\PassOptionsToPackage{table}{xcolor}
\documentclass[11pt]{article}

\usepackage[final]{acl}

\usepackage{times}
\usepackage{latexsym}
\usepackage{booktabs}
\usepackage{tabularx}
\usepackage{multirow}
\usepackage{array}
\usepackage{float}
\usepackage[most]{tcolorbox}
\usepackage[T1]{fontenc}

\usepackage[utf8]{inputenc}
\usepackage{enumitem}
\usepackage[table]{xcolor}

\usepackage{subcaption}
\usepackage[table]{xcolor}
\usepackage{listings}
\usepackage{xspace}
\usepackage{placeins}
\usepackage{titletoc}
\usepackage{amsmath}

\definecolor{ella}{rgb}{0.9764,0.447,0.447}

\setlist[itemize]{leftmargin=*, topsep=2pt, itemsep=1pt, parsep=0pt, partopsep=0pt}
\AtBeginDocument{\setlength{\parskip}{1pt}\setlength{\parsep}{1pt}}

\definecolor{iggHigh}{rgb}{0.91,0.97,0.91}
\definecolor{iggMid}{rgb}{1.0,0.98,0.88}
\definecolor{iggLow}{rgb}{0.99,0.92,0.92}
\definecolor{iggMint}{RGB}{158,216,192}   %
\definecolor{iggYellow}{RGB}{255,216,106} %
\definecolor{iggRed}{RGB}{235,107,107}    %
\newcommand{\scoreHi}[1]{\cellcolor{iggHigh}{#1}}
\newcommand{\scoreMid}[1]{\cellcolor{iggMid}{#1}}
\newcommand{\scoreLo}[1]{\cellcolor{iggLow}{#1}}
\newcommand{\iggConstraintTypeRow}[1]{%
\multicolumn{5}{l}{\rule[-0.2ex]{0pt}{2.1ex}\textbf{#1} \hfill
}\\
\cmidrule(lr){1-5}
}
\newcommand{\iggConstraintSubtypeRow}[1]{%
\multicolumn{5}{l}{\hspace{1.2em}\rule[-0.15ex]{0pt}{2.0ex}
\textit{#1} \hfill \textit{}}\\
}

\newcommand{\keycapaObser}{\raisebox{0.1em}{\includegraphics[height=.8em,trim=0 2em 0em 0]{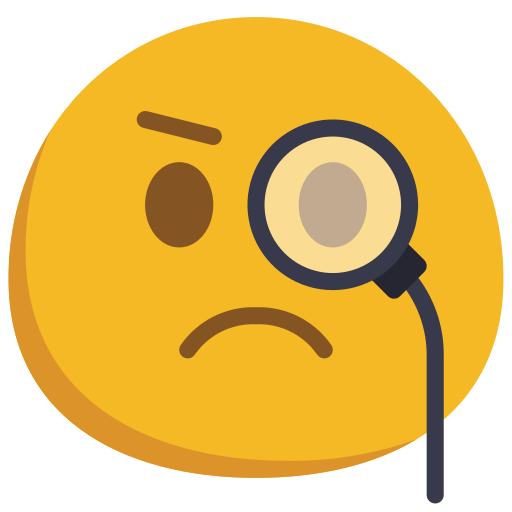}}}
\newcommand{\keycapaDisambiguation}{\raisebox{0.1em}{\includegraphics[height=.8em,trim=0 2em 0em 0]{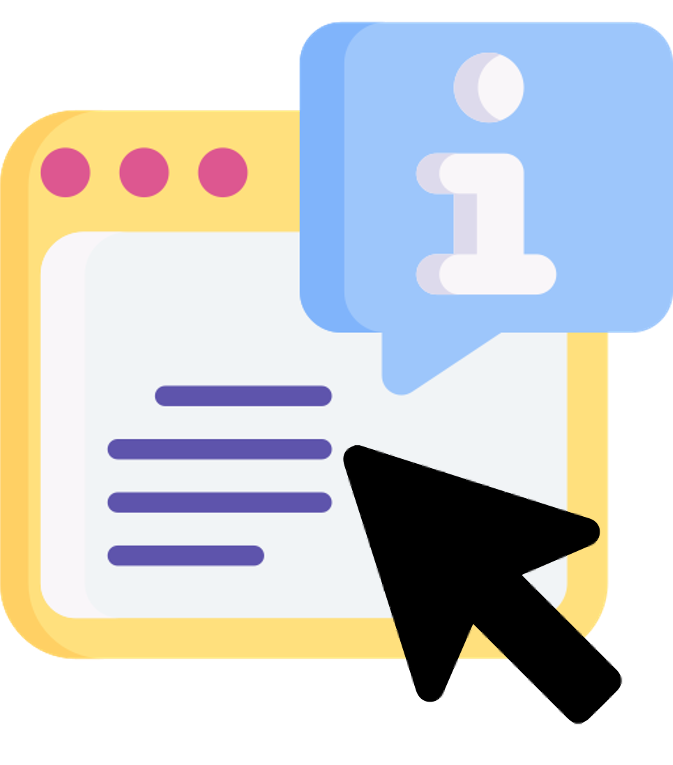}}}
\newcommand{\keycapaState}{\raisebox{0.1em}{\includegraphics[height=.8em,trim=0 2em 0em 0]{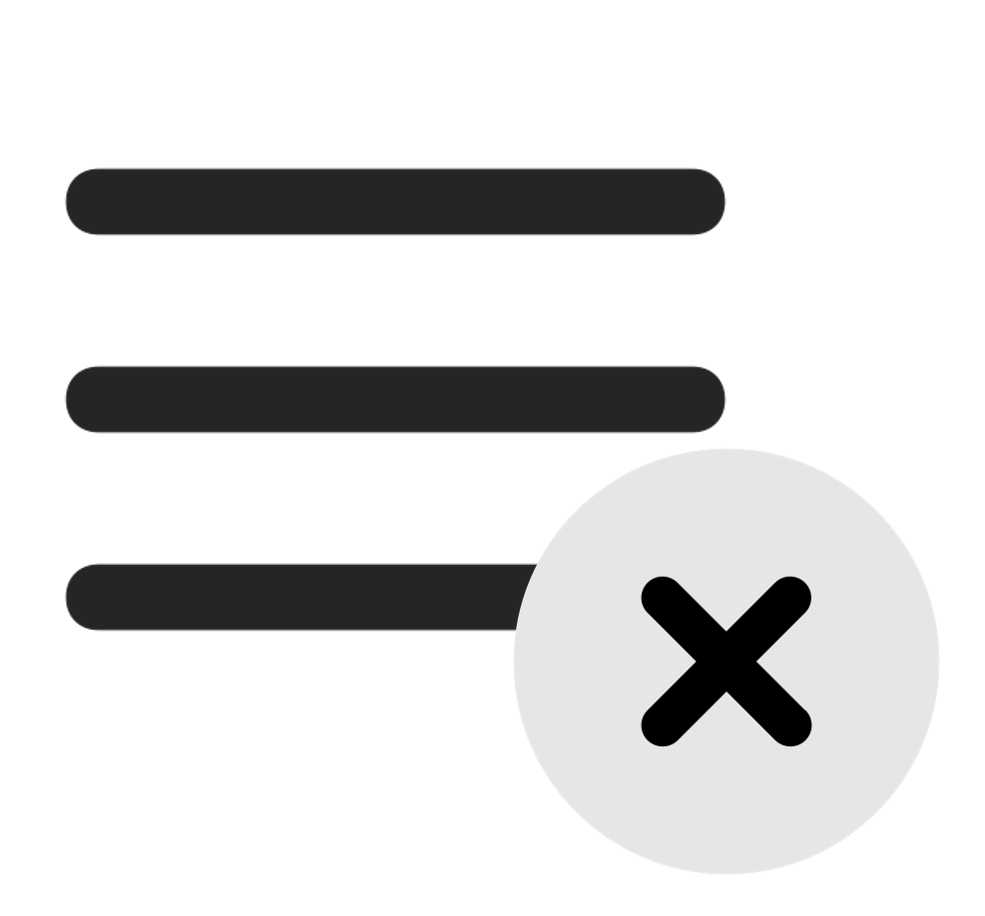}}}
\newcommand{\keycapaTemporal}{\raisebox{0.1em}{\includegraphics[height=.8em,trim=0 2em 0em 0]{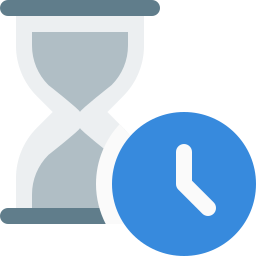}}}

\usepackage{microtype}

\usepackage{inconsolata}

\usepackage{graphicx}
\graphicspath{{figures/}}

\title{
\textcolor{iggMint}{I}\textcolor{iggYellow}{G}\textcolor{iggRed}{G}:
A Benchmark for
\textcolor{iggMint}{I}nteractive
\textcolor{iggYellow}{G}UI
\textcolor{iggRed}{G}rounding
under Visibility Constraints
}
\author{
  \textbf{Kyeongseon Kim\textsuperscript{*, 1}},
  \textbf{Jiyeon Son\textsuperscript{*, 1}},
  \textbf{Tae-Hyun Oh\textsuperscript{1}}
  \\
  \textsuperscript{1}KAIST
  \\
  {\small
    \textbf{Correspondence:}
    \href{mailto:ella94.ai@gmail.com}{ella94.ai@gmail.com}
  }
}

\begin{document}
\maketitle
\begingroup
\renewcommand{\thefootnote}{}
\footnotetext{\textsuperscript{*}Equal contribution.}
\endgroup
\begin{abstract}
\vspace{0.5em}
GUI grounding benchmarks ask whether an agent can localize a target that is already visible in a static screenshot. Real interfaces often violate this assumption: a target may be off-screen, appear only after a state change or hover, remain occluded, or become actionable after a delay. This leaves the ability to recover target visibility unevaluated.
To measure this reveal step, we introduce
\textbf{\textit{Interactive GUI Grounding (IGG)}}%
\footnote{The benchmark and code are available at
\url{https://igg-bench.github.io/}.},
a benchmark in which an agent must first make the target visible through interaction and then click it.
IGG contains 601 tasks across 16 controllable GUI environments and seven visibility-constraint sub-types, with a compact action space of scrolling, moving, dragging, waiting, and clicking.
In our human study, participants achieve 94.6\% accuracy, while GPT-5.5 reaches 79.4\% and the strongest evaluated open-source VLM reaches only 31.6\%.
To identify the source of this gap, we conduct two controlled interventions. Telling Qwen3-VL which interaction to perform raises its accuracy only from 30.8\% to 33.9\%, whereas performing the interaction for the model raises it to 81.9\%. The bottleneck is executing the reveal interaction, not recognizing the target once it is exposed.
We further show that this capability is learnable,
demonstrating that visibility recovery can be strengthened consistently across model families.
\end{abstract}
\section{Introduction}
GUI agents increasingly rely on vision-language models (VLMs) to map natural-language instructions to screen locations~\cite{wu2025gui, seepointrefine, visualwebarena, Kumbhar_2026_CVPR, gao2024enhancing, seeact}.
Existing grounding benchmarks evaluate this mapping within a single screenshot and typically assume that the target is already visible~\cite{screenspot,li2025screenspotpro}.
However, routine interfaces often violate this assumption (Figure~\ref{fig:constraint_type_screen}): a target may lie outside the viewport~\cite{burigat2012dynamic}, sit in a collapsed panel, be covered by a popup, be distinguishable from its neighbors only on hover, or activate only after a delay. In each case, no static grounding model, however accurate, can localize the target in the initial screenshot. The agent must first select and execute an interaction that reveals it, and only then recognize and click it. Static grounding benchmarks leave this reveal step unmeasured, and navigation benchmarks fold it into end-to-end task success, where it cannot be separated from planning. In light of this gap, this paper asks whether current GUI agents can make a target visible before grounding it and, when they cannot, at which stage they fail.
\begin{figure*}[t]
\vspace{-0.5em}
\centering
\includegraphics[width=\textwidth]{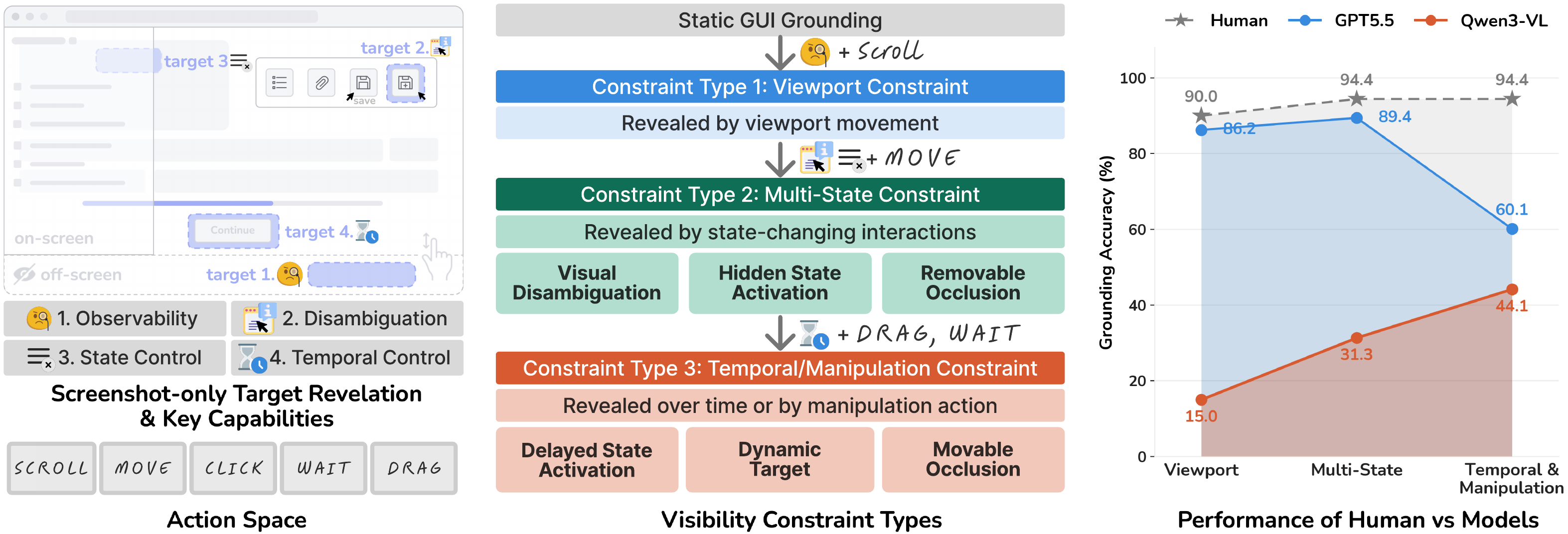}
\caption{\textbf{Overview of Interactive GUI Grounding (IGG).}
\textbf{Left}: representative visibility constraints require observability \keycapaObser, disambiguation \keycapaDisambiguation, state control \keycapaState, or temporal control \keycapaTemporal before the target can be clicked.
\textbf{Center}: these capabilities and a compact action space define \textit{Viewport Constraint}, \textit{Multi-State Constraint}, and \textit{Temporal/Manipulation Constraint}.
\textbf{Right}: human solvability exceeds model grounding accuracy on all three constraint types, and the gap is largest on viewport constraints for Qwen3-VL and on temporal and manipulation constraints for GPT-5.5.
}
\label{fig:overview}
\end{figure*}
To fill this gap, we introduce \textbf{\textit{Interactive GUI Grounding (IGG)}} to measure this missing step.
An IGG episode decomposes success into four stages: \textit{selecting a reveal action}, \textit{executing it at the correct location or with the correct parameters}, \textit{recognizing the target once it appears}, and \textit{localizing the final click}. We call the first two stages \textit{visibility recovery}. The remaining stages, recognizing the target once it appears and localizing the click, coincide with static grounding (Section~\ref{sec:problem_formulation}). Thus, like static grounding, IGG scores a single final click. Unlike static grounding, it additionally requires the agent to recover the target’s visibility beforehand. Because interaction serves only to expose the target, IGG isolates visibility recovery without introducing the broader planning demands of full GUI navigation.

IGG contains 601 tasks across 16 controllable GUI environments modeled on familiar applications. Tasks fall into seven sub-types under three constraint types, \textit{Viewport}, \textit{Multi-State}, and \textit{Temporal/Manipulation}, and are solved with five actions: \textsc{Scroll}, \textsc{Move}, \textsc{Drag}, \textsc{Wait}, and \textsc{Click}. Deterministic initial states and stable target identities make evaluation replayable at click time, while the environments preserve the interaction patterns that make visibility recovery difficult: hover reveals, popups, delayed activation, carousels, and movable occlusions.

Our experiments identify interaction execution as the main bottleneck in IGG.
Telling a model which interaction to perform provides little improvement, whereas performing the reveal action for it recovers most of its accuracy. SFT on successful interaction trajectories followed by online GRPO further improves visibility recovery across Qwen3-VL, InternVL3.5, and GUI-Owl-1.5.

Our contributions are:
\begin{itemize}
    \item We formulate interactive GUI grounding as visibility recovery followed by target recognition and localization, separating it from both static grounding and end-to-end navigation.
    \item We introduce IGG, a benchmark of 601 tasks, 16 controllable GUI environments, and seven visibility-constraint sub-types, together with a click-time evaluator and human-verified ground-truth interaction trajectories.
    \item We provide controlled interventions, heuristic analyses, cross-viewport and paired mock-real evaluations, and training studies that identify where current agents fail and show consistent improvements across three VLM architectures.
\end{itemize}

\section{Related Work}
\paragraph{GUI grounding.}
GUI grounding maps a natural-language instruction to the corresponding interface element~\citep{screenspot,tang2025survey}.
ScreenSpot~\citep{screenspot}, ScreenSpot-Pro~\citep{li2025screenspotpro}, and OSWorld-G~\citep{xie2025osworldg} evaluate localization within a screenshot where the target is visible.
Recent methods improve this setting through rich screenshot supervision~\citep{gao2024enhancing},
diffusion-based prediction~\citep{Kumbhar_2026_CVPR}, UI decomposition~\citep{xie2025osworldg}, and multi-turn refinement~\citep{seepointrefine}.
Universal grounding extends localization to agent settings~\citep{uground}, but still predicts a target from the currently observed state.
IGG instead evaluates the preceding visibility-recovery step: the agent must change the interface before the target can be recognized and localized.

\paragraph{GUI navigation}
Mind2Web~\citep{mind2web}, WebArena~\citep{webarena}, OSWorld~\citep{osworld} evaluate multi-step task completion on web and desktop interfaces, and agent-side improvements such as interaction-history compression~\citep{zhu2025turbocharging} are measured against their end-to-end success.
Because only end-to-end success is scored, planning, exploration, and grounding failures collapse into a single signal.
IGG instead scores one narrower capability, visibility recovery followed by a single evaluated click (Section~\ref{sec:problem_formulation}). Just as static grounding was separated from navigation and became a standard evaluation axis~\citep{screenspot}, IGG isolates a step that navigation benchmarks contain only implicitly, and it complements rather than replaces them.

\section{The IGG Benchmark}
Our goal is to measure visibility recovery in isolation and reproducibly. To this end, we formalize the task (Section~\ref{sec:problem_formulation}), categorize the interactions it requires (Section~\ref{sec:taxonomy}), describe the environments and how tasks were built (Section~\ref{sec:dataset}), specify the evaluation protocol (Section~\ref{sec:eval_setup}), and validate the benchmark with human studies (Section~\ref{sec:human_study}).

\subsection{Problem formulation}
\label{sec:problem_formulation}
An IGG task is specified by an initial interface state in which the target is not visible or not actionable, a goal-only instruction $q$ that identifies the target but not the procedure for exposing it, a target element with a stable identifier, and an evaluator that scores a \textsc{Click} as successful if and only if the clicked element's identifier matches the target's. Each task also carries one human-verified ground-truth trace, which is withheld from the agent and used only for efficiency metrics. At step $t$, the agent receives the screenshot $s_t$, the instruction $q$, and its action history $h_{t-1}=(a_1,\ldots,a_{t-1})$, selects an action from
\[
\mathcal{A} = \{\textsc{Scroll}, \textsc{Move}, \textsc{Drag}, \textsc{Wait}, \textsc{Click}\},
\]
and receives the next screenshot $s_{t+1}$ (Table~\ref{tab:action_space}). IGG uses screenshot-only observations and restricts interaction to these five reveal-and-click actions. The episode ends with success on the first click that hits the target and with failure when the action budget is exhausted. A click elsewhere applies any defined state transition, and the episode continues.

\begin{figure*}[!t]
\centering
\includegraphics[width=\textwidth]{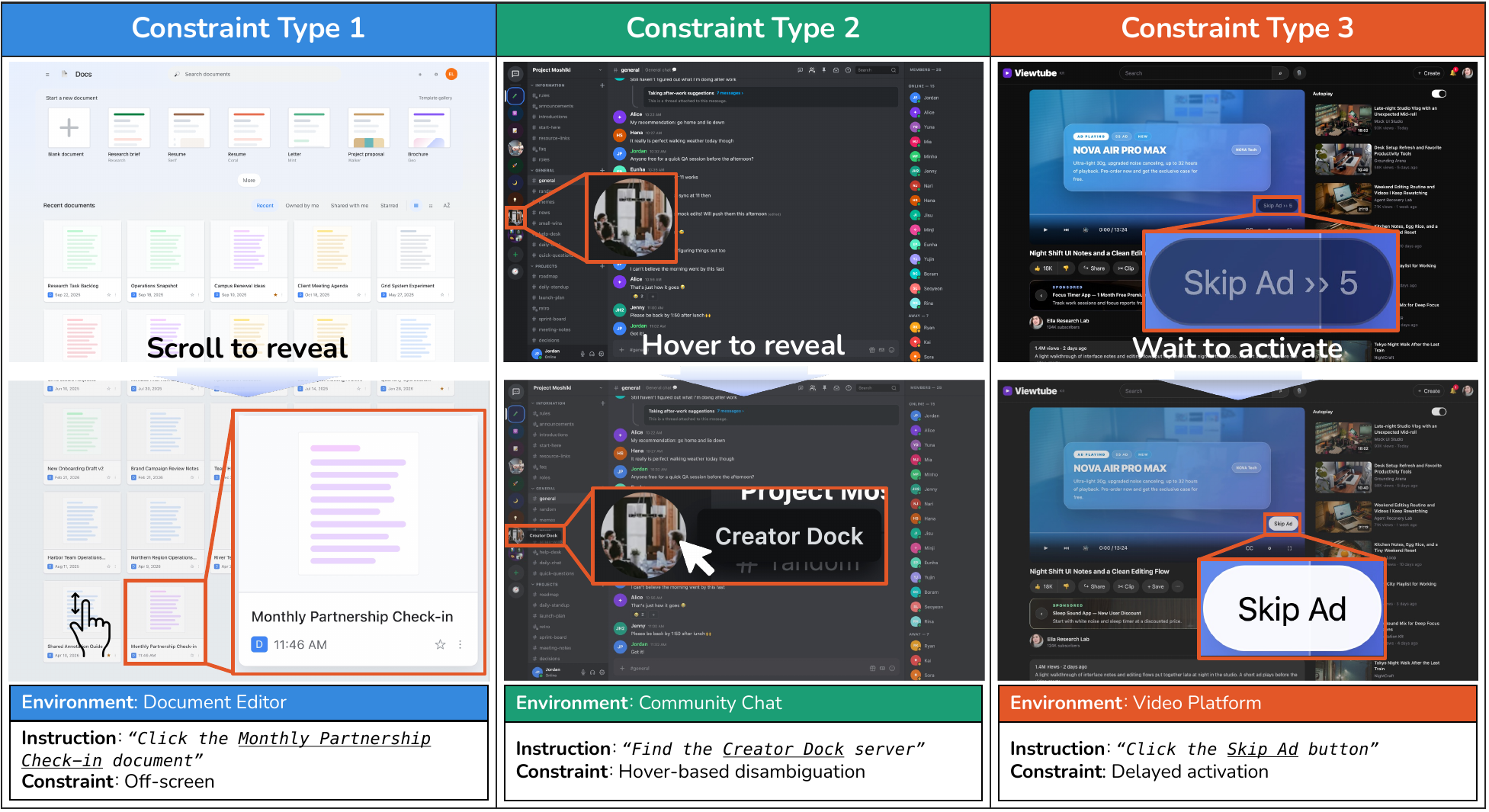}
\caption{Examples of the three IGG constraint types: viewport, multi-state, and temporal or manipulation constraints. Additional examples appear in \hyperref[app:dataset_examples]{Appendix~\ref*{app:dataset_examples}}.
}

\label{fig:constraint_type_screen}
\end{figure*}
\begin{table}[t]
\centering
\footnotesize
\captionsetup{skip=2pt}
\caption{Minimal action space $\mathcal{A}$ for visibility recovery.}
\label{tab:action_space}
\setlength{\tabcolsep}{4pt}
\renewcommand{\arraystretch}{1.2}
\begin{tabularx}{\columnwidth}{@{}p{0.35\columnwidth} >{\raggedright\arraybackslash}X@{}}
\toprule
\textbf{Action} & \textbf{Role} \\
\midrule
\texttt{SCROLL}$(\mathrm{dir}, \Delta_{\mathrm{px}})$ & Scroll to reveal off-screen regions. \\
\texttt{MOVE}$(x_{\mathrm{px}}, y_{\mathrm{px}})$ & Move the pointer to trigger hover states or expose disambiguating cues. \\
\texttt{DRAG}(\texttt{startX}, \texttt{startY}, \texttt{endX}, \texttt{endY})& Drag an element between the specified start and end coordinates. \\
\texttt{WAIT}$(\Delta t_{\mathrm{ms}})$ & Pause for a specified duration to handle delayed or asynchronous UI updates. \\
\texttt{CLICK}$(x_{\mathrm{px}}, y_{\mathrm{px}})$ & Click the specified location; grounding is evaluated at this step. \\
\bottomrule
\end{tabularx}
\vspace{-0.5em}
\end{table}

\subsection{Visibility constraint taxonomy}
\label{sec:taxonomy}
We categorize tasks by the interaction required to recover target visibility (Figures~\ref{fig:overview} and~\ref{fig:constraint_type_screen}).

\paragraph{Constraint Type 1: Viewport Constraint.}
The target is outside the current viewport but reachable within the same UI state.
\begin{itemize}
    \item \textbf{CT1-1. Scroll-based Exploration:} The target is off-screen and must be revealed by scrolling within the current state.
\end{itemize}
\vspace{-0.8em}
\paragraph{Constraint Type 2: Multi-State Constraint.}
The target, or the visual cue needed to identify it, is hidden by the current UI state and must be revealed through a state-changing interaction.
\begin{itemize}
    \item \textbf{CT2-1. Visual Disambiguation:} The target is visually similar to other elements and can only be distinguished through hover-based cues such as tooltips or labels, requiring a \textsc{Move} before the final \textsc{Click}.
    \item \textbf{CT2-2. Hidden State Activation:} The target is contained in a collapsed or inactive UI region, such as a dropdown, tab, or side panel, that must be opened before clicking.
    \item \textbf{CT2-3. Removable Occlusion:} A blocking element, such as a popup or modal, hides the target and must be dismissed before the target can be clicked.
\end{itemize}
\vspace{-0.8em}
\paragraph{Constraint Type 3: Temporal \& Manipulation Constraint.}
Reaching the target requires temporal control or spatial manipulation.
\begin{itemize}
    \item \textbf{CT3-1. Delayed State Activation:} The target is initially inactive and becomes interactable only after a temporal delay or asynchronous event, requiring a \textsc{Wait} before clicking.
    \item \textbf{CT3-2. Dynamic Target Interaction:} The target resides in an automatically transitioning element, such as a carousel or ticker, and must be reached by waiting for the correct state or by using navigation controls.
    \item \textbf{CT3-3. Movable Occlusion:} A UI element occludes the target due to its position or size, requiring a \textsc{Drag} to reposition or resize the occluding element.
\end{itemize}

\subsection{Environments and task construction}
\label{sec:dataset}
\begin{figure}[t!]
    \centering
    \includegraphics[width=\columnwidth]{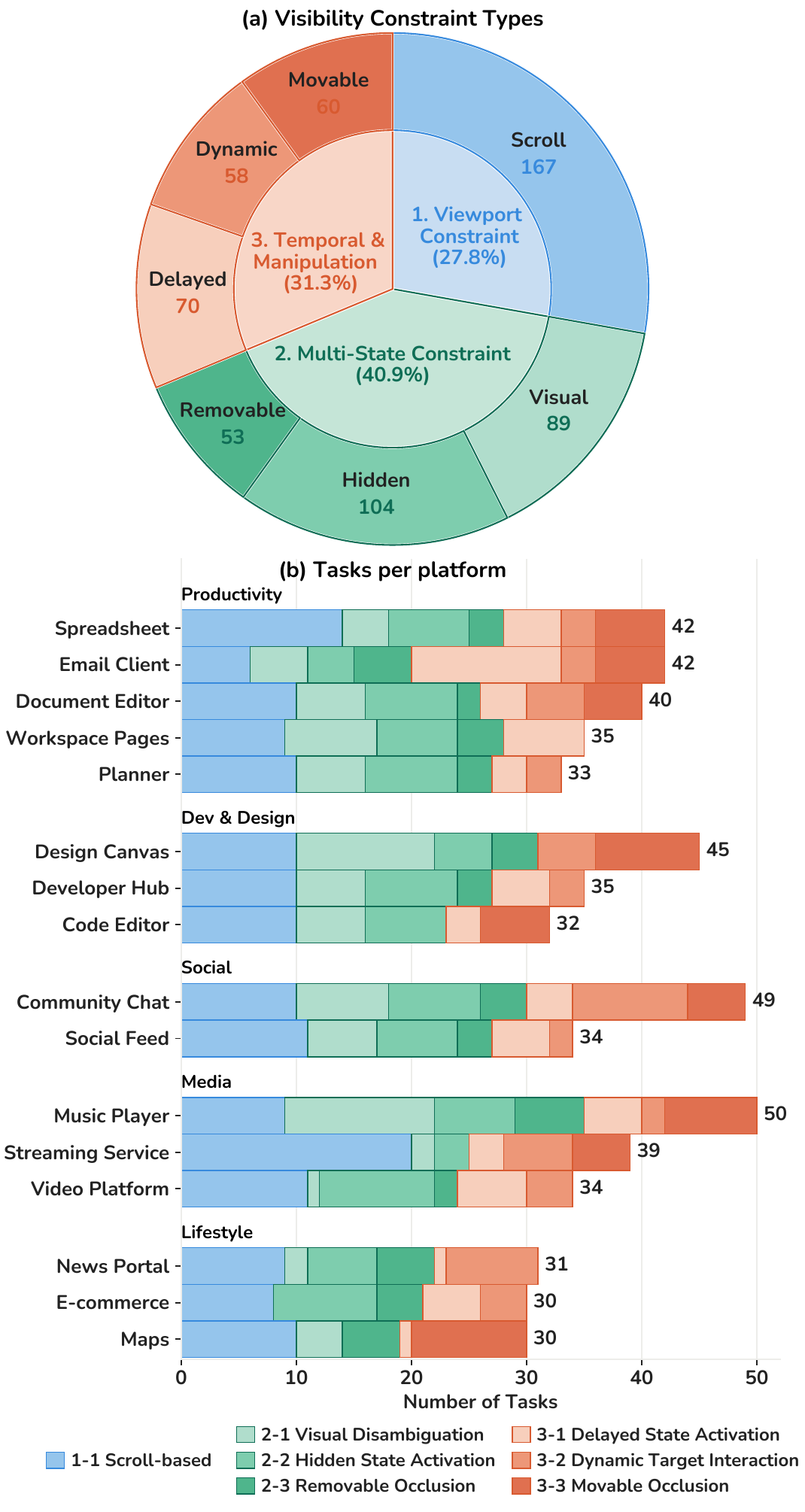}
    \caption{Distribution of IGG tasks. (a) Task distribution across the three constraint types and their sub-types. (b) Number of tasks per platform, broken down by sub-type.}
    \label{fig:distribution}
\end{figure}

We want environments that are deterministic enough to score a click programmatically, yet rich enough in interaction to make visibility recovery difficult. IGG comprises 601 tasks across 16 controllable environments modeled on widely used applications such as Gmail, Google Calendar, VS Code, Figma, Spotify, and YouTube (full list in Appendix~\ref{app:mockups}).\footnote{Real applications are named as canonical references only, and no affiliation with or endorsement by their owners is implied.} Each environment is a deterministic web application: every task starts from a fixed initial state, state transitions are reproducible, and every element carries a stable identifier, so the evaluator decides success at click time without heuristic matching. At the same time, the environments implement the interaction patterns that visibility recovery requires, including hover reveals, collapsible panels, popups, delayed activation, carousels, and draggable occluders.

Tasks are built in three steps. A multimodal agent first drafts a goal-only instruction and a reference trace for each target. Annotators then rewrite instructions that leak the procedure (``Click `Reply' on Minji's message'' rather than ``Hover over the message and click `Reply'\,'', as in Figure~\ref{fig:subtype_l2_1}) and add a natural disambiguating cue, such as an author, title, or surrounding content, when several elements share a label (Appendix~\ref{app:instruction_design}). Finally, annotators replay every trace to confirm that the target is unique and reachable. Figure~\ref{fig:distribution} shows the resulting distribution over sub-types and environments.
\vspace{-0.5em}
\section{Evaluation and Experiments}
\label{sec:evaluation_experiments}

\subsection{Evaluation setup}
\label{sec:eval_setup}
IGG combines task specifications, 16 controllable GUI environments, and a Playwright-driven Chromium evaluator at a $1440\!\times\!900$ viewport. On each \texttt{CLICK}, the evaluator compares the clicked element ID with the target ID; a match ends the episode, while a mismatch returns the updated screenshot.
Models receive only screenshots, the goal-only instruction, and action history. The DOM, accessibility tree, target ID, and ground-truth interaction trajectory remain hidden. Each baseline is evaluated once per task after an environment reset; main runs use an eight-action budget and greedy decoding unless noted. Table~\ref{app:prompt_protocol} provides the prompts, decoding settings, and action formats.

\paragraph{Metrics.}
We report the following metrics:
\begin{itemize}
  \item \textbf{Grounding Accuracy (GA) $\uparrow$ :} the fraction of tasks in which the agent \texttt{clicks} the target element within the action budget.
  \item \textbf{Action Cost (AC) $\downarrow$ :} mean number of actions executed until success or budget exhaustion.
  \item \textbf{Reference Efficiency Ratio (RER) $\uparrow$ :} mean of $|g_i| / |a_i|$ over successful tasks, where $g_i$ is the annotated ground-truth trajectory and $a_i$ is the agent trajectory; values above 1 indicate trajectories shorter than the annotated ground-truth trajectory.
  \item \textbf{Total Time (TT) $\downarrow$ :} mean elapsed wall-clock time per task.
\end{itemize}
GA measures success; AC and RER measure interaction efficiency. TT is reported descriptively: local and API-backed models use different serving stacks, so their runtimes are not hardware-normalized.
\begin{table}[!t]
\centering
\scriptsize
\caption{Results by constraint type and sub-type.}
\label{tab:igg_results}
\setlength{\tabcolsep}{5pt}
\renewcommand{\arraystretch}{1.12}
\begin{tabular}{lcccc}
\toprule
\textbf{Model} & \textbf{GA (\%)} $\uparrow$ & \textbf{AC} $\downarrow$ & \textbf{RER} $\uparrow$ & \textbf{TT (ms)} $\downarrow$ \\
\midrule
\iggConstraintTypeRow{Constraint Type 1: Viewport Constraint}
Qwen2.5-VL-7B      & \scoreLo{10.8} & \scoreLo{7.37} & \scoreMid{0.70} & \scoreHi{\textbf{14718}} \\
Qwen3-VL-8B        & \scoreMid{15.0} & \scoreMid{7.26} & \scoreHi{\textbf{0.73}} & \scoreMid{15246} \\
GUI-Owl-1.5-8B     & \scoreMid{15.0} & \scoreMid{7.27} & \scoreLo{0.62} & \scoreMid{21985} \\
GPT-5.5            & \scoreHi{\textbf{86.2}} & \scoreHi{\textbf{4.29}} & \scoreMid{0.65} & \scoreLo{56355} \\
\midrule
\iggConstraintTypeRow{Constraint Type 2: Multi-State Constraint}
Qwen2.5-VL-7B      & \scoreLo{14.6} & \scoreLo{7.20} & \scoreMid{1.16} & \scoreHi{\textbf{8773}} \\
Qwen3-VL-8B        & \scoreMid{31.3} & \scoreMid{6.21} & \scoreMid{1.13} & \scoreMid{10862} \\
GUI-Owl-1.5-8B     & \scoreMid{31.3} & \scoreMid{6.14} & \scoreHi{\textbf{1.16}} & \scoreMid{14682} \\
GPT-5.5            & \scoreHi{\textbf{89.4}} & \scoreHi{\textbf{3.28}} & \scoreLo{1.08} & \scoreLo{49103} \\
\addlinespace[2pt]
\iggConstraintSubtypeRow{Constraint Type 2-1: Visual Disambiguation}
Qwen2.5-VL-7B      & \scoreLo{18.0} & \scoreLo{6.96} & \scoreMid{1.52} & \scoreHi{\textbf{8048}} \\
Qwen3-VL-8B        & \scoreMid{22.5} & \scoreMid{6.51} & \scoreHi{\textbf{1.73}} & \scoreMid{11287} \\
GUI-Owl-1.5-8B     & \scoreMid{27.0} & \scoreMid{6.22} & \scoreMid{1.71} & \scoreMid{13350} \\
GPT-5.5            & \scoreHi{\textbf{86.5}} & \scoreHi{\textbf{2.91}} & \scoreLo{1.40} & \scoreLo{44982} \\
\addlinespace[2pt]
\iggConstraintSubtypeRow{Constraint Type 2-2: Hidden State Activation}
Qwen2.5-VL-7B      & \scoreLo{9.6} & \scoreLo{7.44} & \scoreMid{0.93} & \scoreHi{\textbf{9181}} \\
Qwen3-VL-8B        & \scoreMid{26.0} & \scoreMid{6.54} & \scoreHi{\textbf{0.97}} & \scoreMid{11251} \\
GUI-Owl-1.5-8B     & \scoreMid{23.1} & \scoreMid{6.67} & \scoreLo{0.87} & \scoreMid{17408} \\
GPT-5.5            & \scoreHi{\textbf{87.5}} & \scoreHi{\textbf{3.73}} & \scoreMid{0.91} & \scoreLo{56739} \\
\addlinespace[2pt]
\iggConstraintSubtypeRow{Constraint Type 2-3: Removable Occlusion}
Qwen2.5-VL-7B      & \scoreLo{18.9} & \scoreLo{7.13} & \scoreLo{0.81} & \scoreHi{\textbf{9193}} \\
Qwen3-VL-8B        & \scoreMid{56.6} & \scoreMid{5.08} & \scoreMid{0.89} & \scoreMid{9384} \\
GUI-Owl-1.5-8B     & \scoreMid{54.7} & \scoreMid{4.96} & \scoreHi{\textbf{0.95}} & \scoreMid{11569} \\
GPT-5.5            & \scoreHi{\textbf{98.1}} & \scoreHi{\textbf{3.02}} & \scoreMid{0.89} & \scoreLo{41038} \\
\midrule
\iggConstraintTypeRow{Constraint Type 3: Temporal \& Manipulation Constraint}
Qwen2.5-VL-7B      & \scoreLo{42.6} & \scoreMid{6.02} & \scoreLo{0.52} & \scoreMid{24467} \\
Qwen3-VL-8B        & \scoreMid{44.1} & \scoreLo{6.08} & \scoreMid{0.63} & \scoreHi{\textbf{10907}} \\
GUI-Owl-1.5-8B     & \scoreMid{46.8} & \scoreMid{5.78} & \scoreMid{0.73} & \scoreMid{11864} \\
GPT-5.5            & \scoreHi{\textbf{60.1}} & \scoreHi{\textbf{4.82}} & \scoreHi{\textbf{0.97}} & \scoreLo{134642} \\
\addlinespace[2pt]
\iggConstraintSubtypeRow{Constraint Type 3-1: Delayed State Activation}
Qwen2.5-VL-7B      & \scoreLo{85.7} & \scoreLo{4.57} & \scoreLo{0.50} & \scoreMid{19820} \\
Qwen3-VL-8B        & \scoreHi{\textbf{95.7}} & \scoreMid{3.74} & \scoreMid{0.63} & \scoreHi{\textbf{6626}} \\
GUI-Owl-1.5-8B     & \scoreMid{94.3} & \scoreMid{3.53} & \scoreMid{0.67} & \scoreMid{7412} \\
GPT-5.5            & \scoreHi{\textbf{95.7}} & \scoreHi{\textbf{2.50}} & \scoreHi{\textbf{0.97}} & \scoreLo{20603} \\
\addlinespace[2pt]
\iggConstraintSubtypeRow{Constraint Type 3-2: Dynamic Target Interaction}
Qwen2.5-VL-7B      & \scoreMid{32.8} & \scoreHi{\textbf{5.81}} & \scoreLo{0.58} & \scoreMid{45702} \\
Qwen3-VL-8B        & \scoreLo{25.9} & \scoreLo{7.02} & \scoreMid{0.60} & \scoreHi{\textbf{12631}} \\
GUI-Owl-1.5-8B     & \scoreMid{36.2} & \scoreMid{6.33} & \scoreMid{0.83} & \scoreMid{13558} \\
GPT-5.5            & \scoreHi{\textbf{37.9}} & \scoreMid{6.47} & \scoreHi{\textbf{0.84}} & \scoreLo{134949} \\
\addlinespace[2pt]
\iggConstraintSubtypeRow{Constraint Type 3-3: Movable Occlusion}
Qwen2.5-VL-7B      & \scoreLo{1.7} & \scoreLo{7.90} & \scoreLo{1.00} & \scoreHi{\textbf{9362}} \\
Qwen3-VL-8B        & \scoreLo{1.7} & \scoreLo{7.90} & \scoreLo{1.00} & \scoreMid{14235} \\
GUI-Owl-1.5-8B     & \scoreLo{1.7} & \scoreMid{7.88} & \scoreHi{\textbf{2.00}} & \scoreMid{15421} \\
GPT-5.5            & \scoreHi{\textbf{40.0}} & \scoreHi{\textbf{5.93}} & \scoreMid{1.10} & \scoreLo{267392} \\
\midrule
\iggConstraintTypeRow{Overall}
Qwen2.5-VL-7B      & \scoreLo{22.3} & \scoreLo{6.88} & \scoreLo{0.72} & \scoreMid{15334} \\
Qwen3-VL-8B        & \scoreMid{30.8} & \scoreMid{6.46} & \scoreMid{0.85} & \scoreHi{\textbf{12094}} \\
GUI-Owl-1.5-8B     & \scoreMid{31.6} & \scoreMid{6.34} & \scoreMid{0.89} & \scoreMid{15830} \\
GPT-5.5            & \scoreHi{\textbf{79.4}} & \scoreHi{\textbf{4.04}} & \scoreHi{\textbf{0.92}} & \scoreLo{77876} \\
\bottomrule
\end{tabular}
\vspace{-2em}
\end{table}

\subsection{Main results}
\label{sec:experiments}
We compare Qwen2.5-VL-7B~\cite{qwen25vl}, Qwen3-VL-8B~\cite{qwen3vl}, and GUI-Owl-1.5-8B~\cite{guiowl15} with the commercial reference model GPT-5.5.

\begin{figure*}[t]
  \centering
  \begin{subfigure}[t]{0.49\textwidth}
    \centering
    \includegraphics[width=\linewidth]{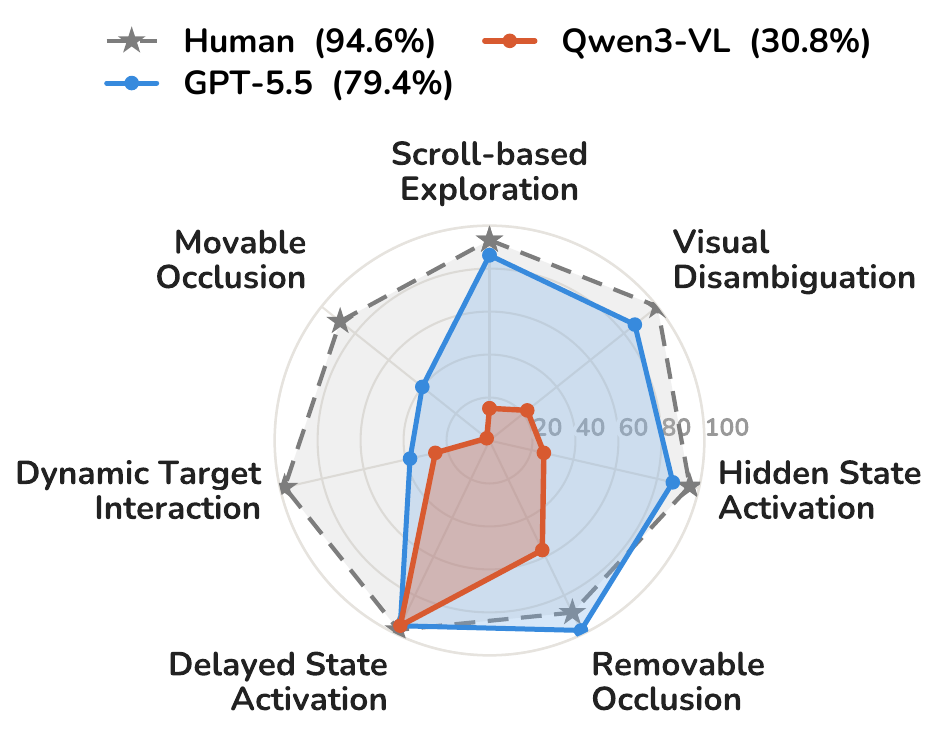}
    \caption{Humans, GPT-5.5, and open-source VLMs.}
    \label{fig:radar-human-gpt-qwen}
  \end{subfigure}\hfill
  \begin{subfigure}[t]{0.49\textwidth}
    \centering
    \includegraphics[width=\linewidth]{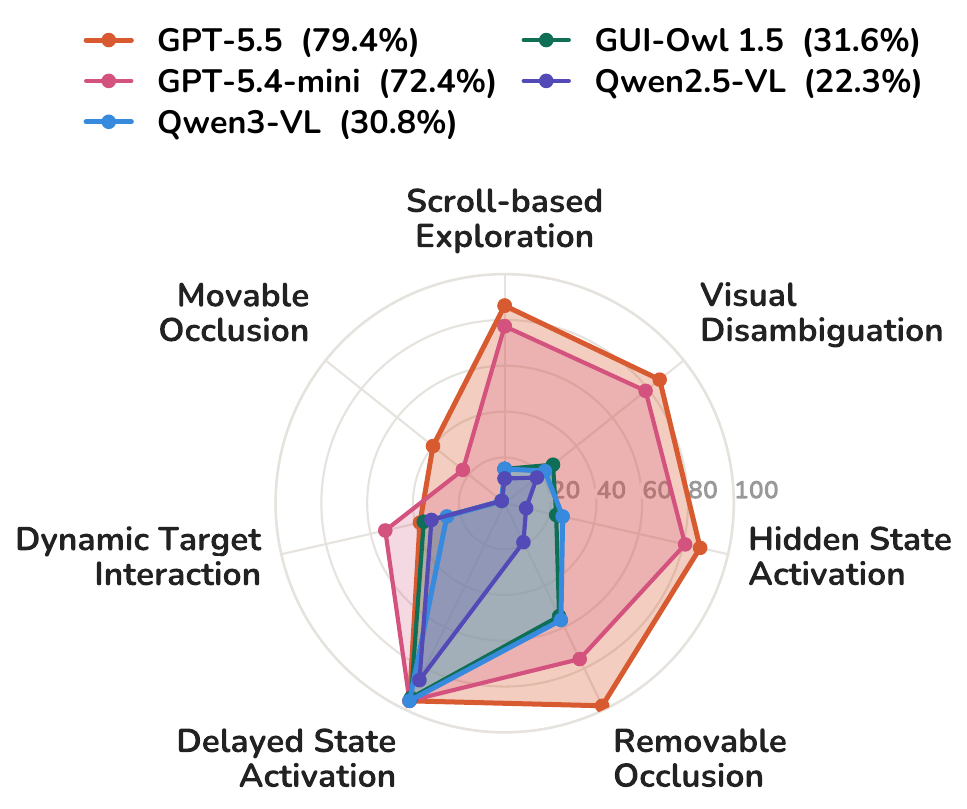}
    \caption{Additional model baselines.}
    \label{fig:radar-all-models}
  \end{subfigure}
  \caption{\textbf{IGG exposes a large interaction gap.} Human results summarize the solvability study, whereas model results span the complete benchmark.
  }
  \label{fig:baseline-radar-comparison}
  \vspace{-0.5em}
\end{figure*}

Table~\ref{tab:igg_results} reports grounding accuracy (GA) and efficiency by constraint type and sub-type; Table~\ref{tab:igg_per_app} aggregates GA by application. Figure~\ref{fig:baseline-radar-comparison} compares humans and model baselines, and \hyperref[tab:per_app_extended_metrics]{Appendix~\ref*{tab:per_app_extended_metrics}} provides extended per-application metrics.

\begin{table}[t]
\vspace{-0.5em}
\centering
\scriptsize
\caption{Grounding Accuracy (GA) (\%) by application.}
\label{tab:igg_per_app}
\setlength{\tabcolsep}{3pt}
\renewcommand{\arraystretch}{1.08}
\begin{tabular}{lcccc}
\toprule
\textbf{App} & \textbf{Qwen2.5-VL} & \textbf{Qwen3-VL} & \textbf{GUI-Owl-1.5} & \textbf{GPT-5.5} \\
\midrule
Social Feed       & 23.5 & 29.4 & 38.2 & \textbf{61.8} \\
Planner           & 15.2 & 21.2 & 24.2 & \textbf{81.8} \\
Developer Hub     & 11.4 & 25.7 & 25.7 & \textbf{57.1} \\
Code Editor       & 25.0 & 21.9 & 28.1 & \textbf{87.5} \\
Community Chat    & 22.4 & 26.5 & 34.7 & \textbf{61.2} \\
Document Editor   & 32.5 & 37.5 & 42.5 & \textbf{85.0} \\
Design Canvas     & 8.9 & 8.9 & 13.3 & \textbf{82.2} \\
Spreadsheet       & 16.7 & 33.3 & 23.8 & \textbf{83.3} \\
Email Client      & 47.6 & 50.0 & 54.8 & \textbf{88.1} \\
News Portal       & 9.7 & 29.0 & 29.0 & \textbf{71.0} \\
Workspace Pages   & 25.7 & 34.3 & 22.9 & \textbf{85.7} \\
Streaming Service & 7.7 & 20.5 & 17.9 & \textbf{87.2} \\
E-commerce        & 16.7 & 43.3 & 33.3 & \textbf{86.7} \\
Music Player      & 20.0 & 36.0 & 32.0 & \textbf{86.0} \\
Maps              & 16.7 & 26.7 & 26.7 & \textbf{70.0} \\
Video Platform    & 55.9 & 50.0 & 58.8 & \textbf{94.1} \\
\midrule
\textbf{Overall (601)} & 22.3 & 30.8 & 31.6 & \textbf{79.4} \\
\bottomrule
\end{tabular}
\vspace{-0.5em}
\end{table}

\begin{figure}[t]
  \centering
\vspace{-0.5em}
  \includegraphics[width=\linewidth]{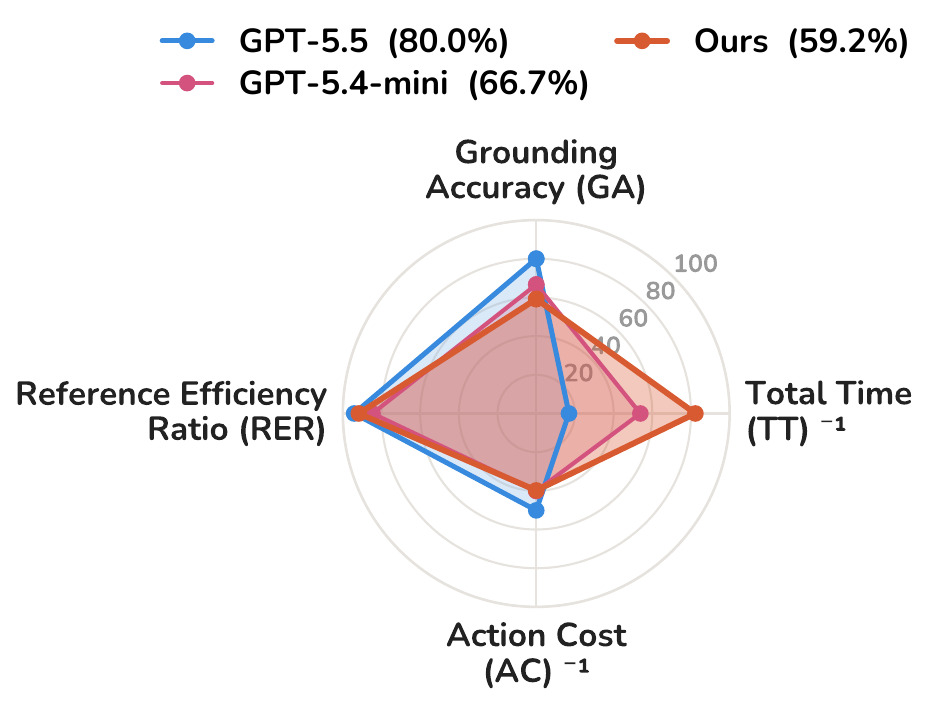}
  \caption{\textbf{Accuracy-efficiency trade-off on the validation set.}
  GPT-5.5 offers substantially higher accuracy, while the trained Qwen3-VL is faster and runs without paid API inference.}
  \label{fig:metrics-radar-speed-cost}
\end{figure}

Figure~\ref{fig:metrics-radar-speed-cost} shows the corresponding accuracy-efficiency trade-off. On the validation comparison, GPT-5.5 is substantially more accurate, but averages 66.5 seconds per task compared with 14.2 seconds for the trained Qwen3-VL and requires paid API inference. Evaluating GPT-5.5 over the complete benchmark cost approximately \$74.9.

\vspace{-0.5em}

\subsection{Where does interactive grounding fail?}
\label{sec:interaction_diagnostic}
We intervene at two points in the four-stage decomposition from Section~\ref{sec:problem_formulation}. In the first condition, the model is told which reveal interaction is required but must determine its location and parameters. In the second, we perform the ground-truth reveal trajectory and ask the model only for the final click.

\begin{table}[t]
\centering
\footnotesize
\caption{Grounding accuracy (\%) when a model solves the complete task, is told the required interaction, or receives the interface after the reveal interaction has been performed.}
\label{tab:interaction_diagnostic}
\setlength{\tabcolsep}{3pt}
\begin{tabular}{lcc}
\toprule
\textbf{Condition} &
\textbf{Qwen3-VL} &
\shortstack{\textbf{GPT-5.4}\\\textbf{Mini (Low)}} \\
\midrule
Full task                         & 30.8 & 72.4 \\
Interaction specified                 & 33.9 & --   \\
Reveal performed                       & 81.9 & 90.4 \\
\bottomrule
\end{tabular}
\vspace{-1em}
\end{table}

Naming the interaction changes Qwen3-VL by only 3.1 points, whereas completing it raises accuracy by 51.1 points. Providing the correct reveal-action category is therefore insufficient. Most remaining errors arise from choosing its location or parameters and executing the reveal correctly, before target recognition and final-click localization are evaluated.

Target-agnostic heuristics further separate easy and hard reveal patterns (Appendix~\ref{app:rule_based_reveal}). Fixed scrolling, hovering, and waiting reveal 68.8\%, 61.4\%, and 87.1\% of their matched tasks, whereas popup dismissal and occluder dragging reach 17.6\% and 16.9\%. The difficult cases require identifying which element controls visibility and how to manipulate it.

Failed Qwen3-VL trajectories show the same pattern. When the ground-truth trajectory requires \textsc{Drag}, \textsc{Scroll}, or \textsc{Move}, failed runs omit that action in 59/59, 207/247, and 71/96 cases, respectively (Appendix~\ref{app:action_prior_diagnostic}). This motivates first expanding the policy's action support with SFT, then improving execution with online feedback.

\subsection{Learning missing interaction priors}
\label{sec:training}
We ask whether training can supply reveal actions that base VLMs rarely use. \textsc{Drag} provides the clearest case: Qwen2.5-VL, Qwen3-VL, and GUI-Owl-1.5 emit no drag actions on the held-out tasks.

\paragraph{Setup.}
After excluding two tasks whose ground-truth trajectories could not be replayed reliably, we use 479 tasks for training and reserve 120 for evaluation. The held-out tasks contribute neither SFT examples nor GRPO rollouts.

\paragraph{Supervised training on successful trajectories~\cite{hu2022lora}.}
We fine-tune the model on human-verified successful trajectories. Let $x_t$ denote the instruction, screenshot, and interaction history at step $t$, and let $a_t$ be the next ground-truth action. We optimize a weighted next-action negative log-likelihood~\cite{pomerleau1991efficient}:
\[
\mathcal{L}_{\mathrm{SFT}}(\theta)
= -\mathrm{E}_{(x_t,a_t)\sim\mathcal{D}}
\left[w(a_t)\log \pi_{\theta}(a_t \mid x_t)\right].
\]
For standard behavior cloning, $w(a_t)=1$. In the frequency-balanced variant~\cite{cui2019class}, we increase $w(a_t)$ for rare reveal actions such as \textsc{Move}, \textsc{Wait}, and \textsc{Drag}; details are in Appendix~\ref{app:training_details}. The model still receives goal-only instructions. Only the training distribution changes.

\paragraph{Refining interactions with online feedback~\cite{shao2024deepseekmath}.}
We then apply online GRPO under the same eight-action interaction budget as main evaluation. The reward is defined over browser-executed trajectories rather than text strings:
\[
R(\tau)
= R_{\mathrm{out}}(\tau)
+ R_{\mathrm{proc}}(\tau)
- \lambda_{\mathrm{step}}|\tau|
- R_{\mathrm{err}}(\tau).
\]
$R_{\mathrm{out}}$ rewards click-time target success, $R_{\mathrm{proc}}$ gives bounded credit for visibility-changing progress, and $R_{\mathrm{err}}$ penalizes wrong clicks, empty clicks, invalid actions, and repeated ineffective actions. Exact weights and optimization hyperparameters are provided in Appendix~\ref{app:training_details}.

The two stages serve different roles. SFT broadens action support, after which GRPO selects among those actions using online feedback. If a seed policy never samples \textsc{Drag}, GRPO observes no successful drag rollout to reinforce. Frequency-balanced SFT brings rare reveal actions into the policy's sampled support, enabling GRPO to optimize their selection and execution.

\begin{figure*}[t]
  \centering
  \begin{subfigure}[t]{0.49\textwidth}
    \centering
    \includegraphics[width=\linewidth]{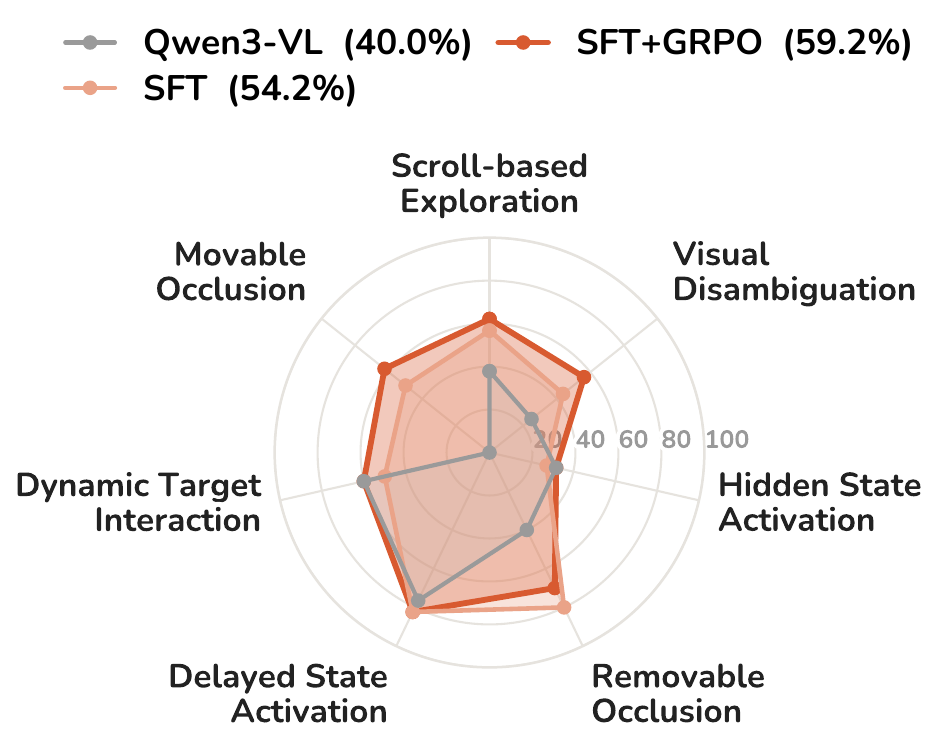}
    \caption{Per-sub-type gains from baseline to SFT and SFT+GRPO.}
    \label{fig:training-progression-radar}
  \end{subfigure}\hfill
  \begin{subfigure}[t]{0.49\textwidth}
    \centering
    \includegraphics[width=\linewidth]{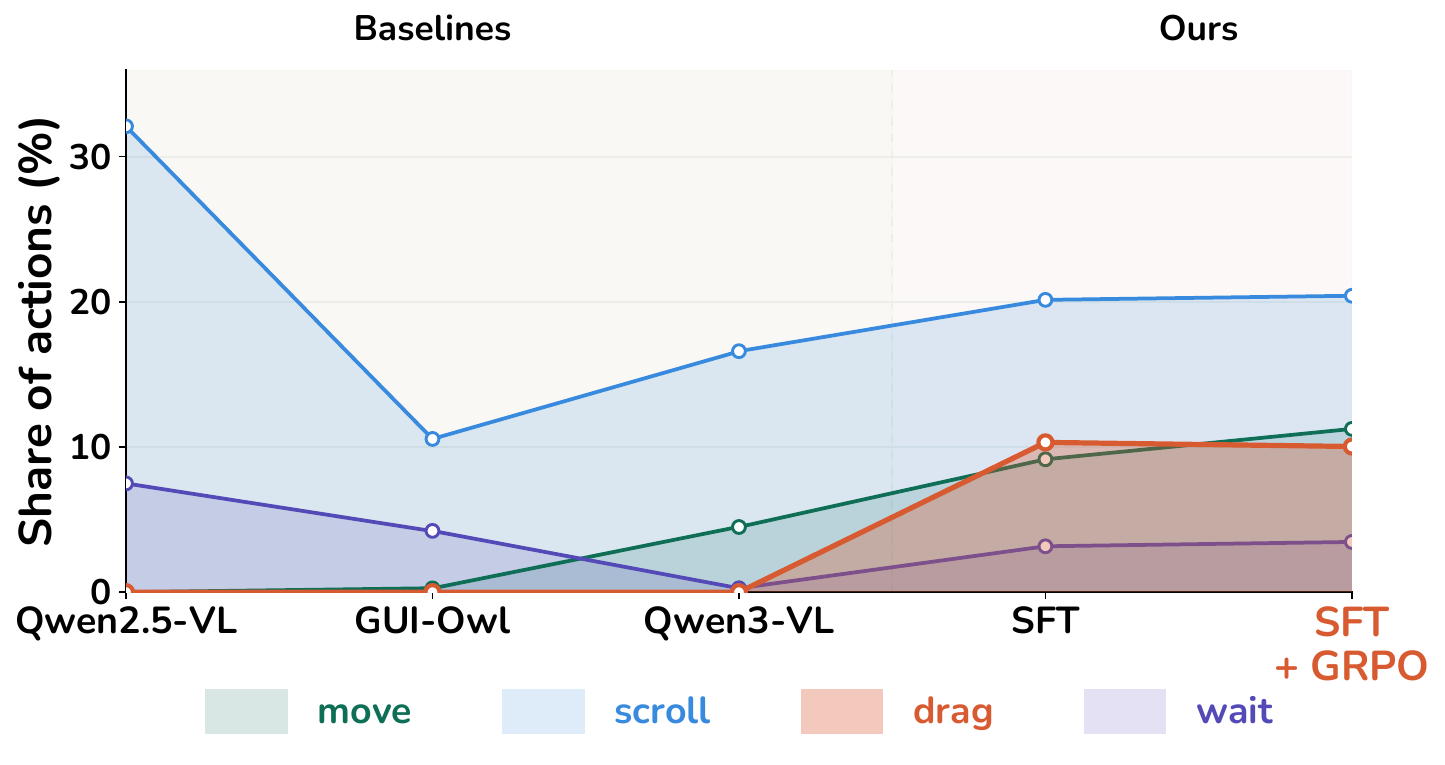}
    \caption{Action-prior shift across baselines, SFT, and SFT+GRPO.}
    \label{fig:action-share-main}
  \end{subfigure}
  \caption{\textbf{SFT expands action support, and GRPO selects among those actions.}
  The best trained model improves Qwen3-VL from 40.0\% to 59.2\% GA on the 120-task validation split.
  Baseline VLMs solve some validation tasks without emitting every action in the ground-truth trajectory, but emit no drag actions. SFT introduces these actions into the policy's support, after which GRPO improves task success.}
  \label{fig:training-main-results}
  \vspace{-0.5em}

\end{figure*}

\paragraph{Results.}
The two-stage procedure raises Qwen3-VL from 40.0\% to 59.2\% GA (Figure~\ref{fig:training-main-results}). On Movable Occlusion, frequency-balanced SFT emits \textsc{Drag} in 7/8 required-drag tasks and solves 4/8; GRPO raises this to 5/8. Initialization also matters on Dynamic Target Interaction: GRPO reaches 60.0\% from the frequency-balanced seed but 30.0\% from the standard SFT seed. The same procedure improves InternVL3.5 from 28.3\% to 47.5\% and GUI-Owl-1.5 from 35.0\% to 52.5\% (Table~\ref{tab:cross_arch}), showing consistent gains across three architectures.

\begin{figure*}[t]
  \centering
  \begin{subfigure}[t]{0.49\textwidth}
    \centering
    \includegraphics[width=\linewidth]{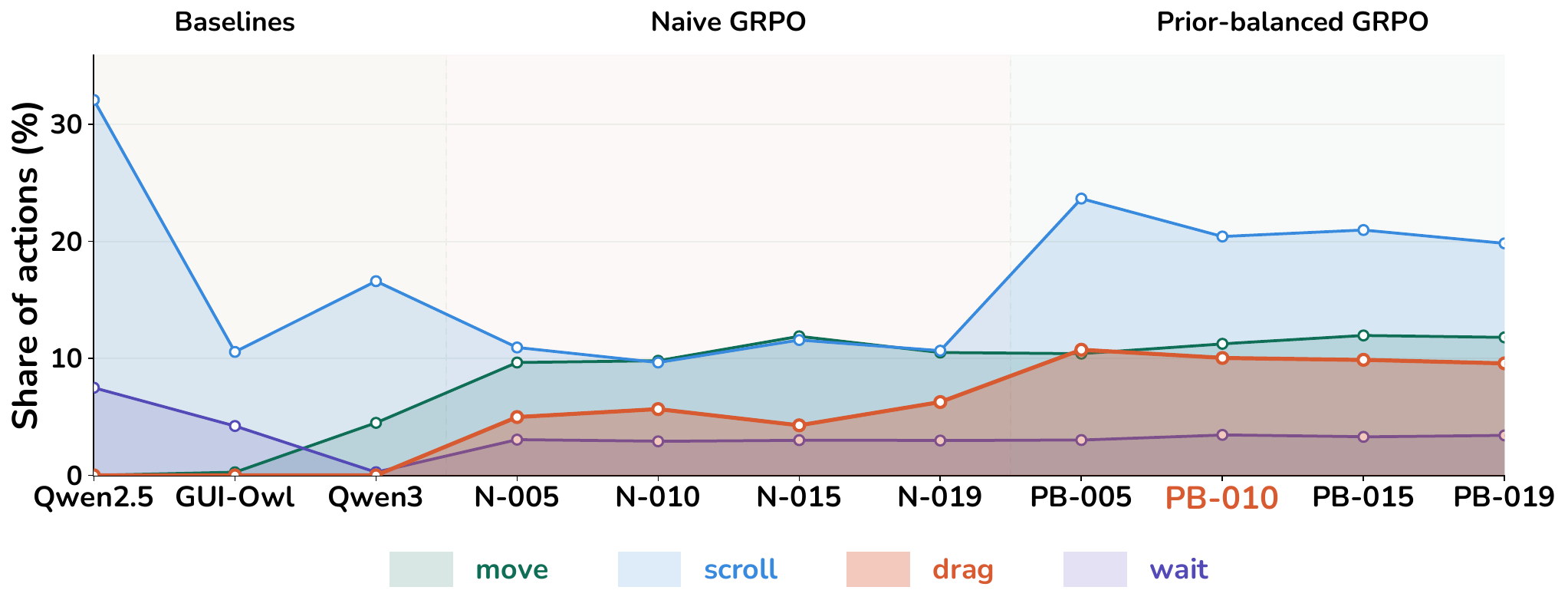}
    \caption{Action share across GRPO variants.}
    \label{fig:grpo-action-share-ablation}
  \end{subfigure}\hfill
  \begin{subfigure}[t]{0.49\textwidth}
    \centering
    \includegraphics[width=\linewidth]{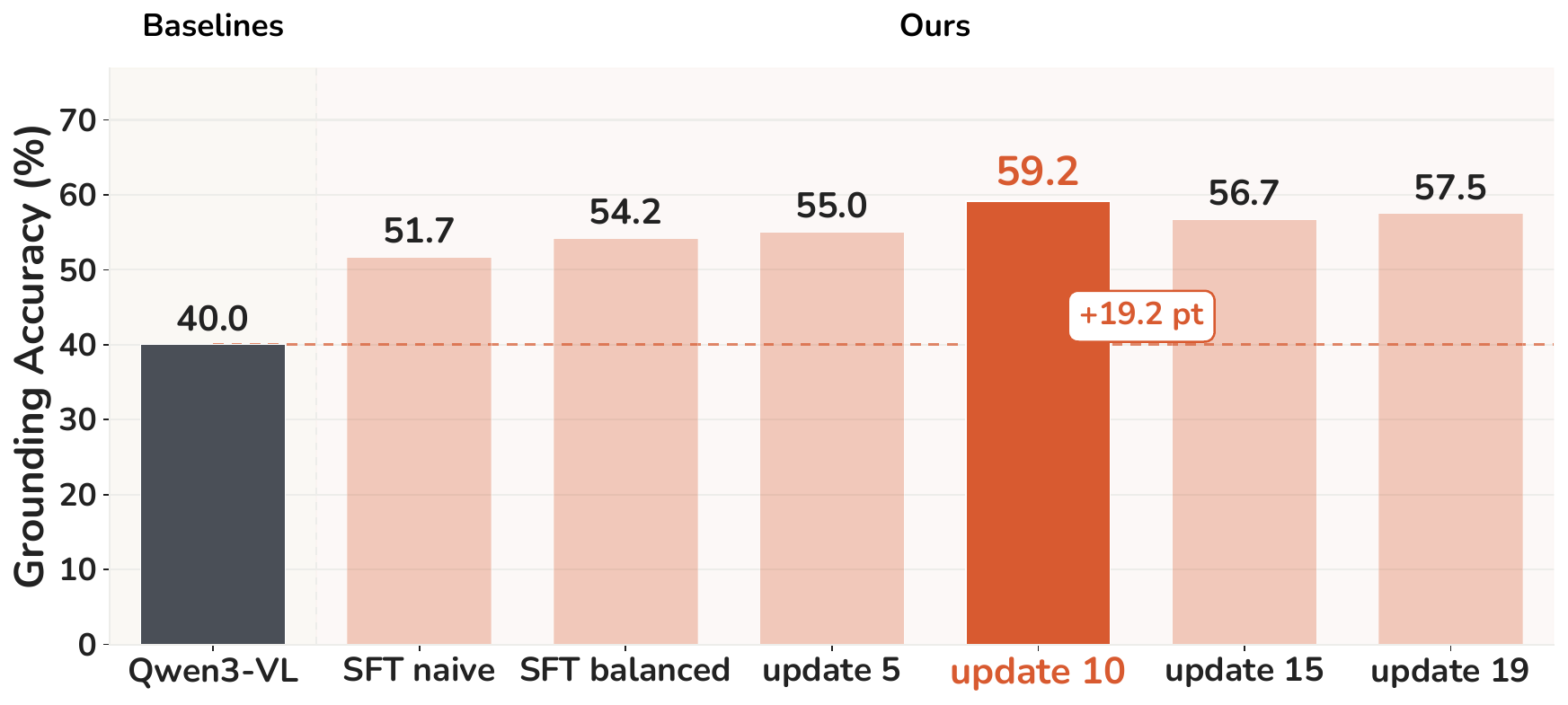}
    \caption{Validation GA across GRPO checkpoints.}
    \label{fig:grpo-ga-ablation}
  \end{subfigure}
  \caption{\textbf{GRPO ablation.}
  GRPO is effective only when initialized from the frequency-balanced SFT seed: it reaches 59.2\% GA, while GRPO from the standard SFT seed does not improve over its seed model.
  The frequency-balanced seed gives the policy access to rare reveal actions, after which online reward improves their selection and execution.}
  \label{fig:grpo-ablation}
  \vspace{-1em}
\end{figure*}

\begin{table}[!t]
\centering
\small
\caption{Human task solvability on the 21-task suite (15 participants, 315 trials).}
\label{tab:human_solvability}
\setlength{\tabcolsep}{4pt}
\renewcommand{\arraystretch}{1.08}
\begin{tabular}{lccc}
\toprule
\textbf{Sub-type} & \textbf{Trials} & \textbf{Success (\%)} $\uparrow$ & \textbf{Time (s)} $\downarrow$ \\
\midrule
CT1-1 & 45 & 93.3 & 23.3 \\
CT2-1 & 45 & 100.0 & 15.9 \\
CT2-2 & 45 & 95.6 & 20.0 \\
CT2-3 & 45 & 88.9 & 21.0 \\
CT3-1 & 45 & 97.8 & 14.2 \\
CT3-2 & 45 & 97.8 & 22.9 \\
CT3-3 & 45 & 88.9 & 21.1 \\
\midrule
Overall & 315 & 94.6 & 19.8 \\
\bottomrule
\end{tabular}
\vspace{-0.5em}
\end{table}

\begin{table}[!t]
\centering
\small
\caption{Human fidelity ratings averaged over 15 participants on a five-point Likert scale: visual realism (VR), interaction realism (IR), and task executability (TE).}
\label{tab:mock_fidelity}
\setlength{\tabcolsep}{5pt}
\renewcommand{\arraystretch}{1.08}
\begin{tabular}{lccc}
\toprule
\textbf{Environment} & \textbf{VR} $\uparrow$ & \textbf{IR} $\uparrow$ & \textbf{TE} $\uparrow$ \\
\midrule
Music Player & 4.93 & 4.80 & 5.00 \\
Video Platform & 4.73 & 4.87 & 4.80 \\
E-commerce & 4.67 & 4.80 & 4.80 \\
Developer Hub & 4.67 & 4.53 & 4.80 \\
Code Editor & 4.67 & 4.80 & 4.87 \\
News Portal & 3.93 & 4.60 & 4.60 \\
Maps & 3.93 & 4.07 & 4.40 \\
\midrule
Average & 4.50 & 4.64 & 4.75 \\
\bottomrule
\end{tabular}
\vspace{-0.5em}
\end{table}

\subsection{Human solvability and environment fidelity}
\label{sec:human_study}
A failure on IGG is informative only if the task is solvable and the environment is faithful to the interface it models. Before comparing models, we therefore verify both with the same 15 participants (protocol details in Appendix~\ref{app:human_study}).
\vspace{-0.5em}
\paragraph{Task solvability.}
Participants solve a fixed suite of 21 tasks, three from each sub-type, using the same instructions and interfaces given to the models, without the constraint label, target ID, target coordinates, or reveal procedure. A trial ends on the correct target click or after eight click-or-drag attempts. Scrolling and hovering are recorded but do not count, so the limit is comparable to, though not identical with, the eight-action model budget. Participants succeed in 298 of 315 trials (94.6\%) with a mean completion time of 19.8 seconds (Table~\ref{tab:human_solvability}). Together with the replay verification of every reference trace, this indicates that the model failures reported below do not stem from unsolvable tasks or ambiguous instructions.

\paragraph{Environment fidelity.}
The same participants compare seven representative environments with their real-interface counterparts and rate visual realism, interaction realism, and task executability on five-point scales, giving mean scores of 4.50, 4.64, and 4.75 (Table~\ref{tab:mock_fidelity}). Appendix~\ref{appendix:fidelity} complements these ratings with visual-similarity metrics and a preliminary paired mock--real evaluation on 16 tasks, one per environment, in which model ordering is preserved across the two settings and the Qwen3-VL model trained in Section~\ref{sec:training} leads in both.

\section{Conclusion}
We introduced IGG to make visibility recovery a measurable component of GUI grounding. Our analyses show that knowing which interaction is required is not enough; current models most often fail in deciding where and how to carry it out. Training on successful interaction trajectories followed by online feedback improves this capability consistently across model families. IGG therefore provides a focused setting for evaluating and learning the interactions that must precede a grounding click.
\section{Limitations}
IGG uses deterministic mockups for controlled and reproducible evaluation. Our fidelity analyses, including paired mock-real evaluation, suggest that these environments capture relevant visual and interaction patterns, but they do not replace evaluation on live interfaces.
Interactive grounding can support accessibility and UI testing, but the same capability may also be misused for automated interaction. We therefore release only controlled mockups that are not connected to external services.
\section*{Acknowledgments}
This work was supported by Institute of Information \& communications Technology Planning \& Evaluation (IITP) grant funded by the Korea government(MSIT) (No. RS-2026-25518317 (50\%), Development of AI memory mechanism that reflects human cognitive principles; No. RS-2024-00457882 (50\%), National AI Research Lab Project), InnoCORE program of the Ministry of Science and ICT(N10250156), and Samsung Electronics Co., Ltd (Project Code: IO260114-15161-01).
\nocite{*}
\bibliography{custom}

@inproceedings{burigat2012dynamic,
  title={Dynamic visualization of large numbers of off-screen objects on mobile devices: an experimental comparison of wedge and overview+ detail},
  author={Burigat, Stefano and Chittaro, Luca and Vianello, Andrea},
  booktitle={Proceedings of the 14th international conference on Human-computer interaction with mobile devices and services},
  pages={93--102},
  year={2012}
}

@inproceedings{seeact,
  title={GPT-4V(ision) is a Generalist Web Agent, if Grounded},
  author={Boyuan Zheng and Boyu Gou and Jihyung Kil and Huan Sun and Yu Su},
  booktitle={Forty-first International Conference on Machine Learning},
  year={2024},
  url={https://openreview.net/forum?id=piecKJ2DlB},
}

@inproceedings{uground,
title={Navigating the Digital World as Humans Do: Universal Visual Grounding for {GUI} Agents},
author={Boyu Gou and Ruohan Wang and Boyuan Zheng and Yanan Xie and Cheng Chang and Yiheng Shu and Huan Sun and Yu Su},
booktitle={The Thirteenth International Conference on Learning Representations},
year={2025},
url={https://openreview.net/forum?id=kxnoqaisCT}
}

@article{shao2024deepseekmath,
  title={Deepseekmath: Pushing the limits of mathematical reasoning in open language models},
  author={Shao, Zhihong and Wang, Peiyi and Zhu, Qihao and Xu, Runxin and Song, Junxiao and Bi, Xiao and Zhang, Haowei and Zhang, Mingchuan and Li, YK and others},
  journal={arXiv preprint arXiv:2402.03300},
  year={2024}
}

@inproceedings{cui2019class,
  title={Class-balanced loss based on effective number of samples},
  author={Cui, Yin and Jia, Menglin and Lin, Tsung-Yi and Song, Yang and Belongie, Serge},
  booktitle={Proceedings of the IEEE/CVF conference on computer vision and pattern recognition},
  pages={9268--9277},
  year={2019}
}

@article{pomerleau1991efficient,
  title={Efficient training of artificial neural networks for autonomous navigation},
  author={Pomerleau, Dean A},
  journal={Neural computation},
  volume={3},
  number={1},
  pages={88--97},
  year={1991},
  publisher={MIT Press}
}

@inproceedings{hu2022lora,
title={Lo{RA}: Low-Rank Adaptation of Large Language Models},
author={Edward J Hu and yelong shen and Phillip Wallis and Zeyuan Allen-Zhu and Yuanzhi Li and Shean Wang and Lu Wang and Weizhu Chen},
booktitle={International Conference on Learning Representations},
year={2022},
url={https://openreview.net/forum?id=nZeVKeeFYf9}
}

@inproceedings{li2025screenspotpro,
  title={Screenspot-pro: Gui grounding for professional high-resolution computer use},
  author={Li, Kaixin and Meng, Ziyang and Lin, Hongzhan and Luo, Ziyang and Tian, Yuchen and Ma, Jing and Huang, Zhiyong and Chua, Tat-Seng},
  booktitle={Proceedings of the 33rd ACM International Conference on Multimedia},
  pages={8778--8786},
  year={2025}
}

@inproceedings{webarena,
  title={Webarena: A realistic web environment for building autonomous agents},
  author={Zhou, Shuyan and Xu, Frank F and Zhu, Hao and Zhou, Xuhui and Lo, Robert and Sridhar, Aman and Cheng, Xianyi and Ou, Tianyue and Bisk, Yonatan and Fried, Daniel and Alon, Uri and Neubig, Graham},
  booktitle={International Conference on Learning Representations},
  year={2024}
}

@inproceedings{visualwebarena,
  title={Visualwebarena: Evaluating multimodal agents on realistic visual web tasks},
  author={Koh, Jing Yu and Lo, Robert and Jang, Lawrence and Duvvur, Vikram and Lim, Ming and Huang, Po-Yu and Neubig, Graham and Zhou, Shuyan and Salakhutdinov, Russ and Fried, Daniel},
  booktitle={Proceedings of the 62nd Annual Meeting of the Association for Computational Linguistics (Volume 1: Long Papers)},
  pages={881--905},
  year={2024}
}

@article{mind2web,
  title={Mind2web: Towards a generalist agent for the web},
  author={Deng, Xiang and Gu, Yu and Zheng, Boyuan and Chen, Shijie and Stevens, Sam and Wang, Boshi and Sun, Huan and Su, Yu},
  journal={Advances in Neural Information Processing Systems},
  volume={36},
  pages={28091--28114},
  year={2023}
}

@article{wu2025gui,
  title={Gui-actor: Coordinate-free visual grounding for gui agents},
  author={Wu, Qianhui and Cheng, Kanzhi and Yang, Rui and Zhang, Chaoyun and Yang, Jianwei and Jiang, Huiqiang and Mu, Jian and Peng, Baolin and Qiao, Bo and Tan, Reuben and others},
  journal={arXiv preprint arXiv:2506.03143},
  year={2025}
}

@inproceedings{screenspot,
  title={Seeclick: Harnessing gui grounding for advanced visual gui agents},
  author={Cheng, Kanzhi and Sun, Qiushi and Chu, Yougang and Xu, Fangzhi and YanTao, Li and Zhang, Jianbing and Wu, Zhiyong},
  booktitle={Proceedings of the 62nd Annual Meeting of the Association for Computational Linguistics (Volume 1: Long Papers)},
  pages={9313--9332},
  year={2024}
}

@article{osworld,
  title={Osworld: Benchmarking multimodal agents for open-ended tasks in real computer environments},
  author={Xie, Tianbao and Zhang, Danyang and Chen, Jixuan and Li, Xiaochuan and Zhao, Siheng and Cao, Ruisheng and Hua, Toh J and Cheng, Zhoujun and Shin, Dongchan and Lei, Fangyu and others},
  journal={Advances in Neural Information Processing Systems},
  volume={37},
  pages={52040--52094},
  year={2024}
}

@article{seepointrefine,
  title={See, Point, Refine: Multi-Turn Approach to GUI Grounding with Visual Feedback},
  author={Mittal, Himangi and Mittal, Gaurav and Troncoso, Nelson Daniel and Hu, Yu},
  journal={arXiv preprint arXiv:2604.13019},
  year={2026}
}

@article{guiowl15,
  title={Mobile-agent-v3. 5: Multi-platform fundamental gui agents},
  author={Xu, Haiyang and Zhang, Xi and Liu, Haowei and Wang, Junyang and Zhu, Zhaozai and Zhou, Shengjie and Hu, Xuhao and Gao, Feiyu and Cao, Junjie and Wang, Zihua and others},
  journal={arXiv preprint arXiv:2602.16855},
  year={2026}
}

@article{qwen3vl,
  title={Qwen3-vl technical report},
  author={Bai, Shuai and Cai, Yuxuan and Chen, Ruizhe and Chen, Keqin and Chen, Xionghui and Cheng, Zesen and Deng, Lianghao and Ding, Wei and Gao, Chang and Ge, Chunjiang and others},
  journal={arXiv preprint arXiv:2511.21631},
  year={2025}
}

@misc{qwen25vl,
      title={Qwen2.5-VL Technical Report}, 
      author={Shuai Bai and Keqin Chen and Xuejing Liu and Jialin Wang and Wenbin Ge and Sibo Song and Kai Dang and Peng Wang and Shijie Wang and Jun Tang and Humen Zhong and Yuanzhi Zhu and Mingkun Yang and Zhaohai Li and Jianqiang Wan and Pengfei Wang and Wei Ding and Zheren Fu and Yiheng Xu and Jiabo Ye and Xi Zhang and Tianbao Xie and Zesen Cheng and Hang Zhang and Zhibo Yang and Haiyang Xu and Junyang Lin},
      year={2025},
      eprint={2502.13923},
      archivePrefix={arXiv},
      primaryClass={cs.CV},
      url={https://arxiv.org/abs/2502.13923}, 
}

@inproceedings{xie2025osworldg,
  title={Scaling Computer-Use Grounding via User Interface Decomposition and Synthesis},
  author={Xie, Tianbao and Deng, Jiaqi and Li, Xiaochuan and Yang, Junlin and Wu, Haoyuan and Chen, Jixuan and Hu, Wenjing and Wang, Xinyuan and Xu, Yuhui and Wang, Zekun and others},
  booktitle={Advances in Neural Information Processing Systems Datasets and Benchmarks Track},
  year={2025}
}

@inproceedings{caron2021emerging,
  title={Emerging Properties in Self-Supervised Vision Transformers},
  author={Caron, Mathilde and Touvron, Hugo and Misra, Ishan and J{\'e}gou, Herv{\'e} and Mairal, Julien and Bojanowski, Piotr and Joulin, Armand},
  booktitle={Proceedings of the IEEE/CVF International Conference on Computer Vision},
  pages={9650--9660},
  year={2021}
}

@inproceedings{zhai2023sigmoid,
  title={Sigmoid Loss for Language Image Pre-Training},
  author={Zhai, Xiaohua and Mustafa, Basil and Kolesnikov, Alexander and Beyer, Lucas},
  booktitle={Proceedings of the IEEE/CVF International Conference on Computer Vision},
  pages={11975--11986},
  year={2023}
}

@inproceedings{fu2023dreamsim,
  title={DreamSim: Learning New Dimensions of Human Visual Similarity using Synthetic Data},
  author={Fu, Stephanie and Tamir, Netanel and Sundaram, Shobhita and Chai, Lucy and Zhang, Richard and Dekel, Tali and Isola, Phillip},
  booktitle={Advances in Neural Information Processing Systems},
  volume={36},
  pages={50742--50768},
  year={2023}
}

@inproceedings{Kumbhar_2026_CVPR,
    author    = {Kumbhar, Shrinidhi and Liao, Haofu and Appalaraju, Srikar and Singh, Kunwar Yashraj},
    title     = {Towards GUI Agents: Vision-Language Diffusion Models for GUI Grounding},
    booktitle = {Proceedings of the IEEE/CVF Conference on Computer Vision and Pattern Recognition (CVPR)},
    month     = {June},
    year      = {2026},
    pages     = {27493-27502}
}

@inproceedings{zhu2025turbocharging,
  title={Turbocharging Web Automation: The Impact of Compressed History States},
  author={Zhu, Xiyue and Tang, Peng and Liao, Haofu and Appalaraju, Srikar},
  booktitle={Findings of the Association for Computational Linguistics: ACL 2025},
  pages={3644--3651},
  year={2025},
  publisher={Association for Computational Linguistics},
  doi={10.18653/v1/2025.findings-acl.187},
  url={https://aclanthology.org/2025.findings-acl.187/}
}

@inproceedings{gao2024enhancing,
  title={Enhancing vision-language pre-training with rich supervisions},
  author={Gao, Yuan and Shi, Kunyu and Zhu, Pengkai and Belval, Edouard and Nuriel, Oren and Appalaraju, Srikar and Ghadar, Shabnam and Tu, Zhuowen and Mahadevan, Vijay and Soatto, Stefano},
  booktitle={2024 IEEE/CVF Conference on Computer Vision and Pattern Recognition (CVPR)},
  pages={13480--13491},
  year={2024},
  organization={IEEE}
}

@article{tang2025survey,
  title={A survey on (m) llm-based gui agents},
  author={Tang, Fei and Xu, Haolei and Zhang, Hang and Chen, Siqi and Wu, Xingyu and Shen, Yongliang and Zhang, Wenqi and Hou, Guiyang and Tan, Zeqi and Yan, Yuchen and others},
  journal={arXiv preprint arXiv:2504.13865},
  year={2025}
}

\clearpage
\appendix
\raggedbottom

\setcounter{figure}{0}
\setcounter{table}{0}
\renewcommand{\thefigure}{A\arabic{figure}}
\renewcommand{\thetable}{A\arabic{table}}

\newcommand{\ctype}[2]{\(\mathrm{CT}_{#1\text{-}#2}\)}
\newcommand{\ctypeb}[2]{\textbf{CT\textsubscript{#1-#2}}}

\section{Mockup Applications}
\label{app:mockups}

\begin{figure*}[!htbp]
\centering
\begin{subfigure}{0.49\linewidth}
  \centering
  \includegraphics[width=\linewidth]{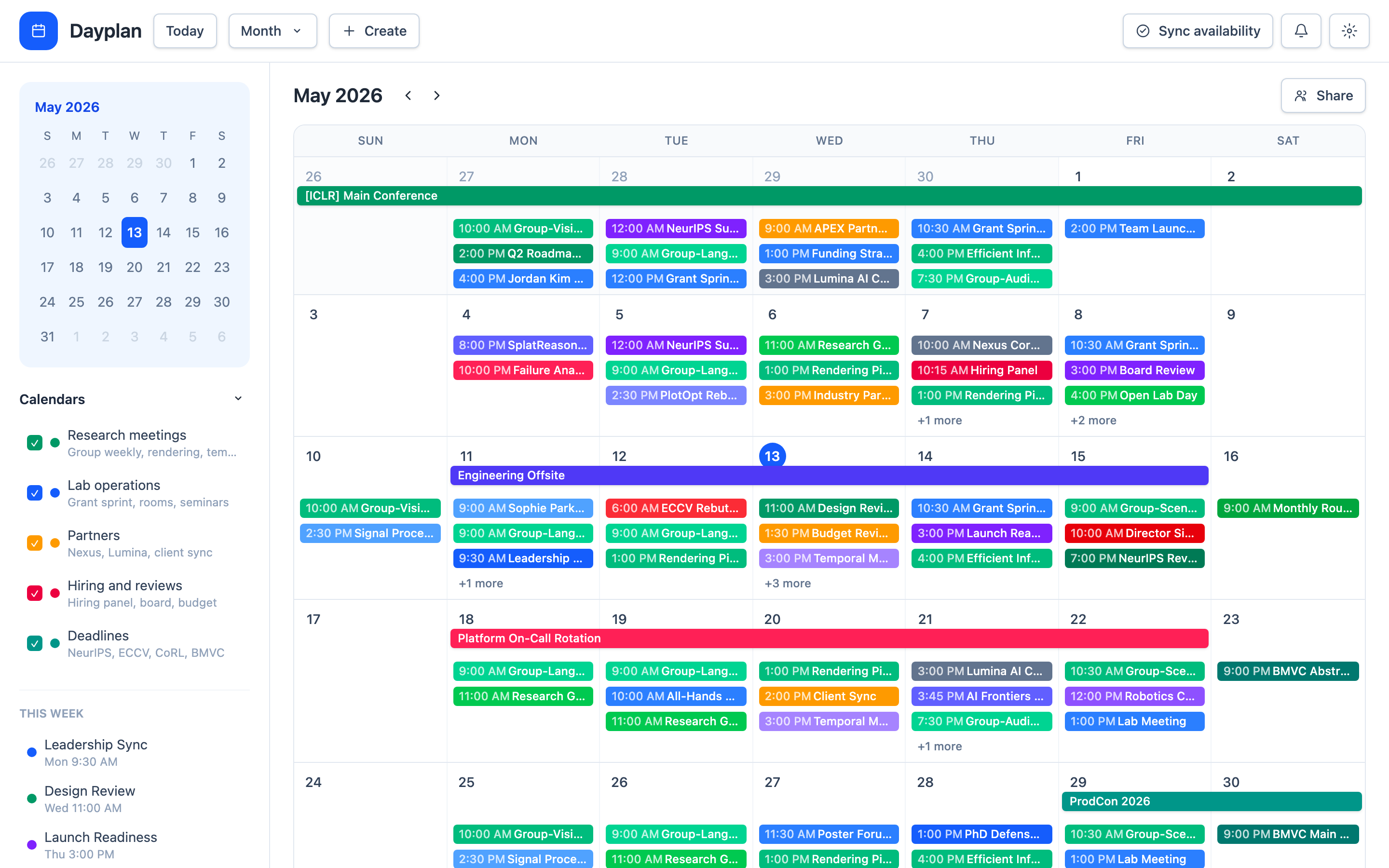}
  \caption{Planner (Google Calendar)}
\end{subfigure}
\hfill
\begin{subfigure}{0.49\linewidth}
  \centering
  \includegraphics[width=\linewidth]{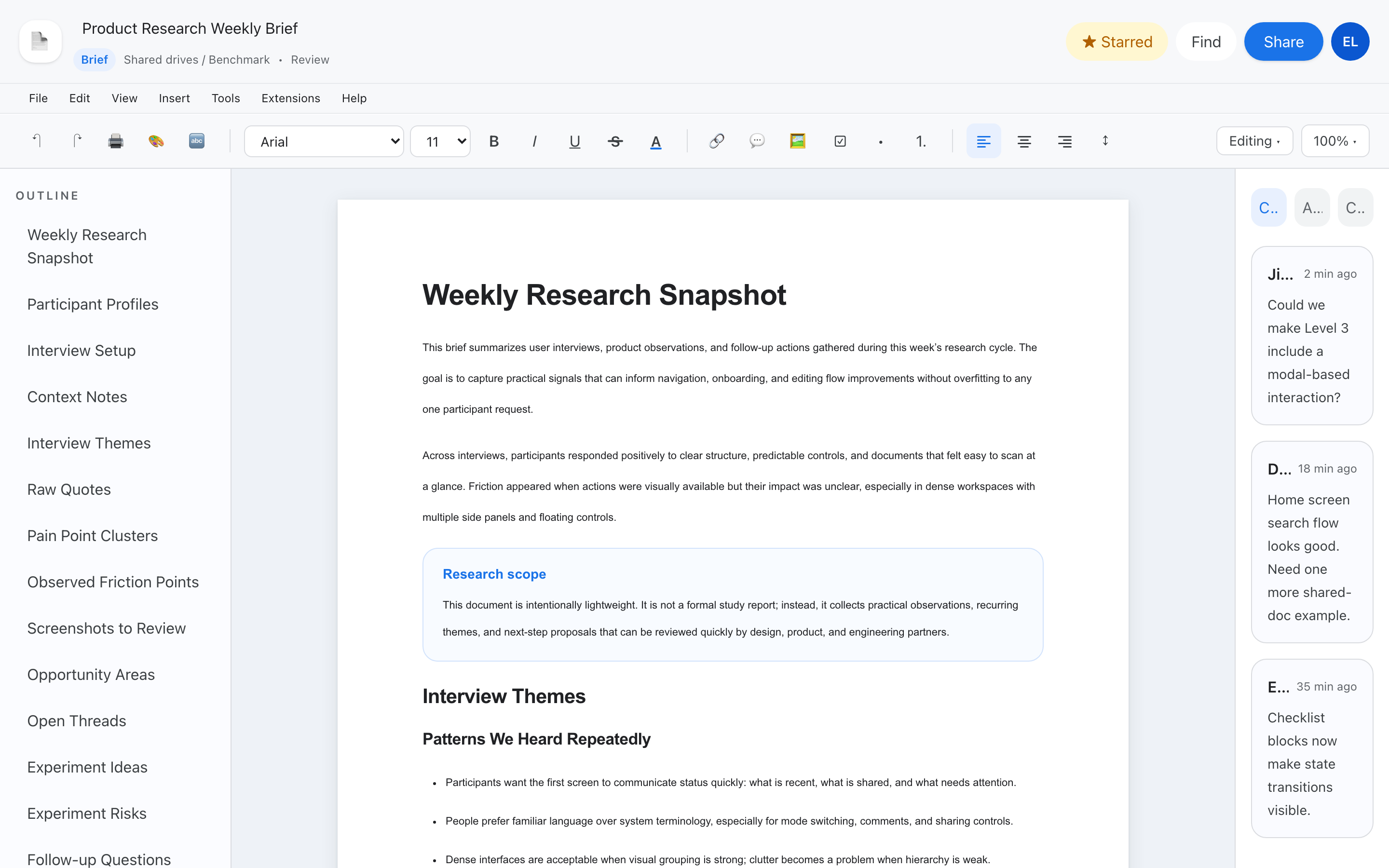}
  \caption{Document Editor (Google Docs)}
\end{subfigure}

\vspace{4pt}

\begin{subfigure}{0.32\linewidth}
  \centering
  \includegraphics[width=\linewidth]{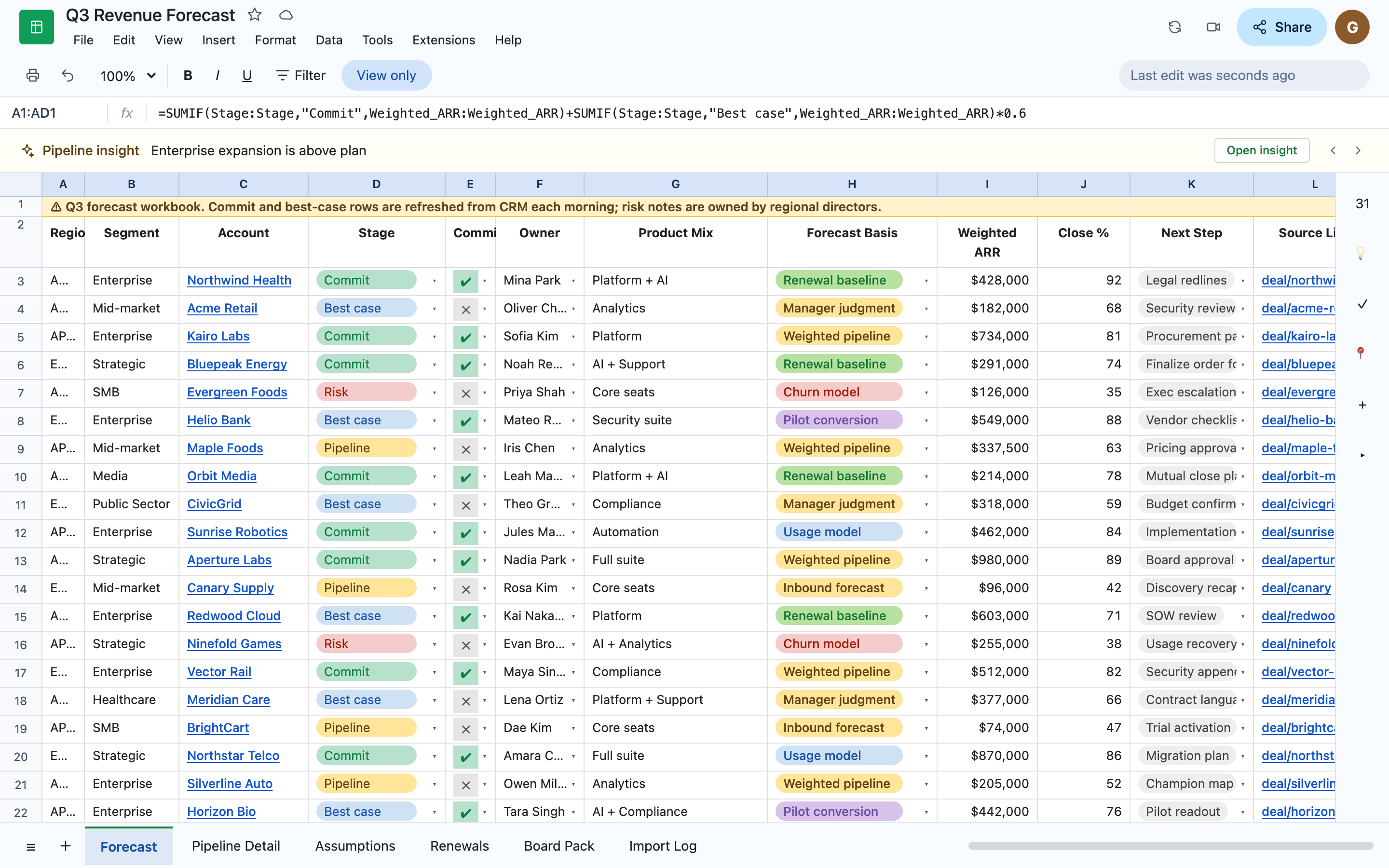}
  \caption{Spreadsheet (Google Sheets)}
\end{subfigure}
\hfill
\begin{subfigure}{0.32\linewidth}
  \centering
  \includegraphics[width=\linewidth]{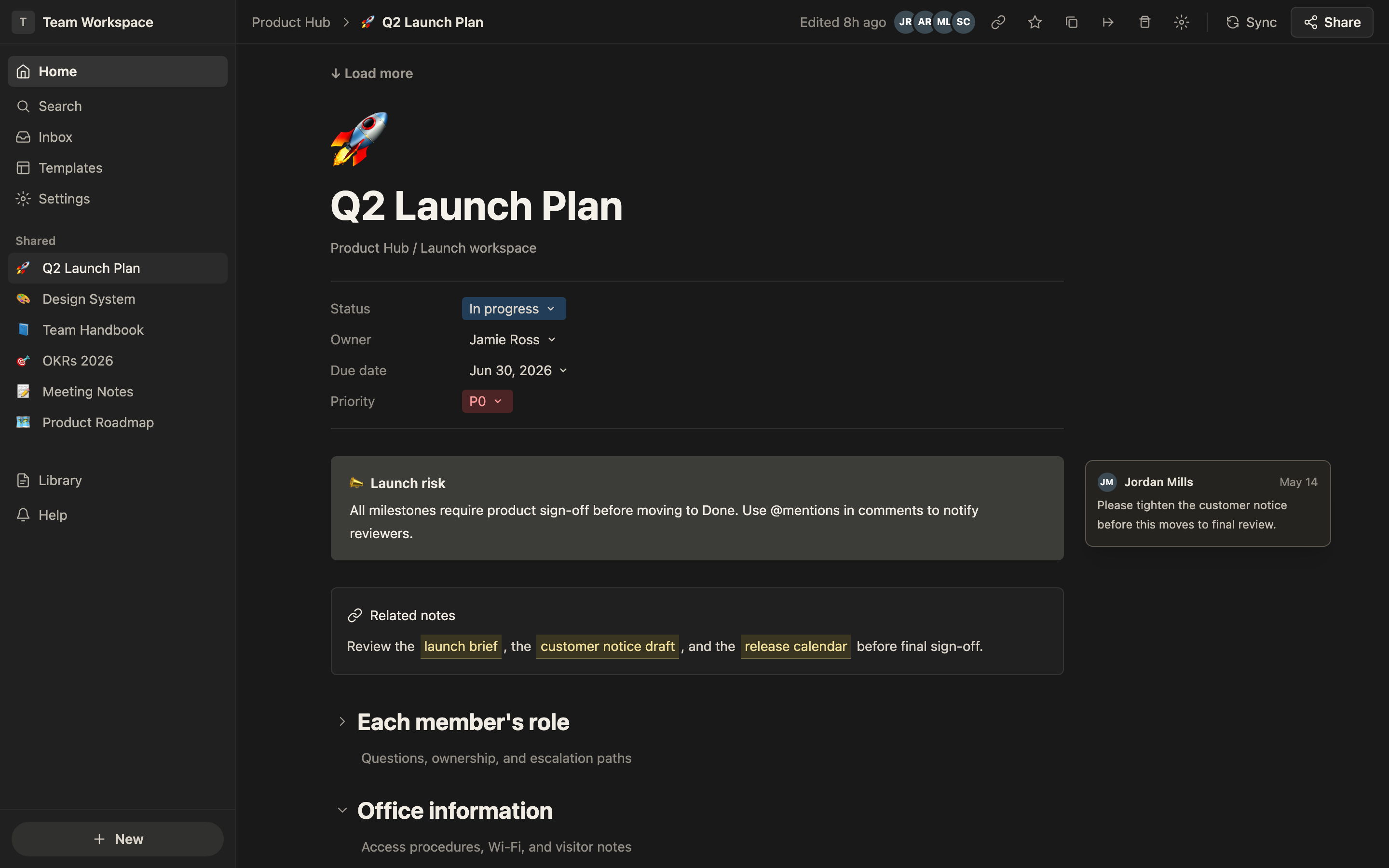}
  \caption{Workspace Pages (Notion)}
\end{subfigure}
\hfill
\begin{subfigure}{0.32\linewidth}
  \centering
  \includegraphics[width=\linewidth]{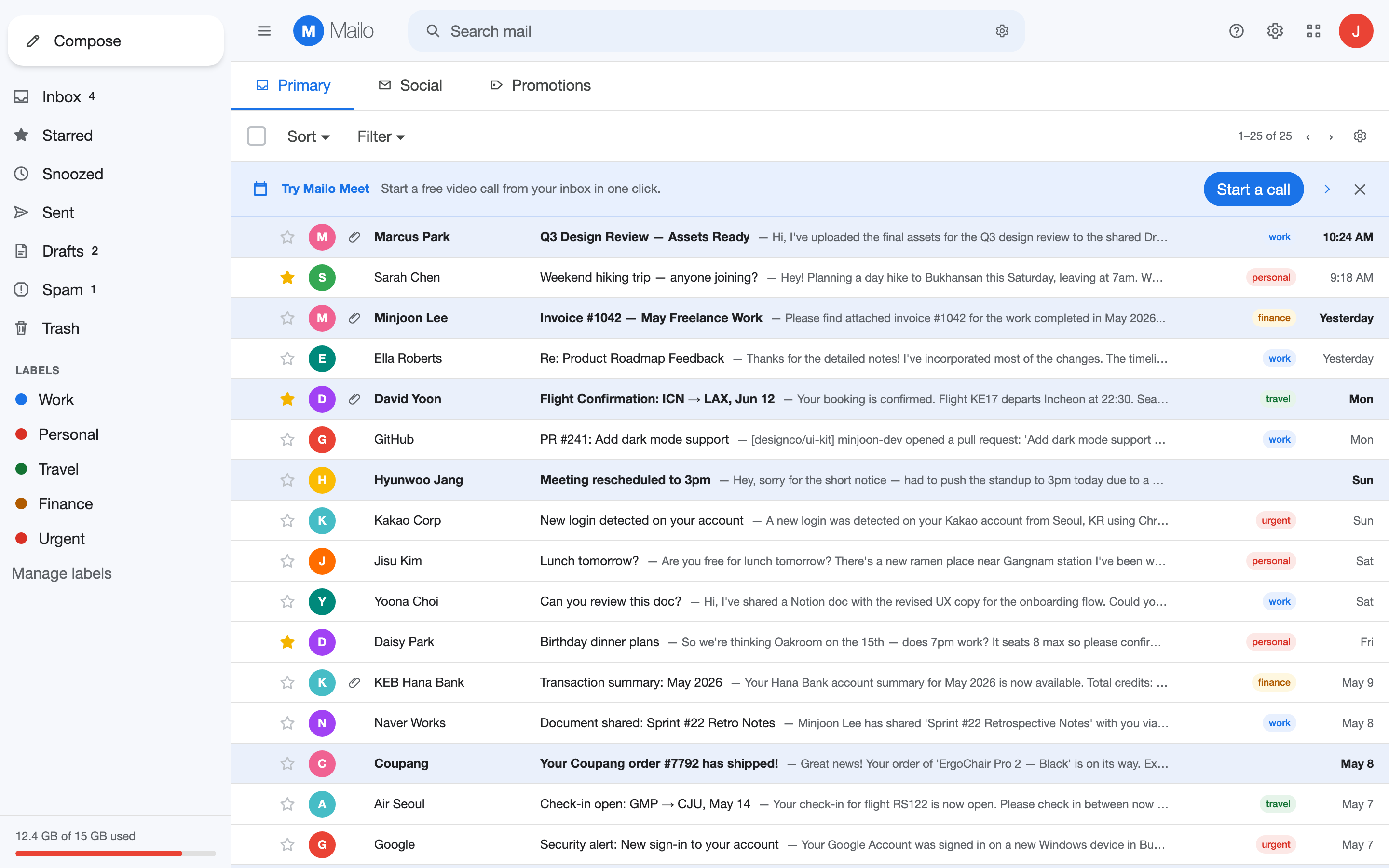}
  \caption{Email Client (Gmail)}
\end{subfigure}
\caption{Representative states from productivity mockup applications.}
\label{fig:mockup_productivity}
\end{figure*}

\begin{figure*}[!htbp]
\centering
\begin{subfigure}{0.49\linewidth}
  \centering
  \includegraphics[width=\linewidth]{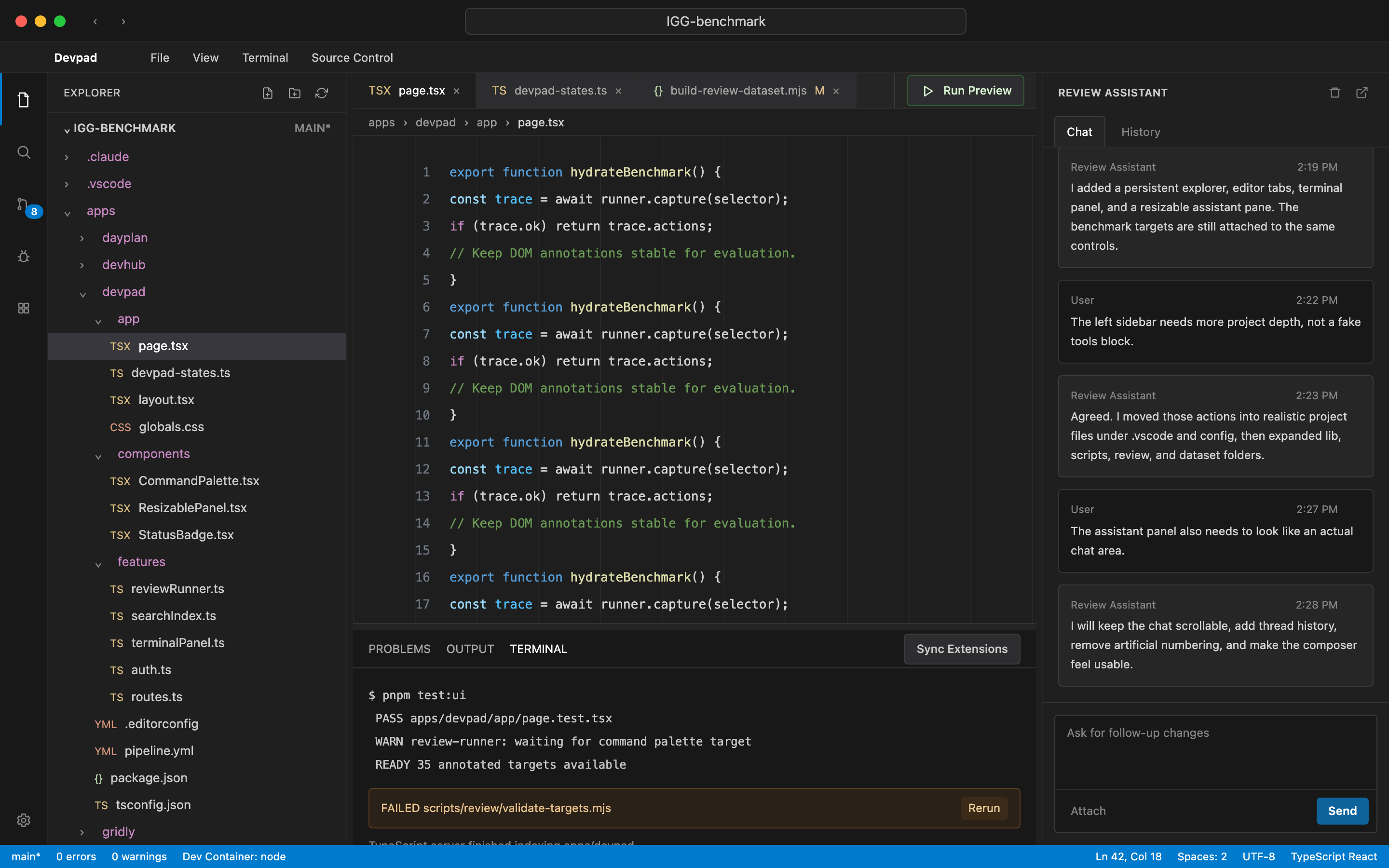}
  \caption{Code Editor (VS Code Web)}
\end{subfigure}
\hfill
\begin{subfigure}{0.49\linewidth}
  \centering
  \includegraphics[width=\linewidth]{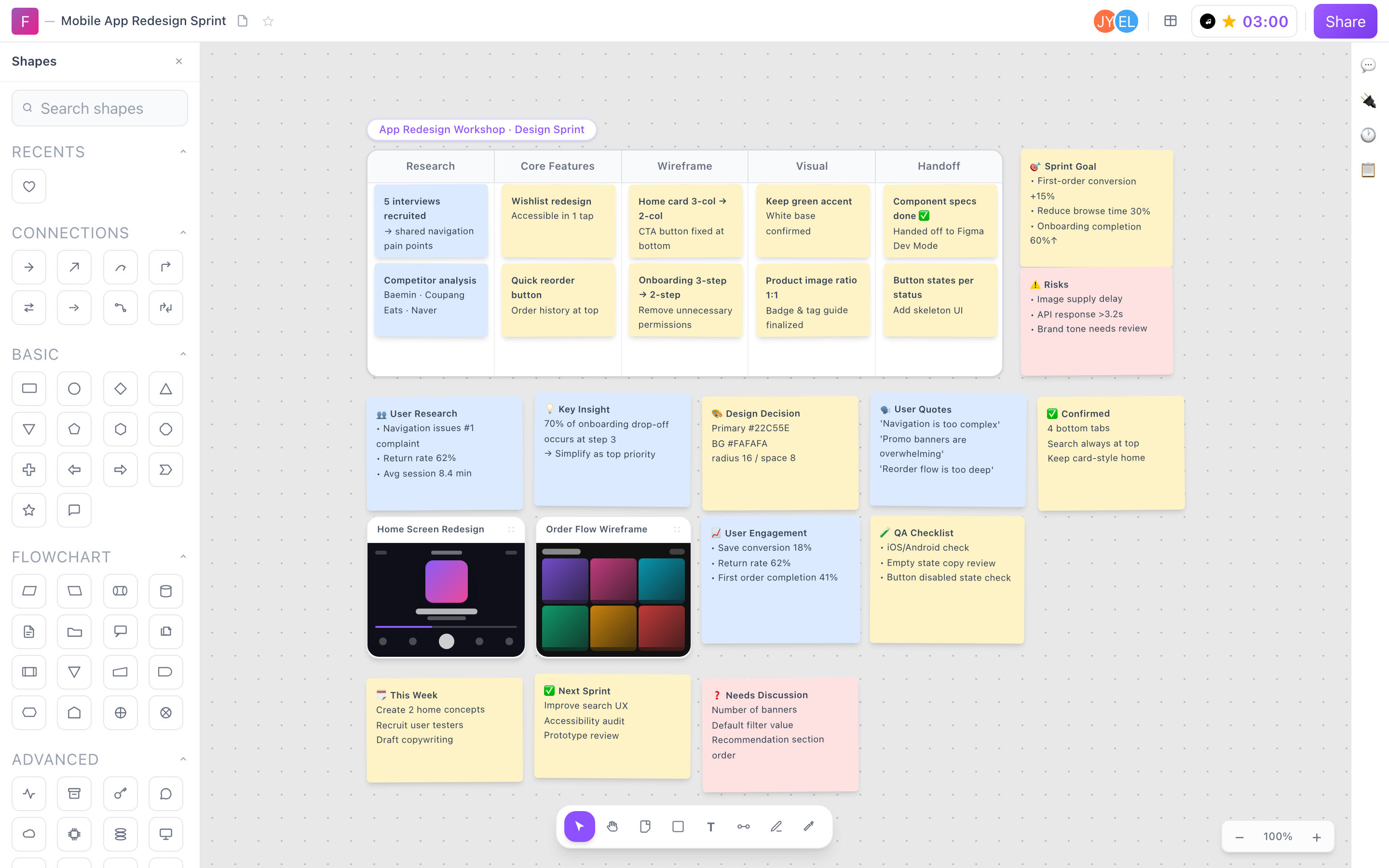}
  \caption{Design Canvas (Figma)}
\end{subfigure}

\vspace{4pt}

\begin{subfigure}{0.49\linewidth}
  \centering
  \includegraphics[width=\linewidth]{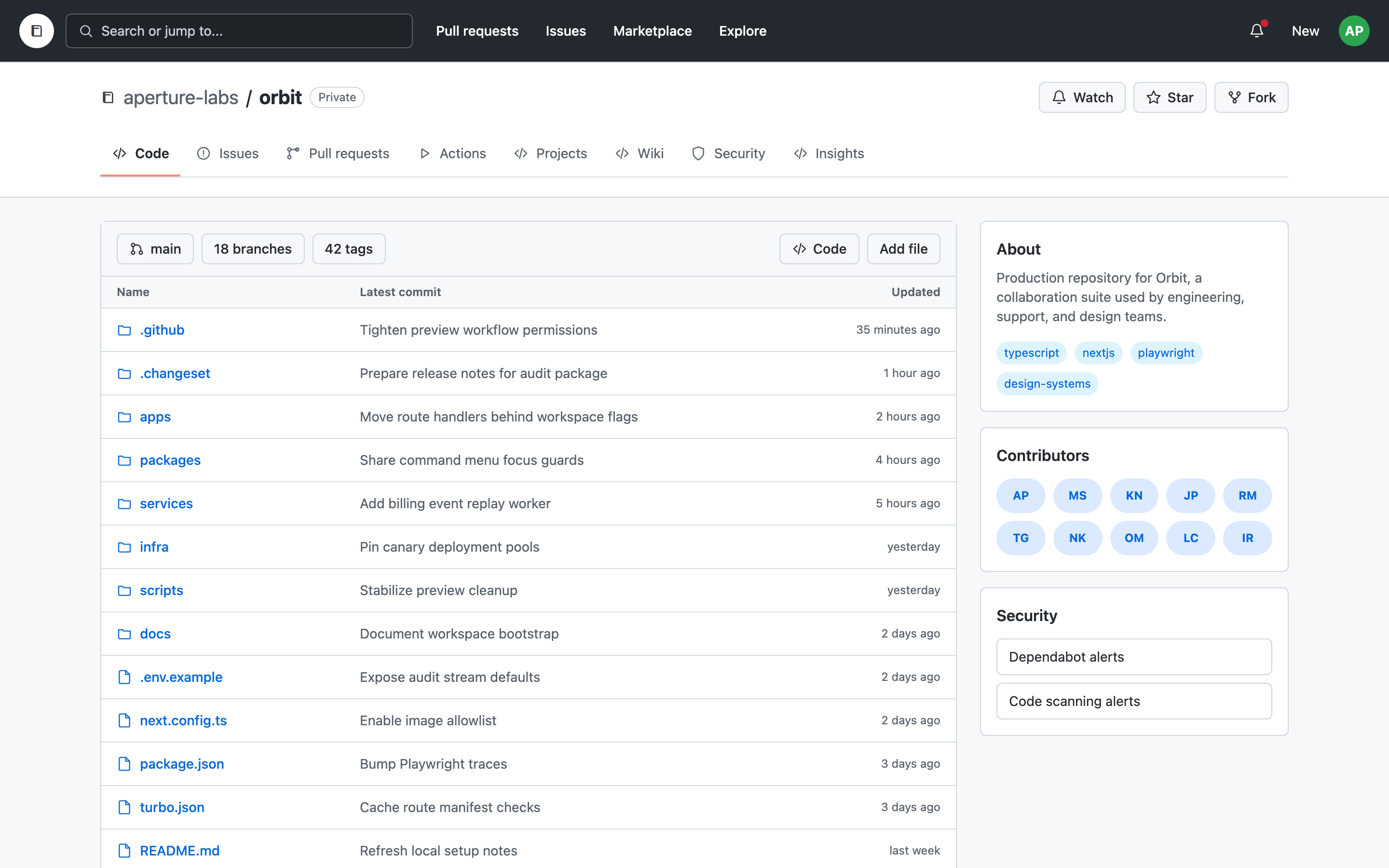}
  \caption{Developer Hub (GitHub)}
\end{subfigure}
\caption{Representative states from development and design mockup applications.}
\label{fig:mockup_dev_design}
\end{figure*}

\begin{figure*}[!htbp]
\centering
\begin{subfigure}{0.49\linewidth}
  \centering
  \includegraphics[width=\linewidth]{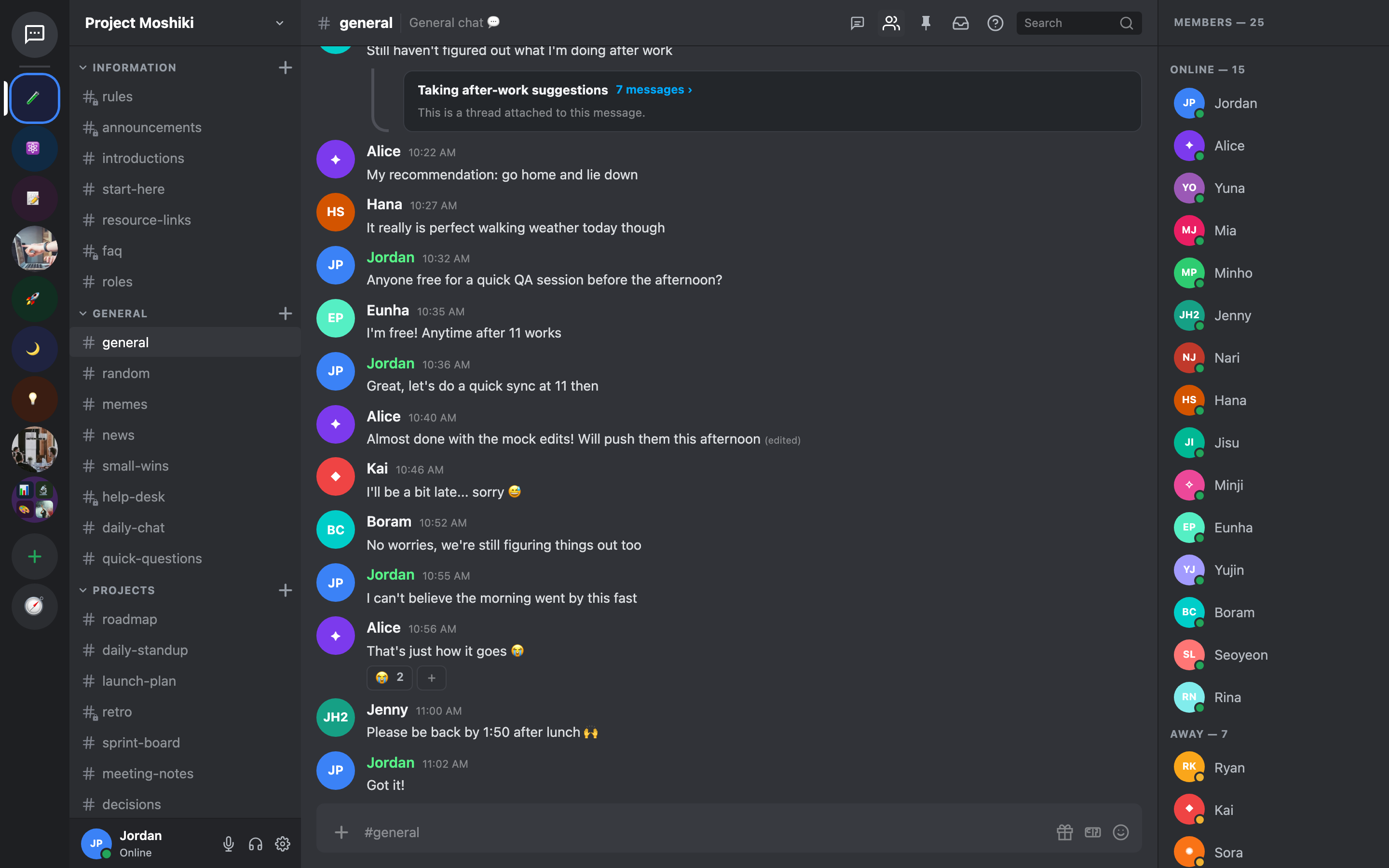}
  \caption{Community Chat (Discord)}
\end{subfigure}
\hfill
\begin{subfigure}{0.49\linewidth}
  \centering
  \includegraphics[width=\linewidth]{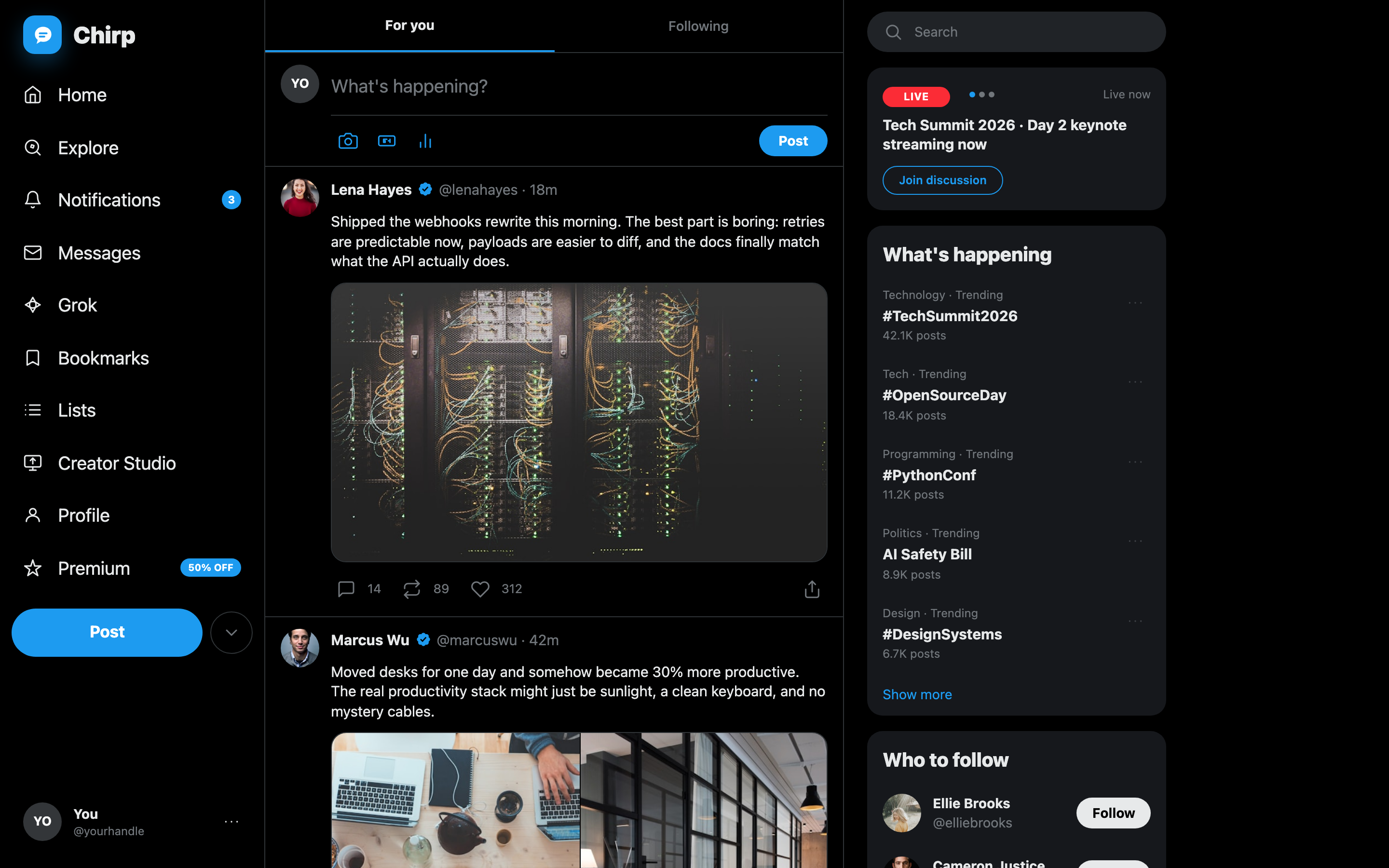}
  \caption{Social Feed (Twitter/X)}
\end{subfigure}
\caption{Representative states from social mockup applications.}
\label{fig:mockup_social}
\end{figure*}

\begin{figure*}[!htbp]
\centering
\begin{subfigure}{0.49\linewidth}
  \centering
  \includegraphics[width=\linewidth]{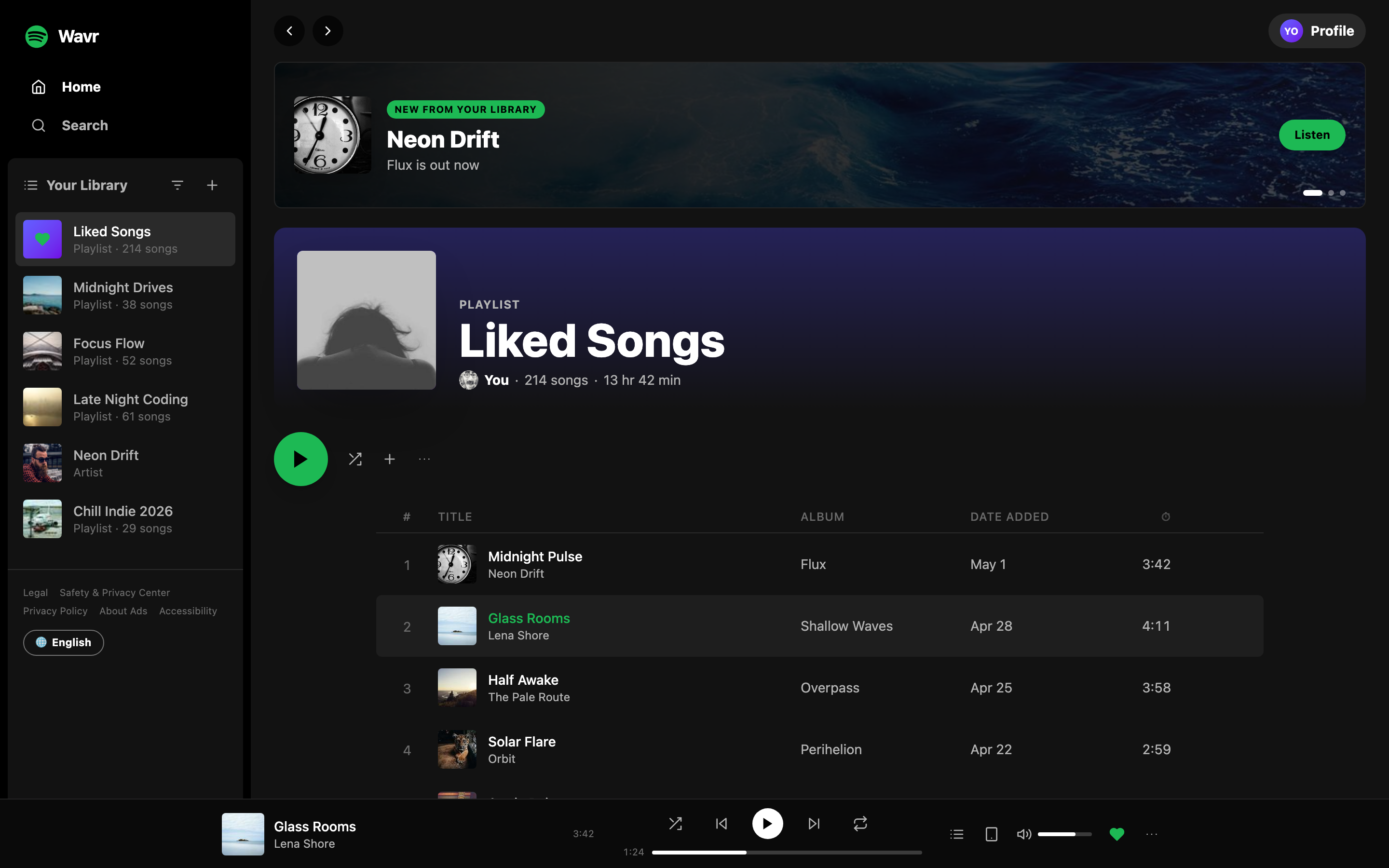}
  \caption{Music Player (Spotify)}
\end{subfigure}
\hfill
\begin{subfigure}{0.49\linewidth}
  \centering
  \includegraphics[width=\linewidth]{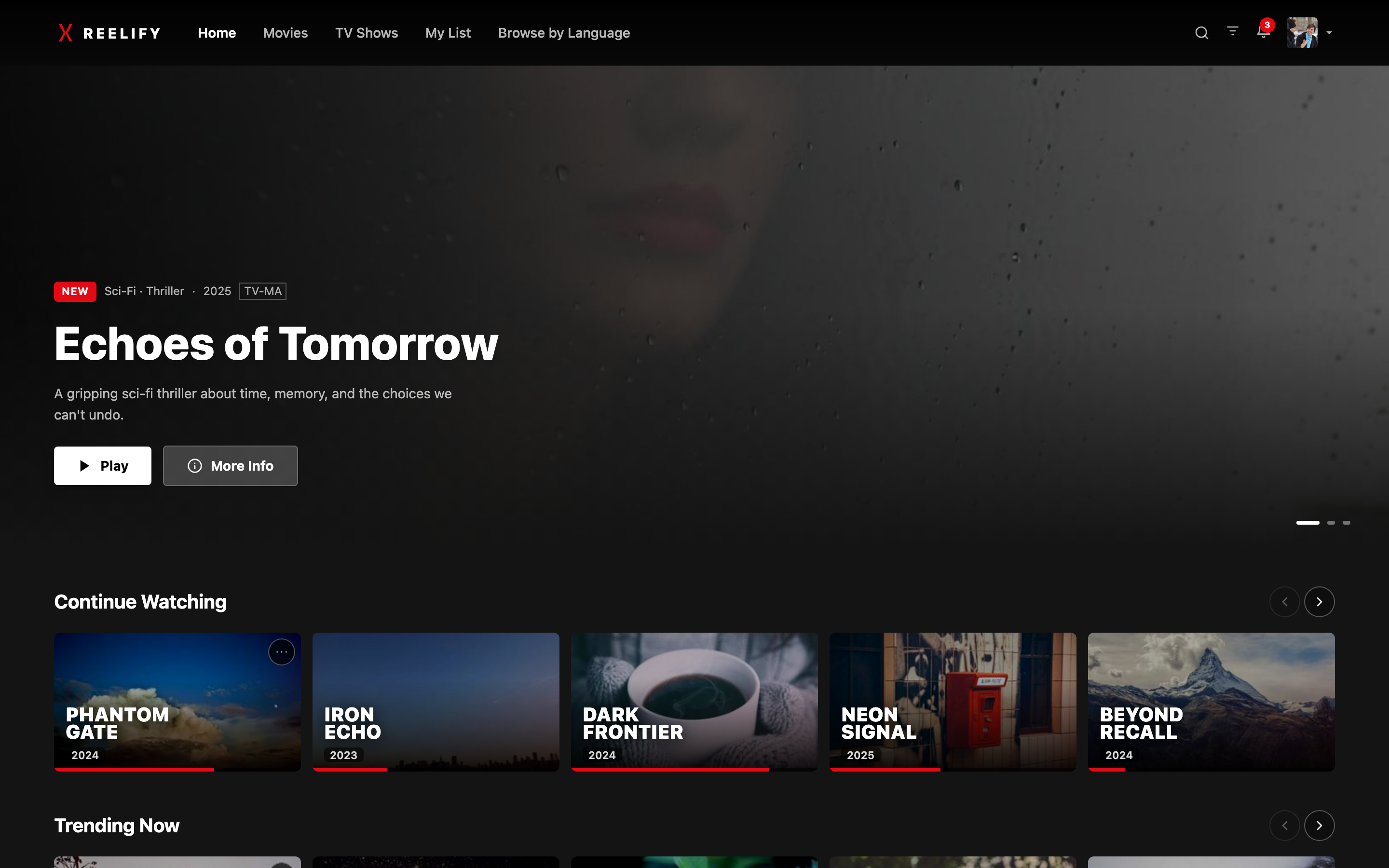}
  \caption{Streaming Service (Netflix)}
\end{subfigure}
\vspace{4pt}

\begin{subfigure}{0.49\linewidth}
  \centering
  \includegraphics[width=\linewidth]{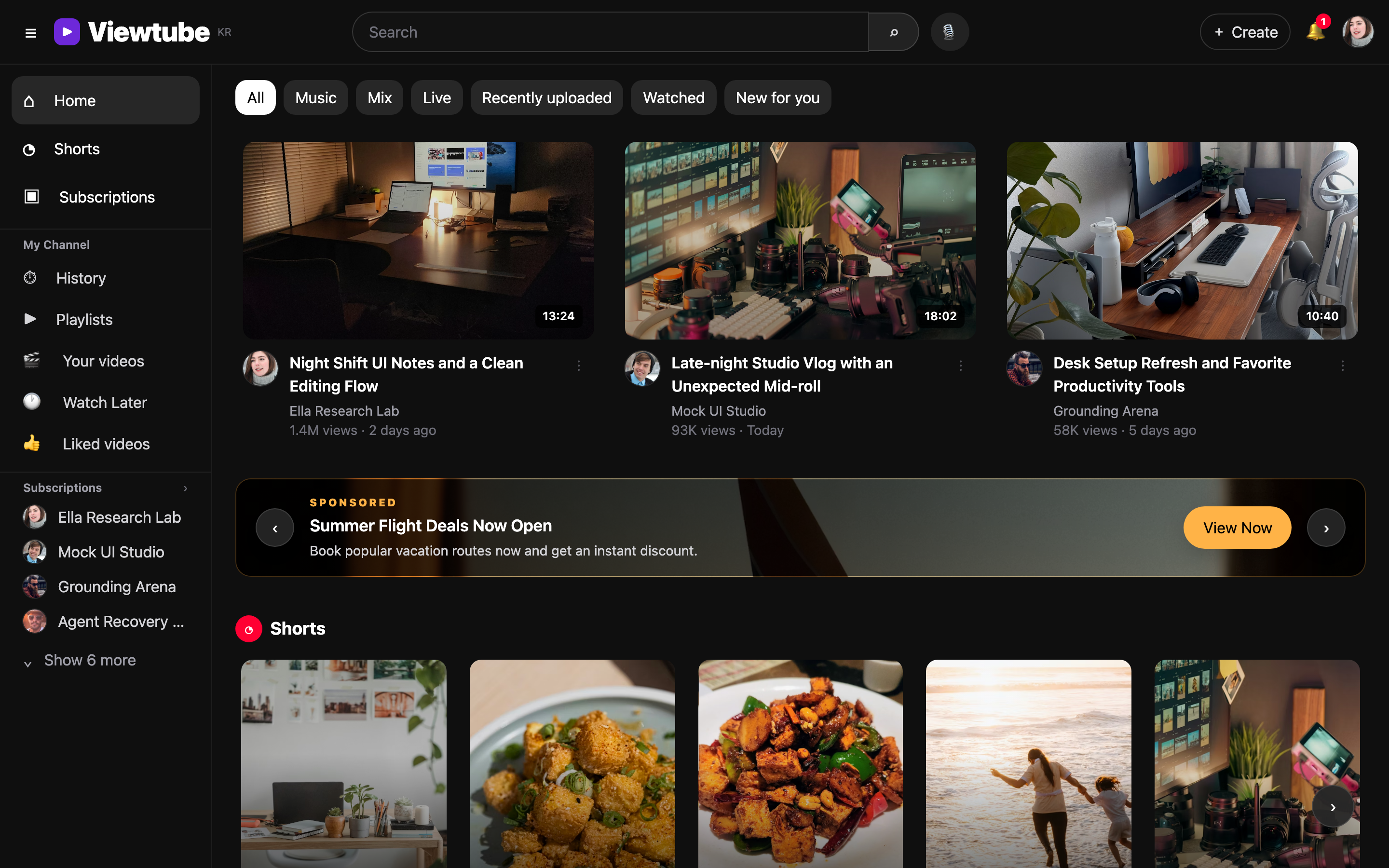}
  \caption{Video Platform (YouTube)}
\end{subfigure}
\caption{Representative states from media mockup applications.}
\label{fig:mockup_media}
\end{figure*}

\begin{figure*}[!htbp]
\centering
\begin{subfigure}{0.32\linewidth}
  \centering
  \includegraphics[width=\linewidth]{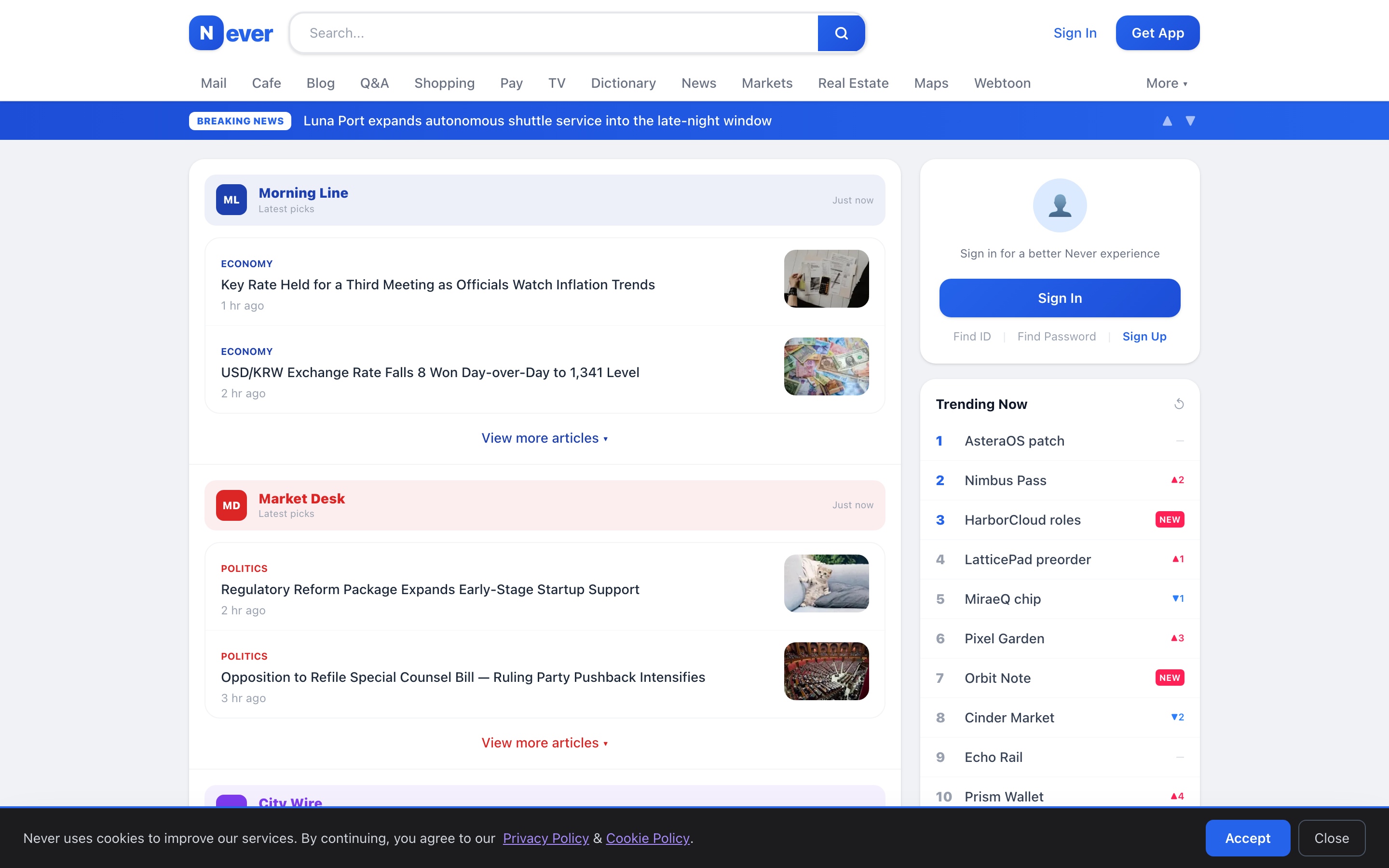}
  \caption{News Portal (Google)}
\end{subfigure}
\hfill
\begin{subfigure}{0.32\linewidth}
  \centering
  \includegraphics[width=\linewidth]{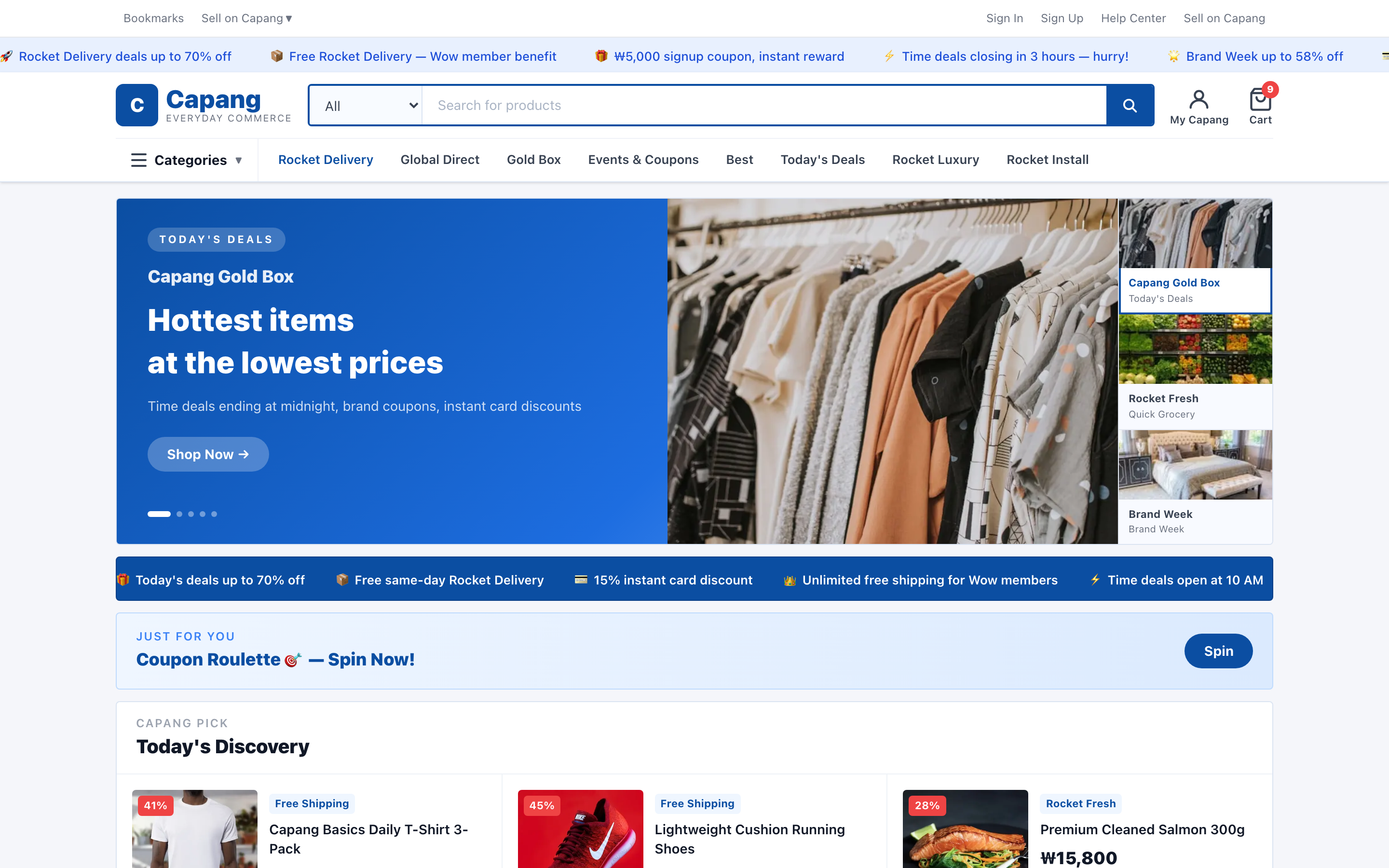}
  \caption{E-commerce (Walmart)}
\end{subfigure}
\hfill
\begin{subfigure}{0.32\linewidth}
  \centering
  \includegraphics[width=\linewidth]{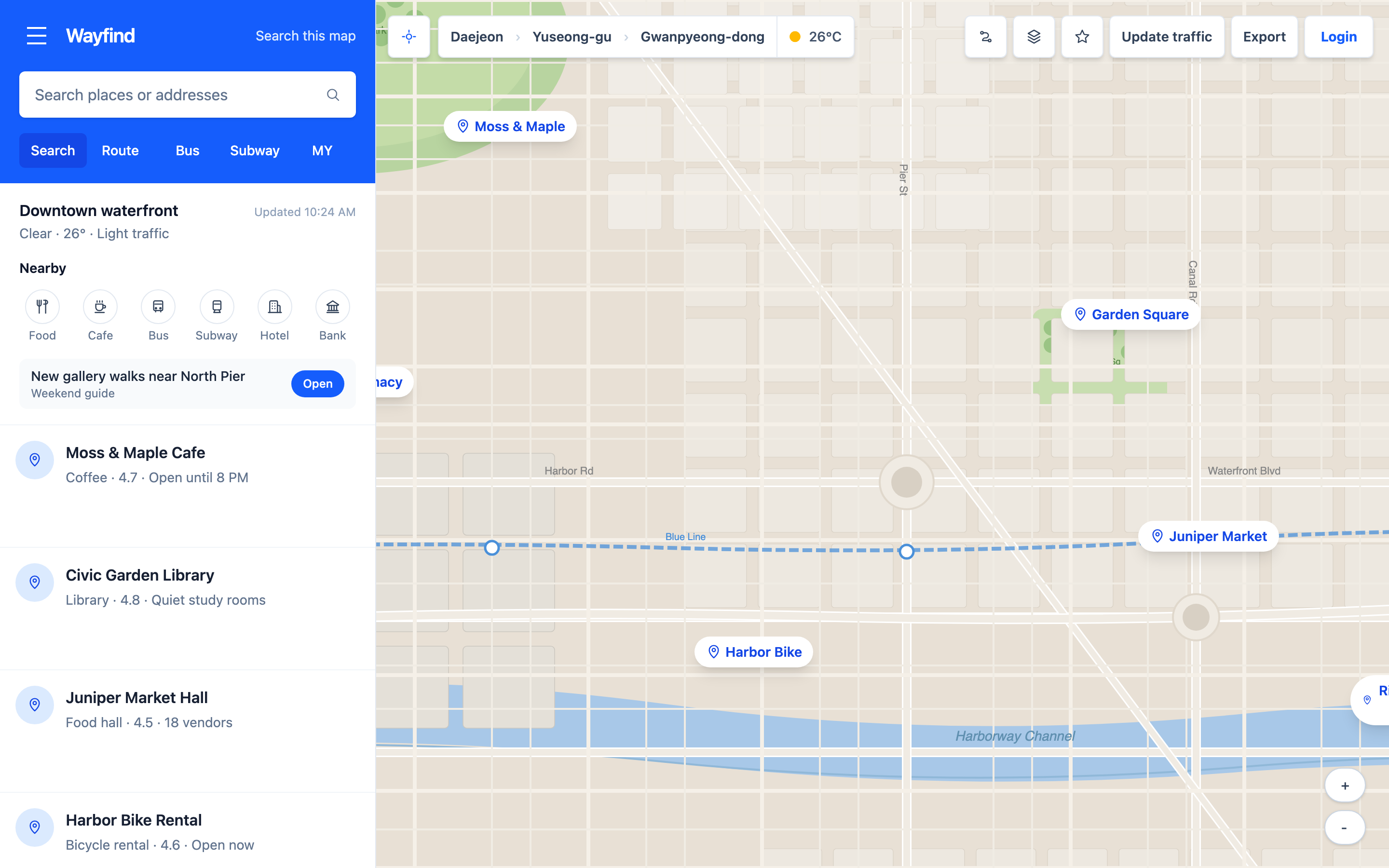}
  \caption{Maps (Google Maps)}
\end{subfigure}
\caption{Representative states from lifestyle mockup applications.}
\label{fig:mockup_lifestyle}
\end{figure*}

IGG comprises 16 controllable applications spanning common GUI domains. Each environment combines deterministic state transitions and stable target identities with interaction patterns modeled on a familiar interface. Table~\ref{tab:app_summary} lists the references, and Figures~\ref{fig:mockup_productivity}-\ref{fig:mockup_lifestyle} show representative states.

\begin{table}[!htbp]
\centering
\small
\setlength{\tabcolsep}{5pt}
\renewcommand{\arraystretch}{1.05}
\caption{The 16 mockup environments and their reference interfaces.}
\label{tab:app_summary}
\begin{tabular}{@{}ll@{}}
\toprule
\textbf{Mockup} & \textbf{Reference} \\
\midrule
Social Feed       & Twitter/X \\
Planner           & Google Calendar \\
Developer Hub     & GitHub \\
Code Editor       & VS Code Web \\
Community Chat    & Discord \\
Document Editor   & Google Docs \\
Design Canvas     & Figma \\
Spreadsheet       & Google Sheets \\
Email Client      & Gmail \\
News Portal       & Google \\
Workspace Pages   & Notion \\
Streaming Service & Netflix \\
E-commerce        & Walmart \\
Music Player      & Spotify \\
Maps              & Google Maps \\
Video Platform    & YouTube \\
\bottomrule
\end{tabular}
\end{table}

\section{Design Choices}
\label{app:design_choices}

\subsection{Viewport robustness}
\label{app:viewport}
The main benchmark fixes the viewport at $1440 \times 900$ so that initial target visibility and constraint type remain constant across models. Responsive layouts can change whether a target is hidden and which interaction reveals it, so the viewport is part of the task specification.

To test whether the learned behavior depends on this layout, we regenerate the 601 tasks at five viewport sizes corresponding to desktop, mobile portrait, tablet, split-screen, and zoomed displays, and keep the 127 tasks that remain valid in every layout (i.e., the target is initially hidden and the reference trace replays successfully). The subset contains 103 training and 24 held-out tasks, so the study measures whether behavior learned at $1440 \times 900$ survives a layout change rather than held-out generalization. On this subset, the trained Qwen3-VL model outperforms its base model by 11.1-17.3 points at every viewport, including mobile portrait (Table~\ref{tab:viewport_robustness}).

\begin{table}[t]
\centering
\small
\caption{Grounding accuracy (\%) at five viewport sizes.}
\label{tab:viewport_robustness}
\setlength{\tabcolsep}{4pt}
\begin{tabular}{lccc}
\toprule
\textbf{Viewport} & \textbf{Base} & \textbf{Trained} & \textbf{Gain} \\
\midrule
$1440\times900$ & 43.3 & 59.8 & +16.5 \\
$390\times844$  & 43.3 & 60.6 & +17.3 \\
$1024\times768$ & 42.5 & 55.9 & +13.4 \\
$720\times900$  & 48.0 & 59.1 & +11.1 \\
$1152\times720$ & 39.4 & 52.8 & +13.4 \\
\bottomrule
\end{tabular}
\end{table}

\section{Additional Results and Analyses}
\label{app:additional_results}
\subsection{Per-environment results}
\label{app:per_app}
Table~\ref{tab:per_app_extended_metrics} breaks the main results down by environment. Open-source models trail GPT-5.5 in every environment. Design Canvas is the hardest environment for open-source models (8.9--13.3\% GA) and Video Platform and Email Client the easiest, whereas GPT-5.5 is weakest on Developer Hub, Community Chat, and Social Feed.

\subsection{Mockup fidelity}
\label{appendix:fidelity}
Each task starts from a resettable state, executes deterministic transitions, and uses a stable target ID for click-time evaluation. The environments implement the visibility-changing interactions measured by IGG, including hidden panels, hover reveals, popups, delayed elements, carousels, and movable occlusions. We evaluate their fidelity to real-interface counterparts using visual similarity, human ratings, and paired task execution.

\paragraph{Visual similarity.}
We measure visual similarity between each mockup and its corresponding real-application reference listed in Table~\ref{tab:app_summary}. We compare each matched mock-real pair against non-matched pairs
formed by pairing the mockup with the other 15 real-application references. We compute feature-space cosine similarities using DINO~\citep{caron2021emerging} and SigLIP~\citep{zhai2023sigmoid}, and perceptual distance using DreamSim~\citep{fu2023dreamsim}. As shown in Table~\ref{tab:mock_visual_similarity}, matched pairs are substantially more similar than non-matched pairs, with higher DINO similarity (0.7724 vs.\ 0.5442), higher SigLIP similarity (0.8094 vs.\ 0.6211), and lower DreamSim distance (0.2529 vs.\ 0.5229). These results indicate that the mockups preserve application-level visual structure beyond generic webpage appearance.

\begin{table}[t]
\centering
\small
\caption{Visual similarity between mockups and real-application references. DINO and SigLIP are cosine similarities, where higher is better. DreamSim is a perceptual distance, where lower is better.}
\label{tab:mock_visual_similarity}
\setlength{\tabcolsep}{3pt}
\renewcommand{\arraystretch}{1.08}
\begin{tabular}{lcccc}
\toprule
\textbf{Pairing} & \textbf{Count} & \textbf{DINO} $\uparrow$ & \textbf{SigLIP} $\uparrow$ & \textbf{DreamSim} $\downarrow$ \\
\midrule
Matched & 16  & 0.7724 & 0.8094 & 0.2529 \\
Non-matched & 240 & 0.5442 & 0.6211 & 0.5229 \\
\bottomrule
\end{tabular}
\end{table}

\paragraph{Human fidelity.}
Participants compare seven environments with their real-interface references and rate visual realism, interaction realism, and task executability. The three mean ratings are 4.50, 4.64, and 4.75 out of 5 (Table~\ref{tab:mock_fidelity}).

\begin{figure*}[!hbtp]
\centering

\begin{subfigure}{0.98\linewidth}
  \centering
  \includegraphics[width=\linewidth]{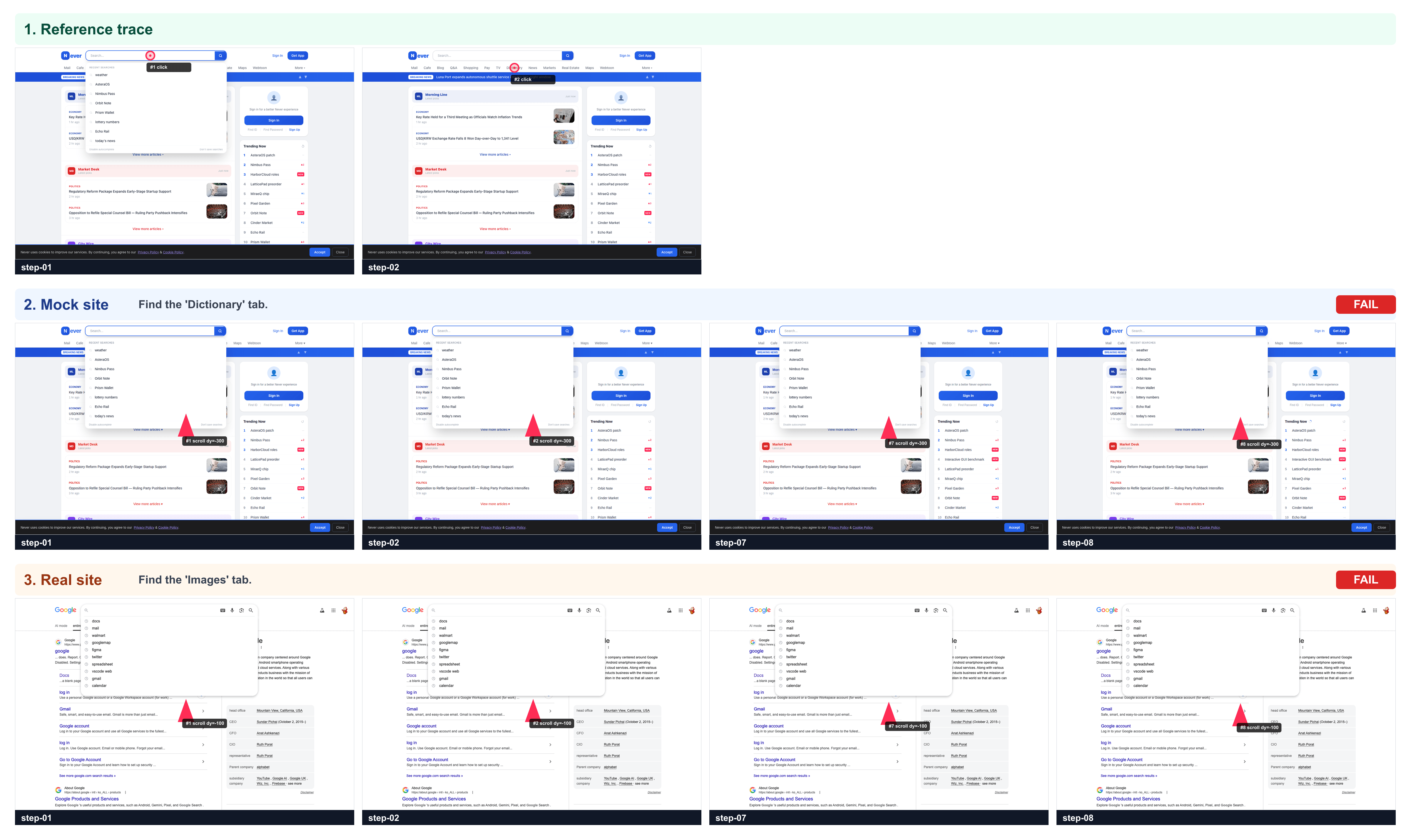}
  \caption{Qwen2.5-VL}
\end{subfigure}
\begin{subfigure}{0.98\linewidth}
  \centering
  \includegraphics[width=\linewidth]{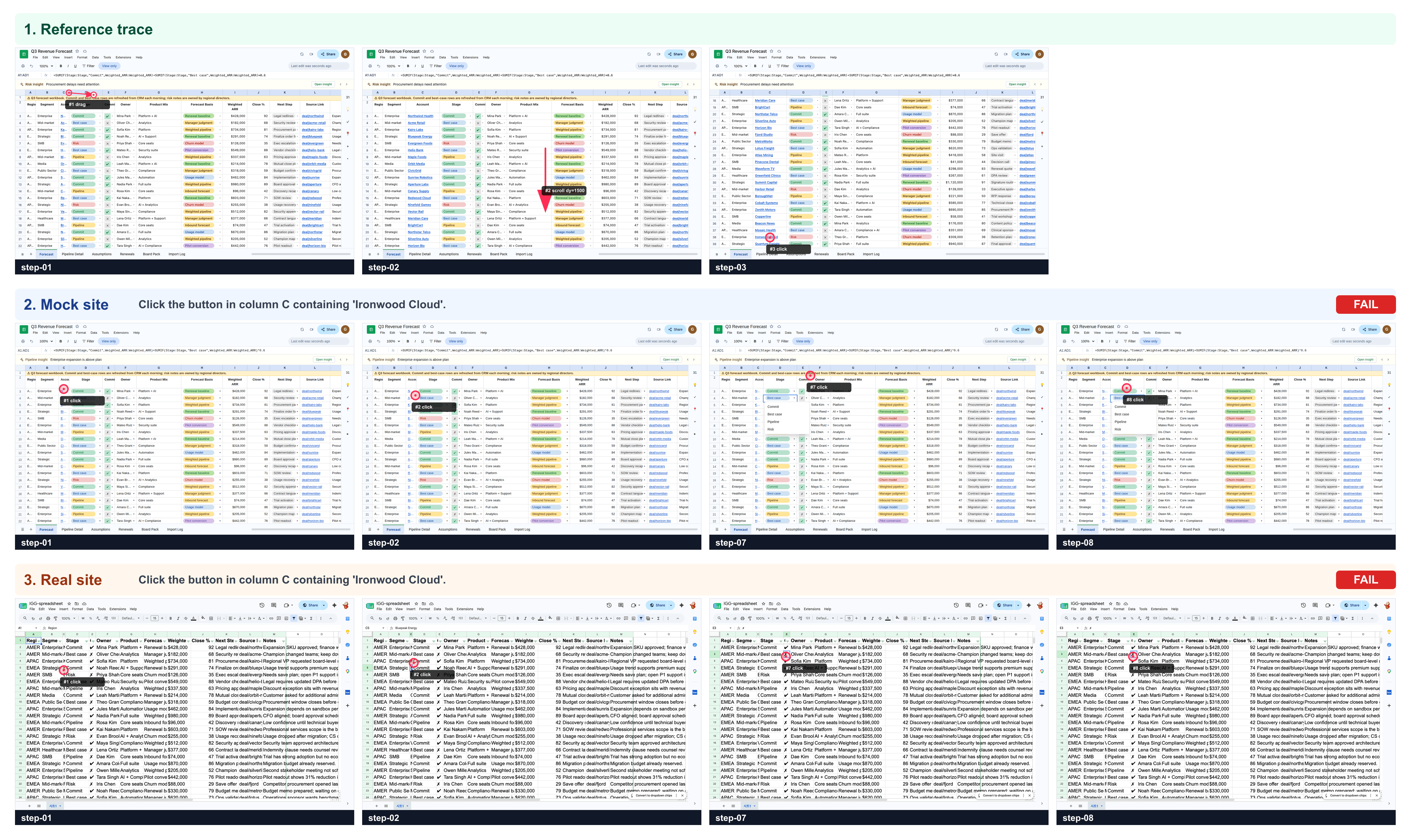}
  \caption{Qwen3-VL}
\end{subfigure}
\caption{Paired examples in which the model fails in both the controlled and real-interface task.}
\label{fig:mock_real_failure_examples}
\end{figure*}

\begin{figure*}[!hbtp]
\centering

\begin{subfigure}{0.98\linewidth}
  \centering
  \includegraphics[width=\linewidth]{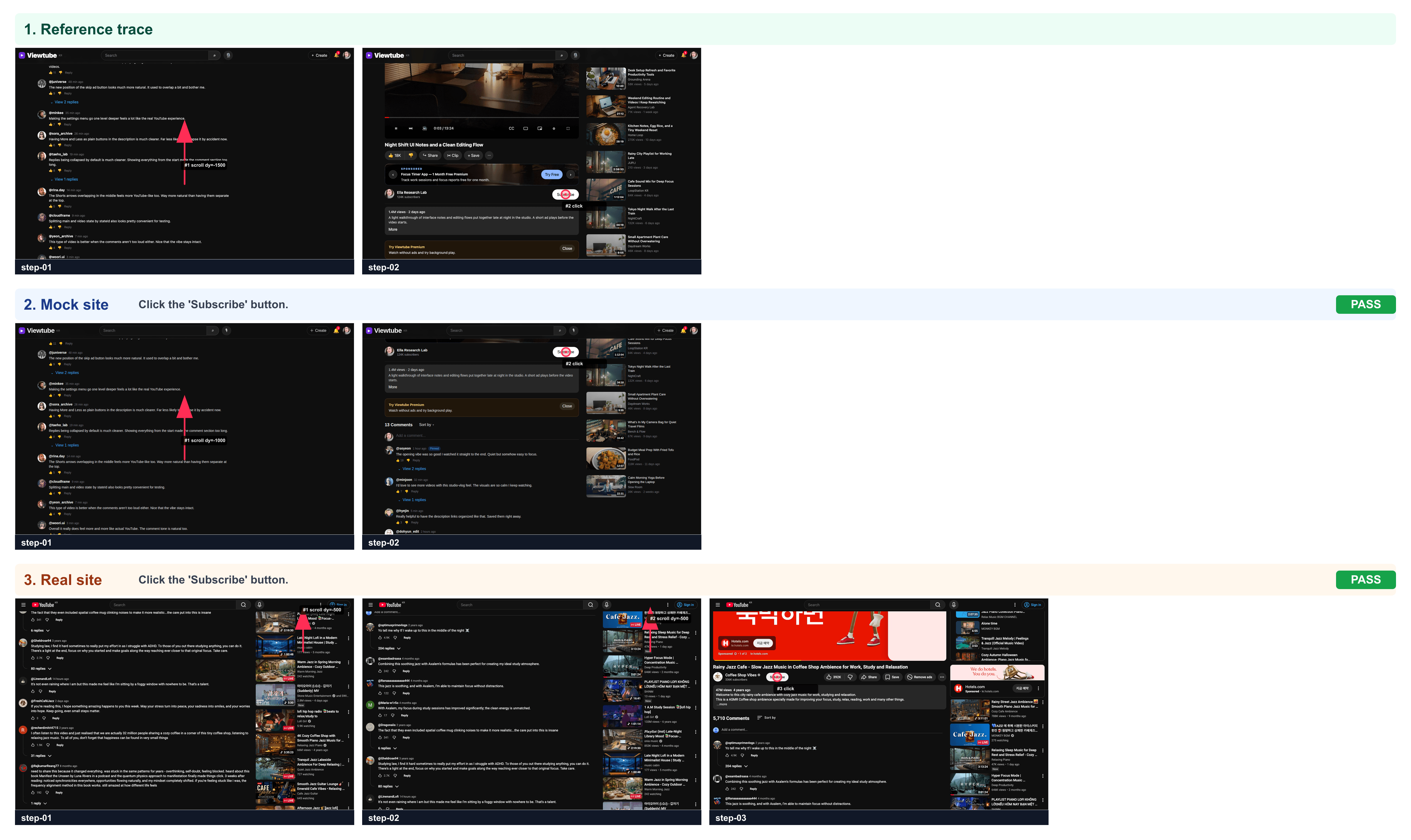}
  \caption{GUI-Owl-1.5}
\end{subfigure}

\vspace{6pt}

\begin{subfigure}{0.98\linewidth}
  \centering
  \includegraphics[width=\linewidth]{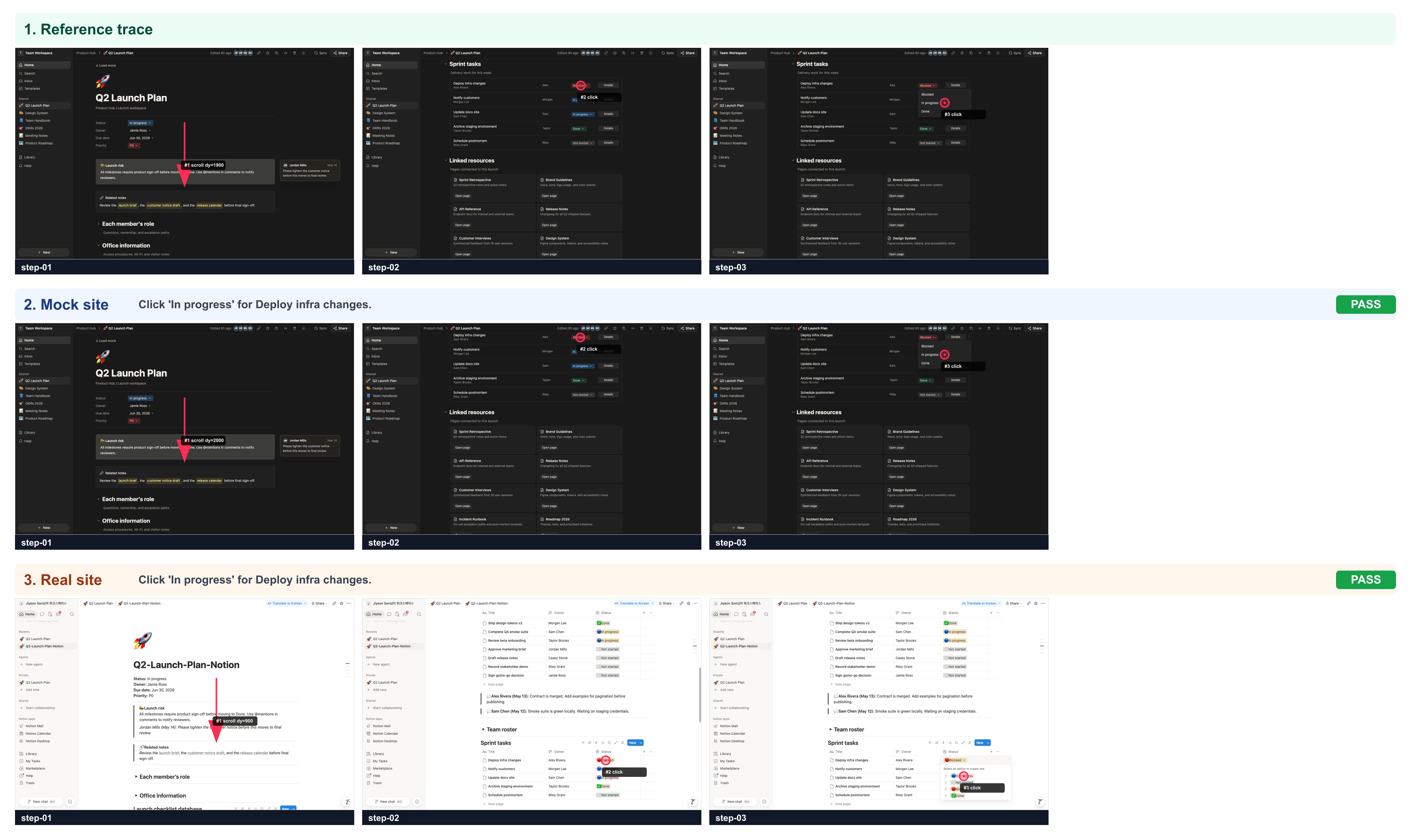}
  \caption{Qwen3-VL SFT+GRPO}
\end{subfigure}
\caption{Paired examples in which the model succeeds in both the controlled and real-interface task.}
\label{fig:mock_real_success_examples}
\end{figure*}

\paragraph{Task-level mock-real fidelity.}
We run four models on 16 paired mock and real-interface tasks. Model ordering is preserved: Qwen3-VL SFT+GRPO leads in both settings with 56.3\% GA, followed by GUI-Owl-1.5, Qwen3-VL, and Qwen2.5-VL (Table~\ref{tab:mock_real_results}). Across the four model families, mock and real GA are strongly consistent (Pearson $r=0.98$; Spearman $\rho=1.00$). The paired set contains one task per application. Figures~\ref{fig:mock_real_failure_examples} and~\ref{fig:mock_real_success_examples} show representative outcomes.

\begin{table}[t]
\centering
\scriptsize
\caption{Task-level fidelity on 16 paired mock-real tasks. Succ. denotes the
number of successful trials out of 16.}
\label{tab:mock_real_results}
\setlength{\tabcolsep}{3.5pt}
\renewcommand{\arraystretch}{1.05}
\begin{tabular}{llcccc}
\toprule
\textbf{Model} & \textbf{Env.} & \textbf{Succ.} &
\textbf{GA} $\uparrow$ & \textbf{AC} $\downarrow$ &
\textbf{RER} $\uparrow$ \\
\midrule
\multirow{2}{*}{Qwen2.5-VL}
& Mock & 3/16 & 18.8 & 7.25 & 0.45 \\
& Real & 3/16 & 18.8 & 7.69 & 0.42 \\
\midrule
\multirow{2}{*}{Qwen3-VL}
& Mock & 4/16 & 25.0 & 7.06 & 0.54 \\
& Real & 5/16 & 31.3 & 7.06 & 0.35 \\
\midrule
\multirow{2}{*}{GUI-Owl-1.5}
& Mock & 6/16 & 37.5 & 6.13 & 0.55 \\
& Real & 6/16 & 37.5 & 6.88 & 0.57 \\
\midrule
\multirow{2}{*}{Qwen3 SFT+GRPO}
& Mock & 9/16 & 56.3 & 5.06 & 0.72 \\
& Real & 9/16 & 56.3 & 5.00 & 0.70 \\
\bottomrule
\end{tabular}
\end{table}

\subsection{Instruction design}
\label{app:instruction_design}

Instructions name the target but not the reveal procedure. Annotators avoid action hints such as ``scroll down,'' ``hover over the icon,'' or ``wait until the button appears,'' and add a natural cue, such as an author, document title, or surrounding content, when several elements share a label. For example, ``Scroll down and click Kai'' becomes ``Find Kai in the direct message list,'' and ``Drag the sticky note away and click QA Checklist'' becomes ``Click the QA Checklist note without deleting any notes.'' Multimodal LLMs propose candidate instructions and reference traces, and annotators correct the wording, repair traces where needed, and replay every task to verify that the target is unique and reachable.

\paragraph{Effect of procedural hints.}
As a diagnostic, we also evaluate instructions that state the intermediate procedure. Unlike the \emph{reveal action given} condition of Section~\ref{sec:interaction_diagnostic}, which names only the action type, these hints describe the procedure. With such hints, Qwen3-VL rises from 30.8\% to 56.9\% GA over all 601 tasks, but hints do not solve precise manipulation: hinted Qwen3-VL solves 0 of 8 held-out movable-occlusion tasks, whereas SFT+GRPO solves 5 of 8 without any test-time hint. The main benchmark therefore retains goal-only instructions, and the training gains reflect an executable interaction prior rather than instruction following.

\section{Evaluation Protocol}
\label{app:prompt_protocol}
The open-source baselines are \texttt{Qwen/Qwen2.5-VL-7B-Instruct}, \texttt{Qwen/Qwen3-VL-8B-Instruct}, and \texttt{mPLUG/GUI-Owl-1.5-8B-Instruct}, and the commercial baseline is \texttt{gpt-5.5}. Each model is evaluated once per task with greedy decoding from a reset environment.

At each turn, the agent receives the system prompt, task instruction, screenshot, step index, coordinate space, and action history. It returns one JSON action. Figure~\ref{fig:igg_prompt} shows the prompt, and the shared action schema is:
\begin{lstlisting}[
    basicstyle=\footnotesize\ttfamily,
    backgroundcolor=\color{gray!10},
    frame=single,
    framerule=0pt,
    breaklines=true,
    showstringspaces=false,
    columns=fullflexible
]
{"type": "scroll", "x": 720, "y": 480, "dy": 50}
{"type": "move", "x": 512, "y": 340}
{"type": "drag", "startX": 10, "startY": 20, "endX": 80, "endY": 100}
{"type": "wait", "ms": 1500}
{"type": "click", "x": 512, "y": 340}
\end{lstlisting}
A \texttt{click} succeeds only when the clicked element ID matches the annotated target ID. Otherwise the environment applies any defined state transition and returns the next screenshot. Navigation outside the task environment is disabled, and the episode ends on success or budget exhaustion. The \texttt{drag} schema (\texttt{startX}, \texttt{startY}, \texttt{endX}, \texttt{endY}) is identical in the prompt, the evaluator, and Table~\ref{tab:action_space}.

\begin{figure*}[t]
\centering
\begin{tcolorbox}[
colback=gray!8,
colframe=gray!65,
boxrule=0.8pt,
arc=4pt,
left=6pt,
right=6pt,
top=6pt,
bottom=6pt,
width=0.96\textwidth
]
\scriptsize
\ttfamily
You are an Interactive GUI Grounding agent.

\vspace{0.4em}
Your ONLY goal is to locate the single UI element described in the instruction and click on it. The element may NOT be visible in the current screenshot.
In that case, take the minimum interaction needed to reveal it (scroll, hover, expand, wait, drag away an occluder, etc.) and then click the element.

\vspace{0.4em}
This is a grounding task, not a navigation task. Do NOT try to complete any higher-level workflow; just find the target and click it.

\hspace*{1em}...

\vspace{0.4em}
Action space (emit exactly ONE action per turn):\\
\hspace*{1em}- Scroll a region: \{"type": "scroll", "x": <int>, "y": <int>, "dy": <int>\}\\
\hspace*{2em}dy > 0 scrolls down, dy < 0 scrolls up. Use (x, y) to pick which\\
\hspace*{2em}scroll container gets the event (e.g. a left sidebar vs the main feed).\\
\hspace*{1em}- Move the cursor: \{"type": "move", "x": <int>, "y": <int>\}\\
\hspace*{2em}Triggers hover effects, tooltips, and hover-reveal menus.\\
\hspace*{1em}- Drag: \{"type": "drag", "startX": <int>, "startY": <int>, "endX": <int>, "endY": <int>\}\\
\hspace*{2em}Use to nudge an occluding popup aside without fully closing it, or to move sliders/panels.\\
\hspace*{1em}- Wait: \{"type": "wait", "ms": <int>\}\\
\hspace*{2em}Use when the target is gated by a timer, async load, ad countdown, auto-rotating carousel, etc.\\
\hspace*{1em}- Click: \{"type": "click", "x": <int>, "y": <int>\}\\
\hspace*{2em}Grounding trigger. If (x, y) lands on the target element, the task ends immediately with success.\\
\hspace*{2em}If it lands on anything else, the task continues and the screen state may change.

\vspace{0.4em}
Strategy:\\
\hspace*{1em}1. Read the instruction. Identify a unique visual description of the target.\\
\hspace*{1em}2. Scan the screenshot. If the target is visible, click its center.\\
\hspace*{1em}3. If not visible, reason about why: is it offscreen, in a collapsed panel,\\
\hspace*{2em}hidden behind a popup, revealed only on hover, or not loaded yet?\\
\hspace*{1em}4. Pick the single cheapest action that could make it visible or confirm it.\\
\hspace*{1em}5. Avoid exploratory clicks on random UI elements; each click is evaluated.

\vspace{0.4em}
Output format (STRICT):\\
\hspace*{1em}Reply with exactly ONE line containing a single JSON object for the next action.\\
\hspace*{1em}No prose, no markdown, no code fences, no explanation.\\
\hspace*{1em}ALL coordinates are SEPARATE integer keys, NEVER a list and NEVER a nested object.\\
\hspace*{1em}...
\end{tcolorbox}
\caption{Excerpt of the system prompt used for IGG model evaluation. Ellipses indicate omitted formatting examples.}
\label{fig:igg_prompt}
\end{figure*}

\section{Dataset Examples}
\subsection{Task specification format}
\label{app:task_spec}
Each task record contains a goal-only instruction, a target ID, an initial state, and one successful reference trace (Listing~\ref{lst:task_spec}). The evaluator keeps the target ID hidden from the model and uses the trace only to compute interaction-efficiency metrics. Reference traces may refer to elements by stable selectors, which the replay tool resolves to the center of the element's current bounding box, whereas models must emit screenshot coordinates.

\begin{lstlisting}[
    caption={Task specification example. The task targets ``Kai'' in the direct message list (Community Chat application, \ctype{1}{1}: Scroll-based Exploration). The ground-truth trajectory consists of a scroll action followed by a click on the revealed target.},
    label={lst:task_spec},
    basicstyle=\footnotesize\ttfamily,
    backgroundcolor=\color{gray!10},
    frame=single,
    framerule=0pt,
    breaklines=true,
    showstringspaces=false,
    columns=fullflexible
]
{
  "id": "messaging-01-find-kai-direct-message-list",
  "sub_type": "1-1",
  "instruction": "Find `Kai' in the direct message list.",
  "target_id": "messaging.dm.kai",
  "initial_state_id": "1",
  "ground-truth trajectory": [
    {"action": "SCROLL", "dy": 1000},
    {"action": "CLICK",  "selector": "messaging.dm.kai"}
  ]
}
\end{lstlisting}
\subsection{Sub-type examples}
\label{app:dataset_examples}
\begin{figure*}[!htbp]
\centering
\begin{subfigure}[t]{0.49\linewidth}
  \centering
  \includegraphics[width=\linewidth]{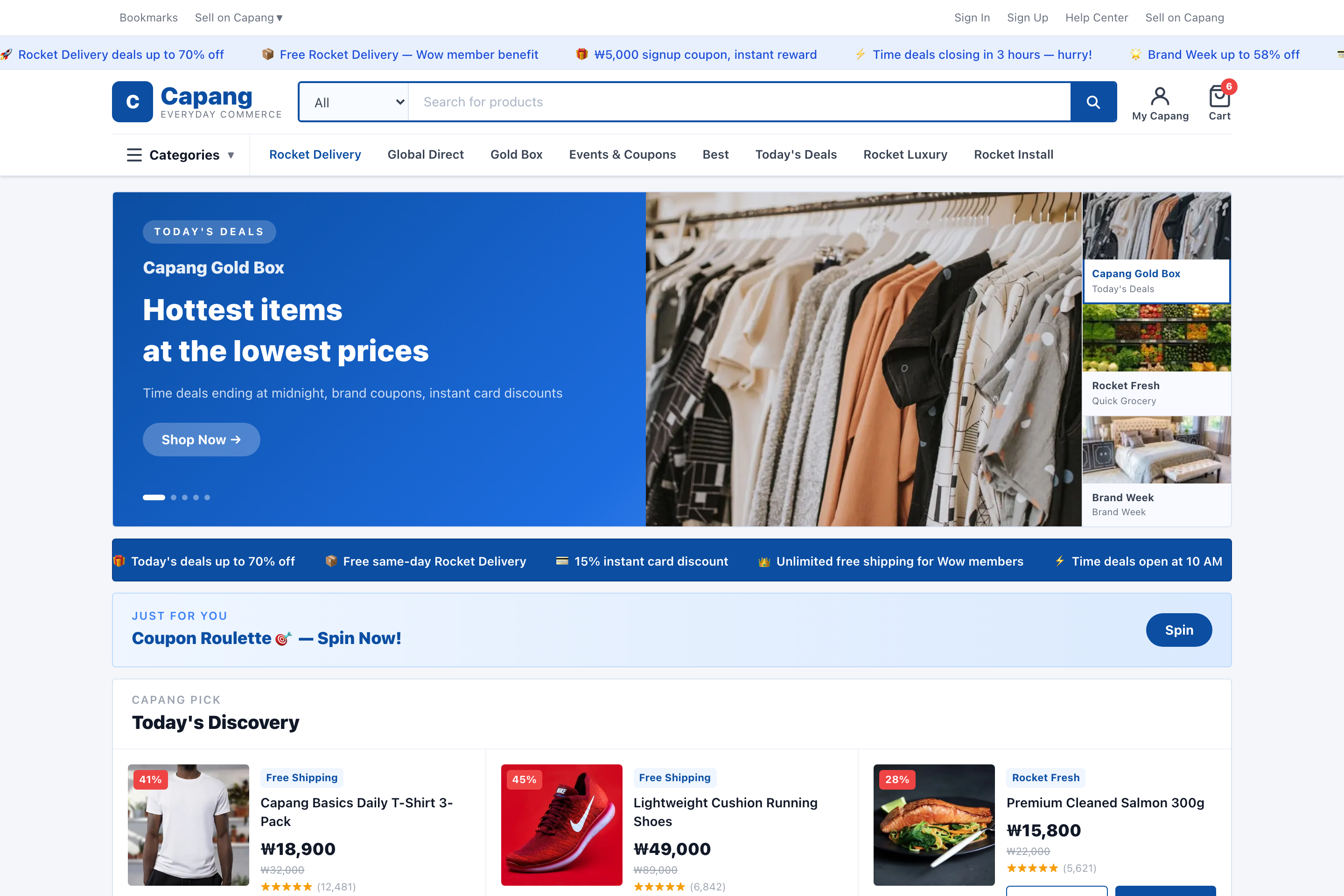}
  \caption{Initial state}
\end{subfigure}
\hfill
\begin{subfigure}[t]{0.49\linewidth}
  \centering
  \includegraphics[width=\linewidth]{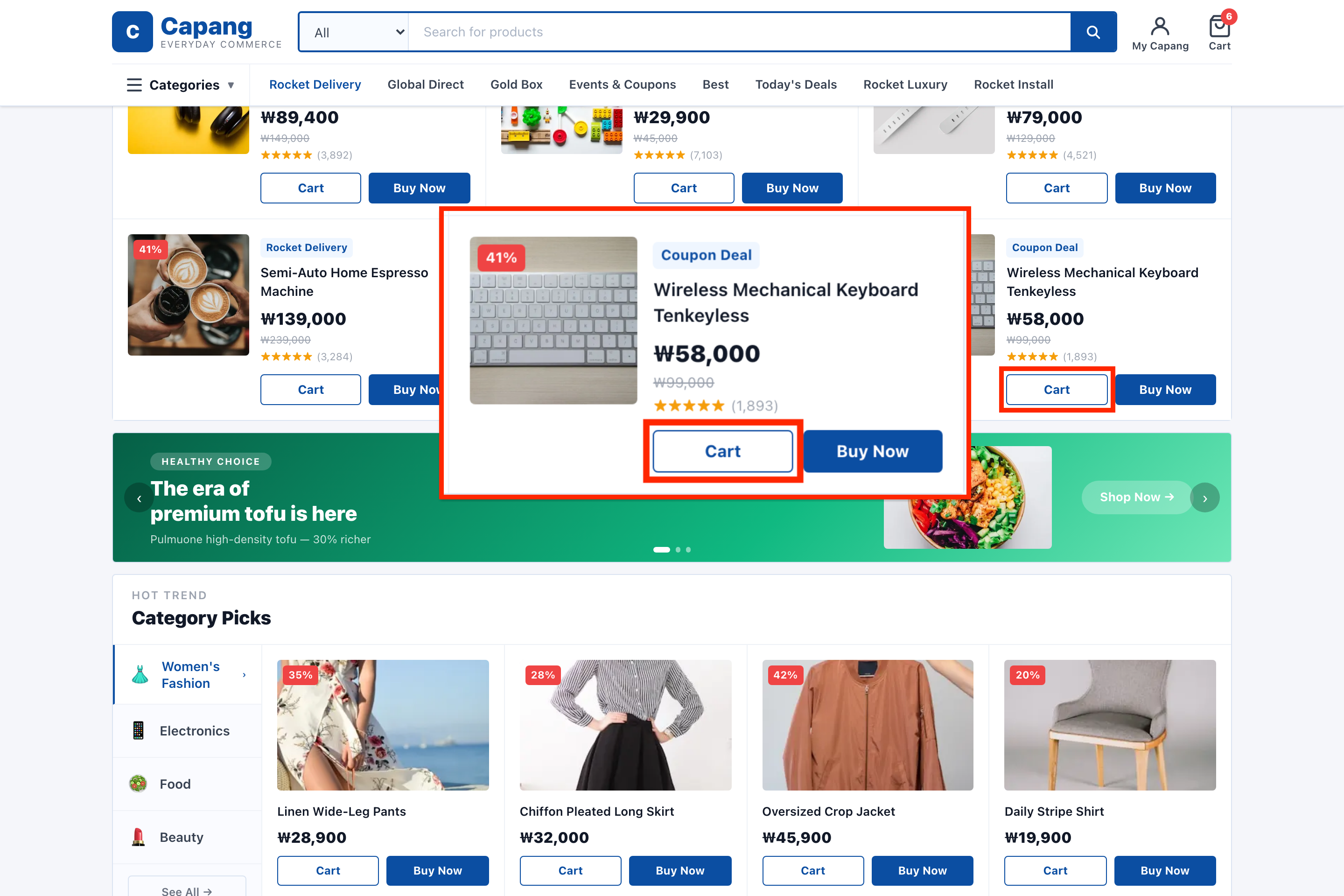}
  \caption{Final state at click time}
\end{subfigure}
\caption{\textbf{\ctypeb{1}{1}: Scroll-based Exploration} (E-commerce). \textit{``Click `Cart' for `Wireless Mechanical Keyboard'.''}}
\label{fig:subtype_l1_1}
\end{figure*}

\begin{figure*}[!htbp]
\centering
\begin{subfigure}[t]{0.49\linewidth}
  \centering
  \includegraphics[width=\linewidth]{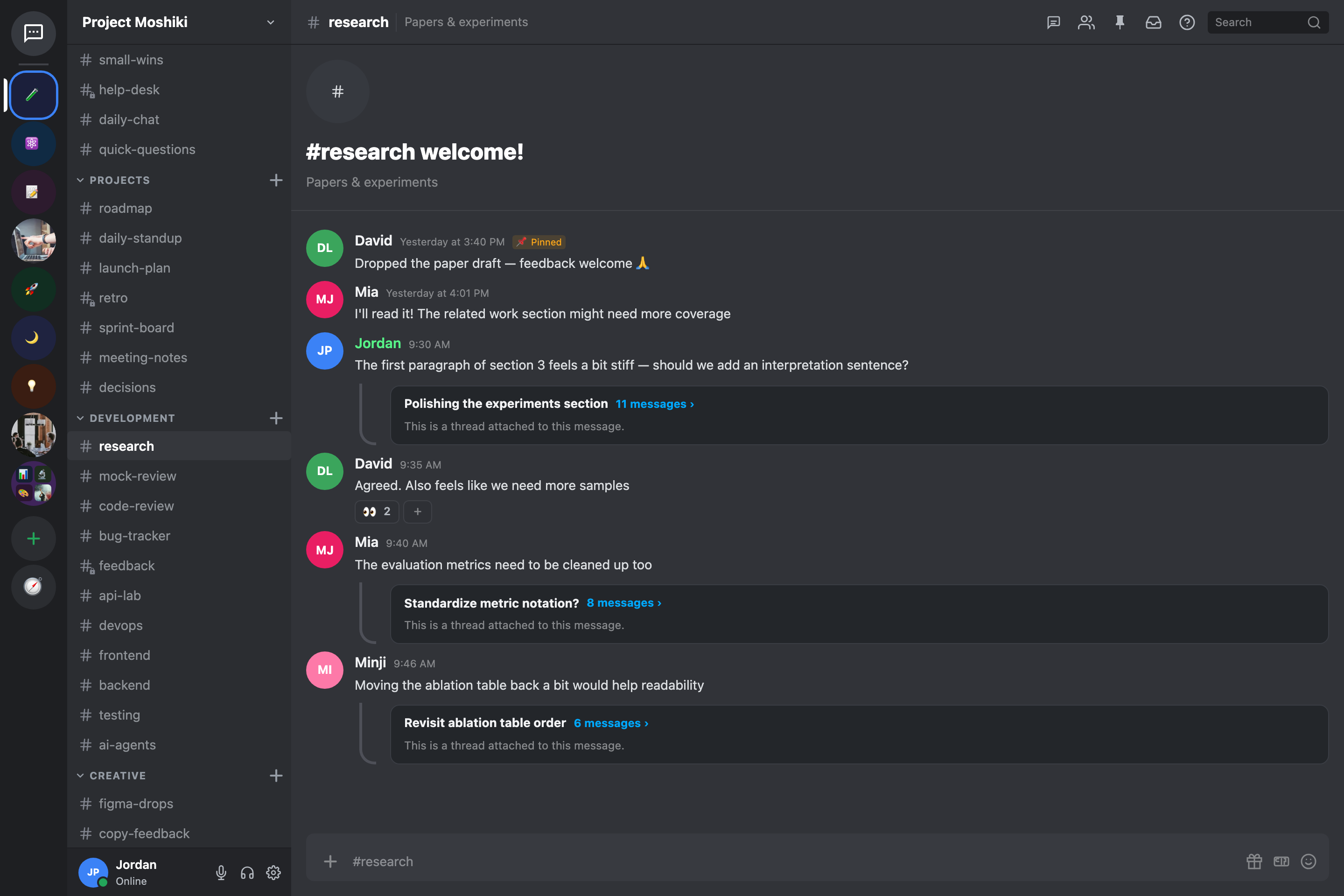}
  \caption{Initial state}
\end{subfigure}
\hfill
\begin{subfigure}[t]{0.49\linewidth}
  \centering
  \includegraphics[width=\linewidth]{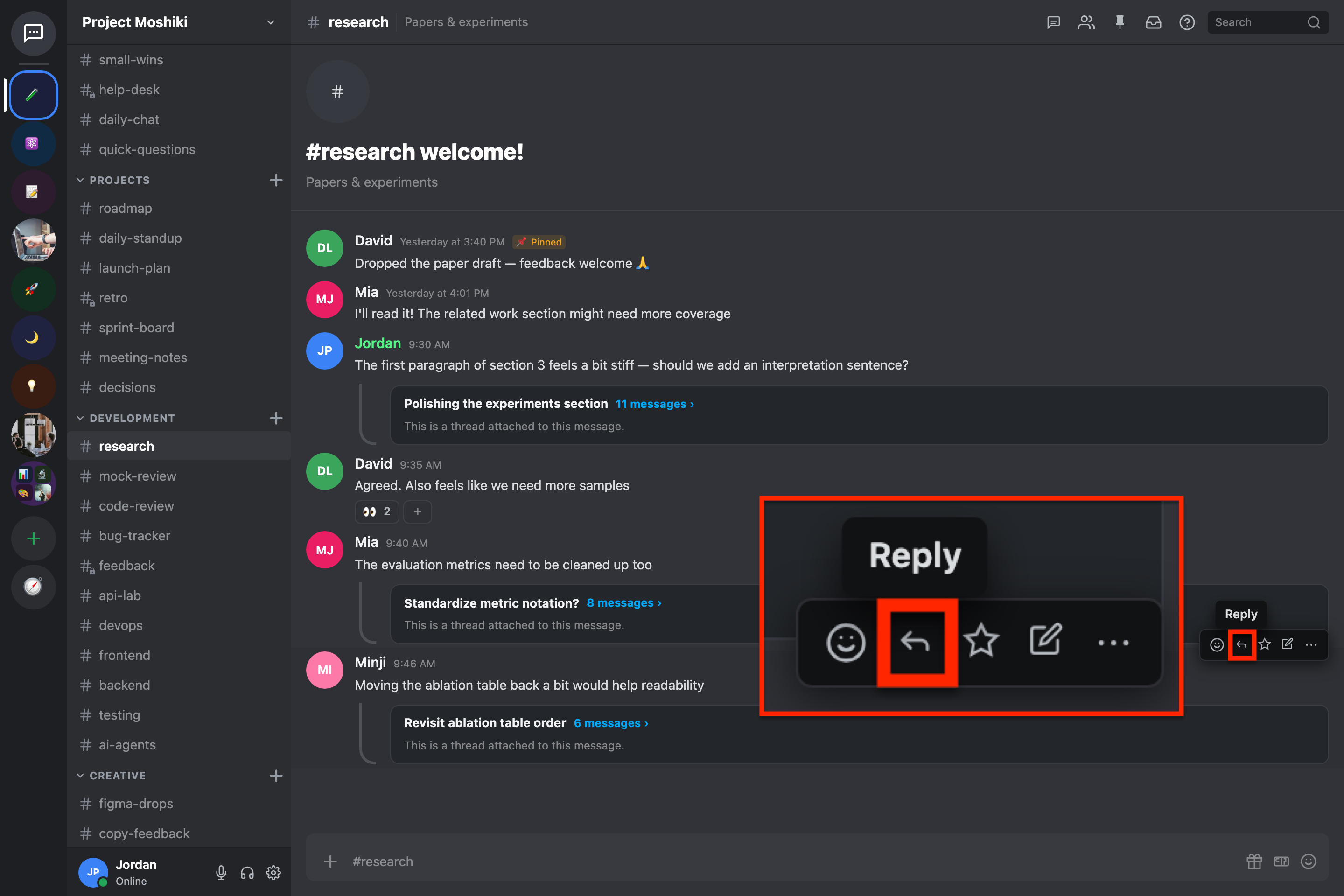}
  \caption{Final state at click time}
\end{subfigure}
\caption{\textbf{\ctypeb{2}{1}: Visual Disambiguation} (Community Chat). \textit{``Click `Reply' on Minji's message.''}}
\label{fig:subtype_l2_1}
\end{figure*}

\begin{figure*}[!htbp]
\centering
\begin{subfigure}[t]{0.49\linewidth}
  \centering
  \includegraphics[width=\linewidth]{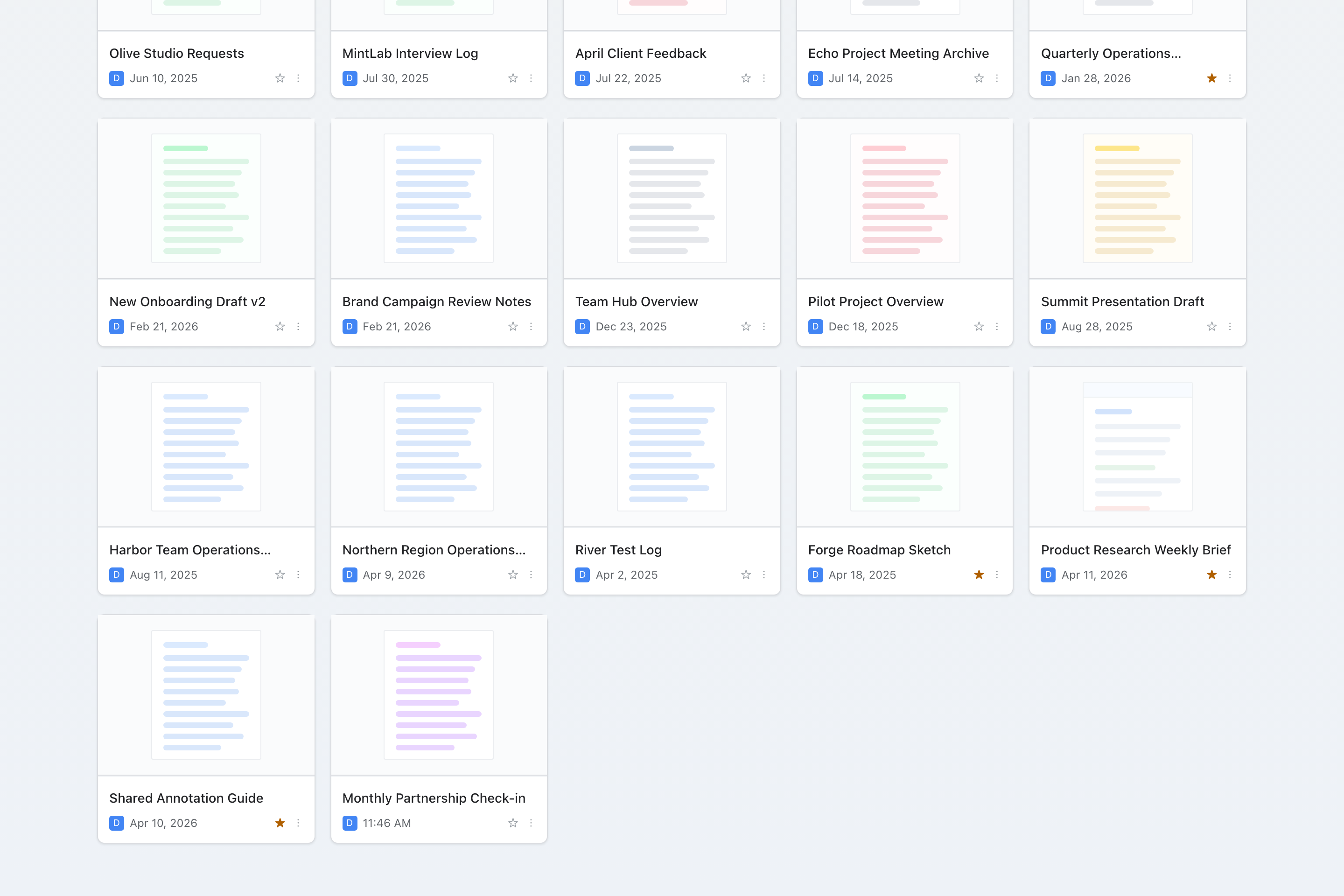}
  \caption{Initial state}
\end{subfigure}
\hfill
\begin{subfigure}[t]{0.49\linewidth}
  \centering
  \includegraphics[width=\linewidth]{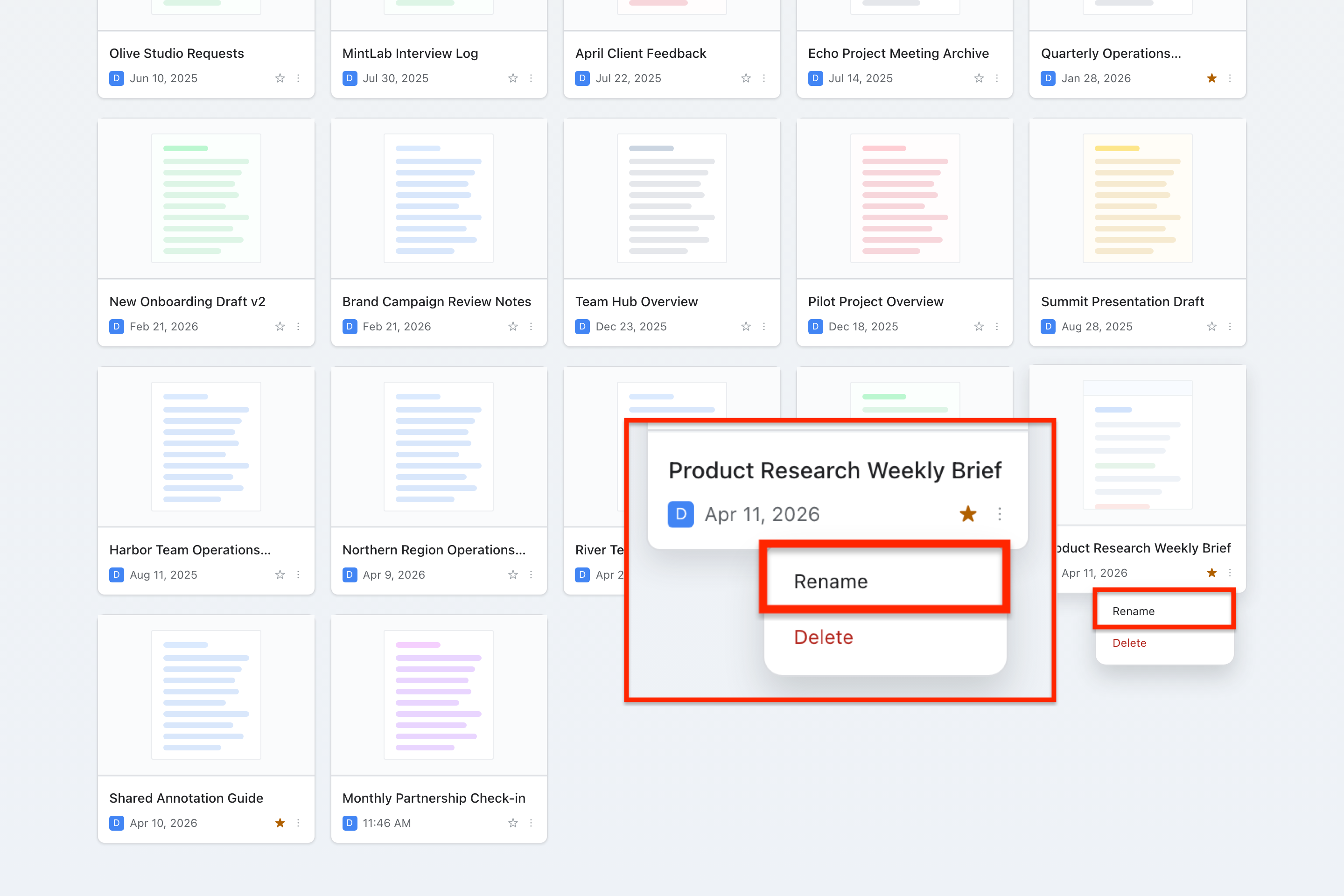}
  \caption{Final state at click time}
\end{subfigure}
\caption{\textbf{\ctypeb{2}{2}: Hidden State Activation} (Document Editor). \textit{``Click `Rename' on `Product Research Weekly Brief'.''}}
\label{fig:subtype_l2_2}
\end{figure*}

\begin{figure*}[!htbp]
\centering
\begin{subfigure}[t]{0.49\linewidth}
  \centering
  \includegraphics[width=\linewidth]{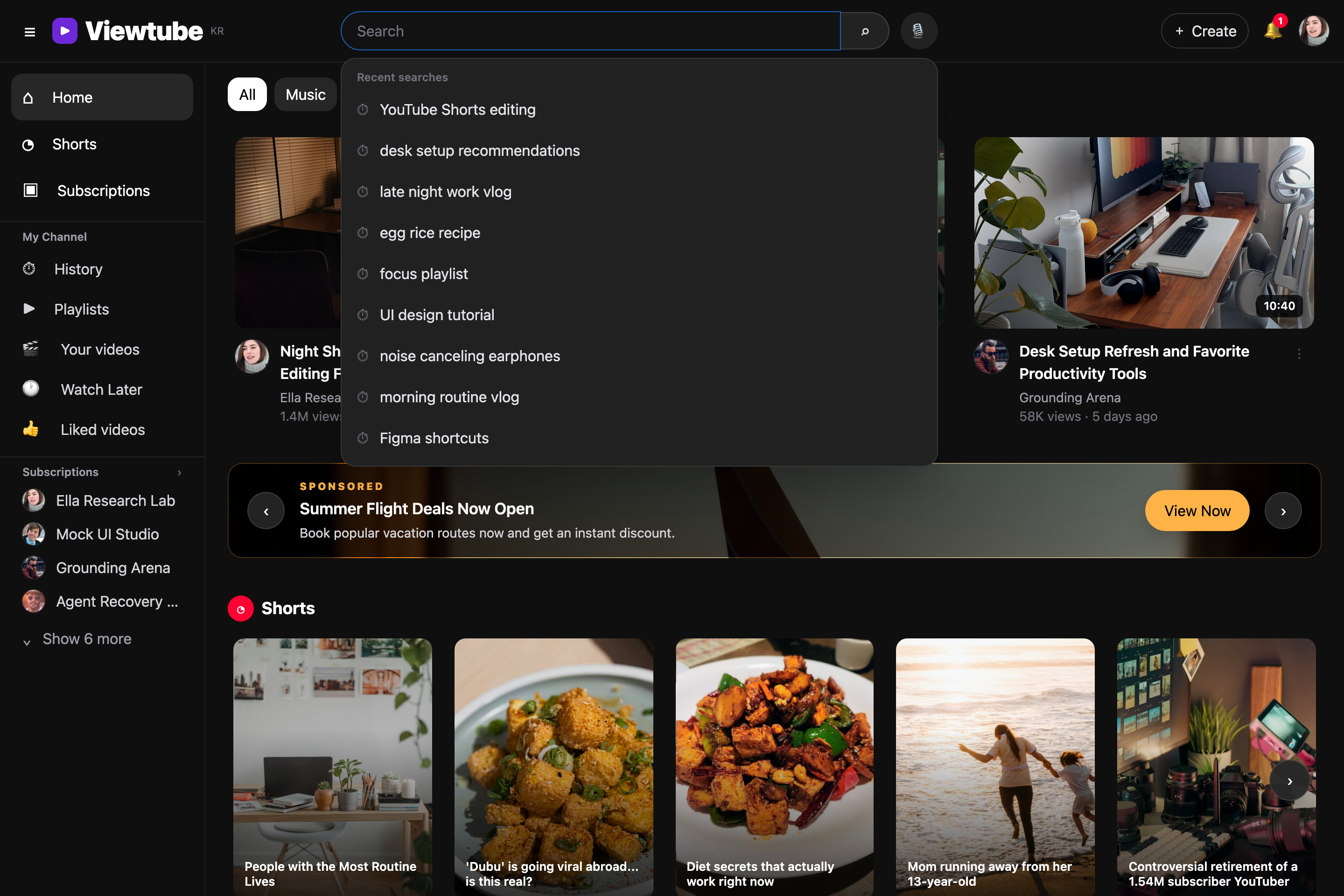}
  \caption{Initial state}
\end{subfigure}
\hfill
\begin{subfigure}[t]{0.49\linewidth}
  \centering
  \includegraphics[width=\linewidth]{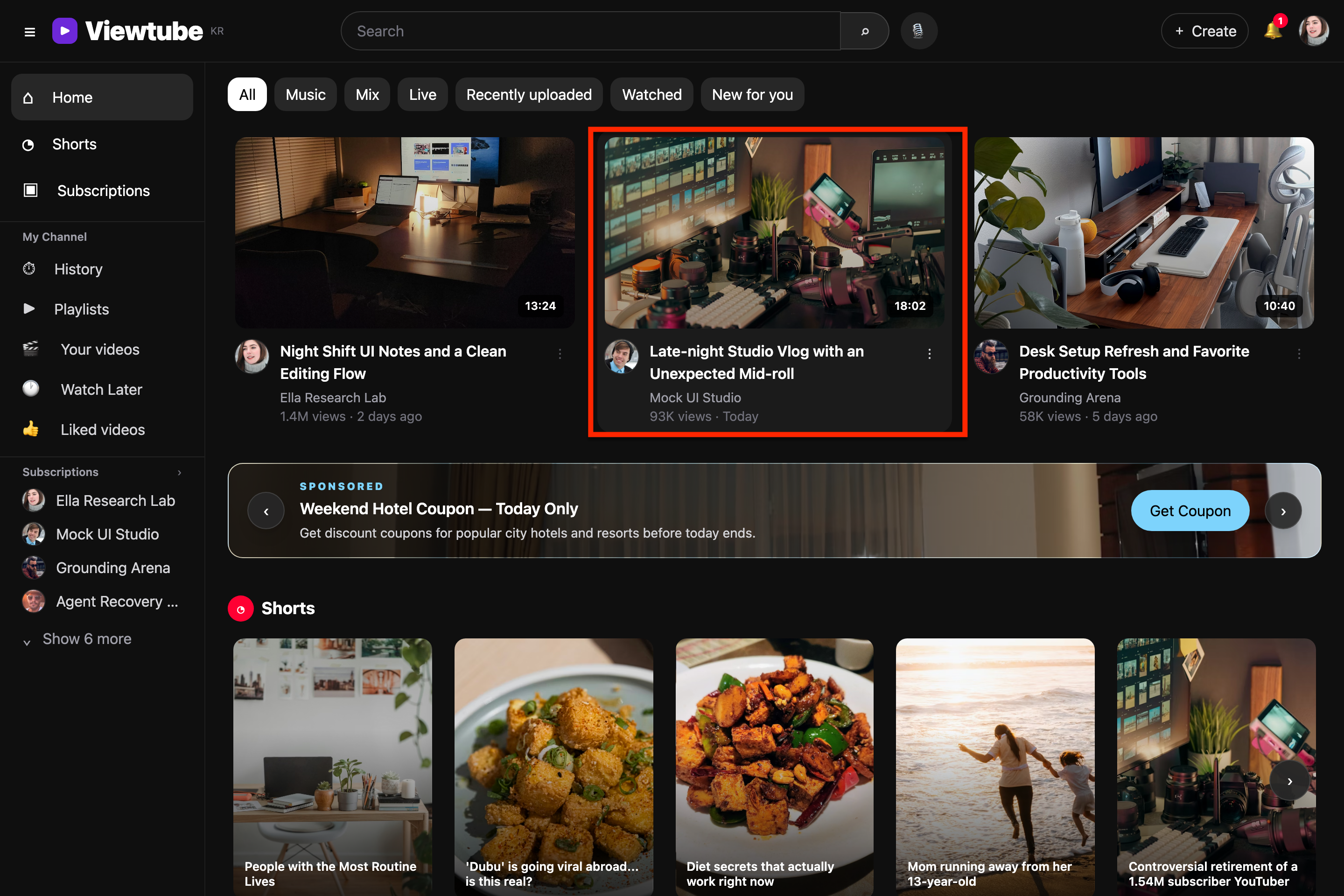}
  \caption{Final state at click time}
\end{subfigure}
\caption{\textbf{\ctypeb{2}{3}: Removable Occlusion} (Video Platform). \textit{``Click the `Late-night Studio Vlog' video card.''}}
\label{fig:subtype_l2_3}
\end{figure*}

\begin{figure*}[!htbp]
\centering
\begin{subfigure}[t]{0.49\linewidth}
  \centering
  \includegraphics[width=\linewidth]{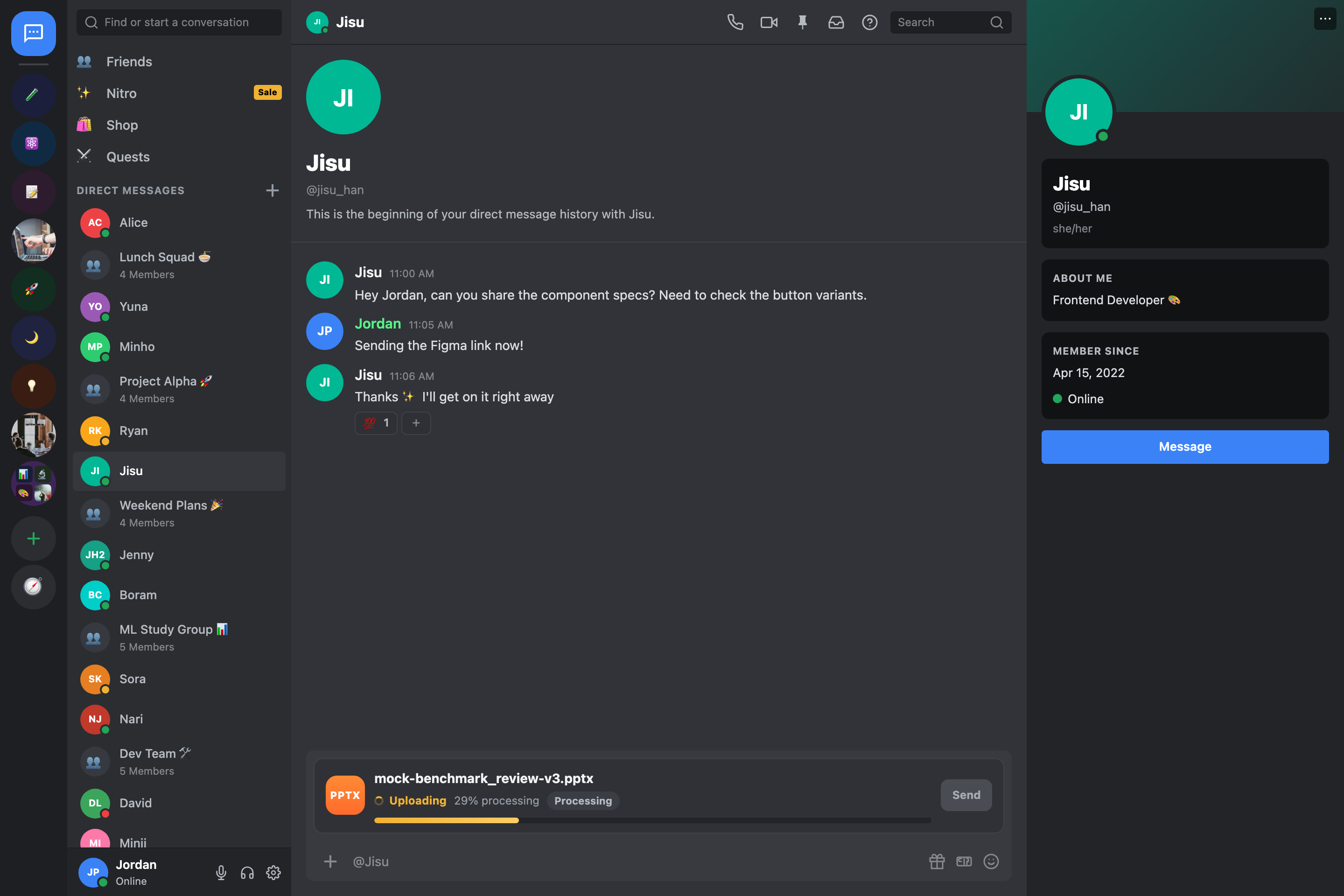}
  \caption{Initial state}
\end{subfigure}
\hfill
\begin{subfigure}[t]{0.49\linewidth}
  \centering
  \includegraphics[width=\linewidth]{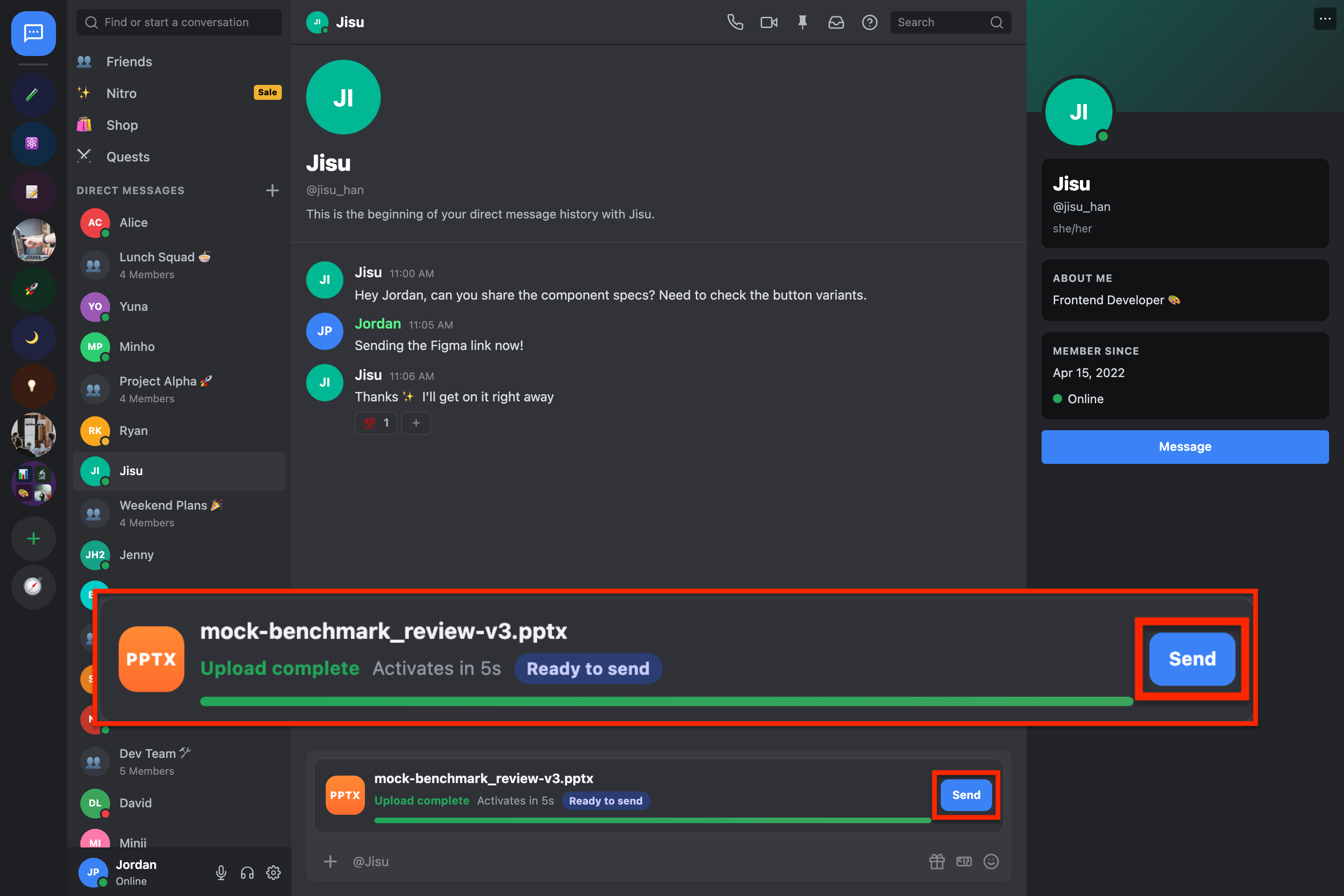}
  \caption{Final state at click time}
\end{subfigure}
\caption{\textbf{\ctypeb{3}{1}: Delayed State Activation} (Community Chat). \textit{``Click `Send' for the pptx upload.''}}
\label{fig:subtype_l3_1}
\end{figure*}

\begin{figure*}[!htbp]
\centering
\begin{subfigure}[t]{0.49\linewidth}
  \centering
  \includegraphics[width=\linewidth]{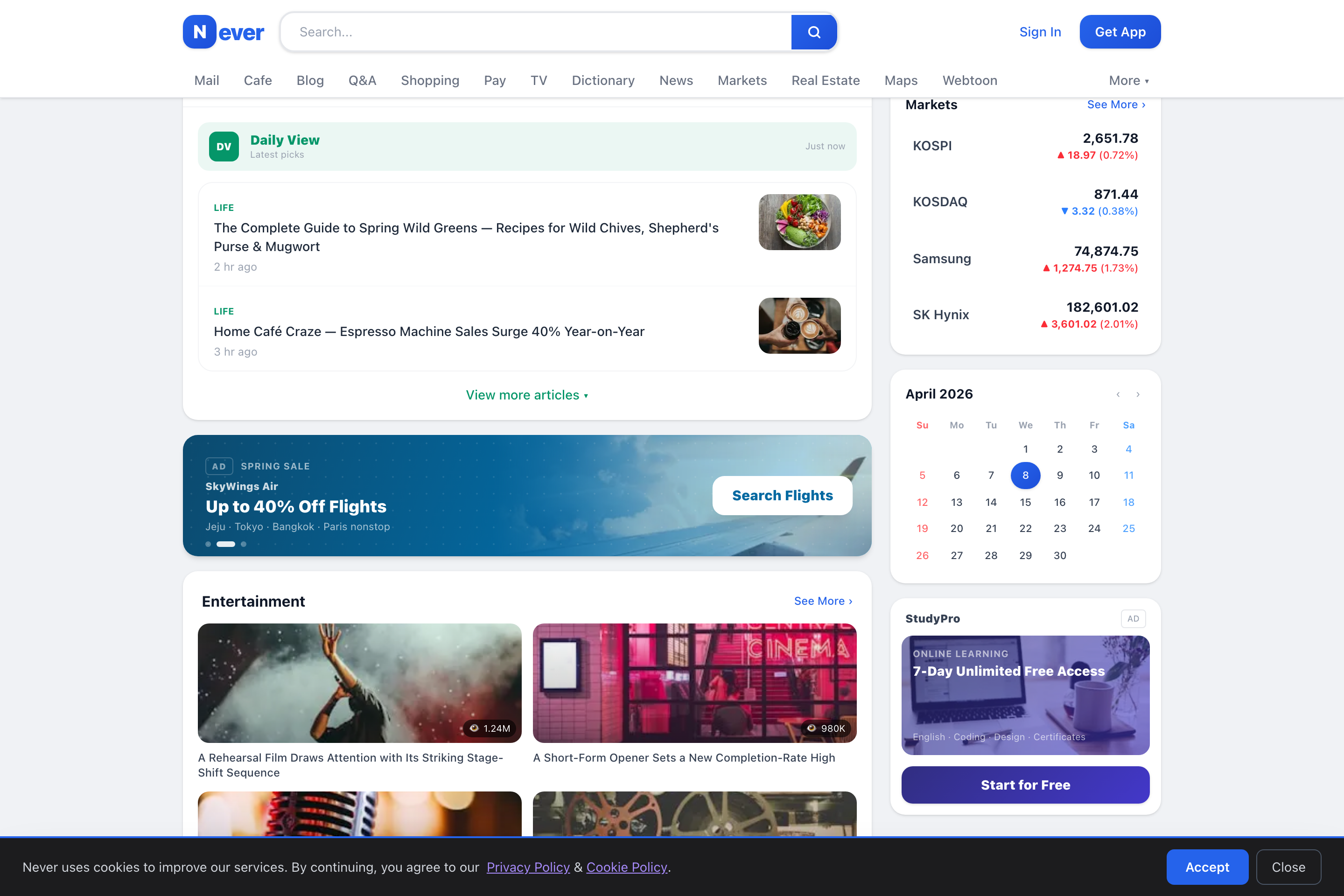}
  \caption{Initial state}
\end{subfigure}
\hfill
\begin{subfigure}[t]{0.49\linewidth}
  \centering
  \includegraphics[width=\linewidth]{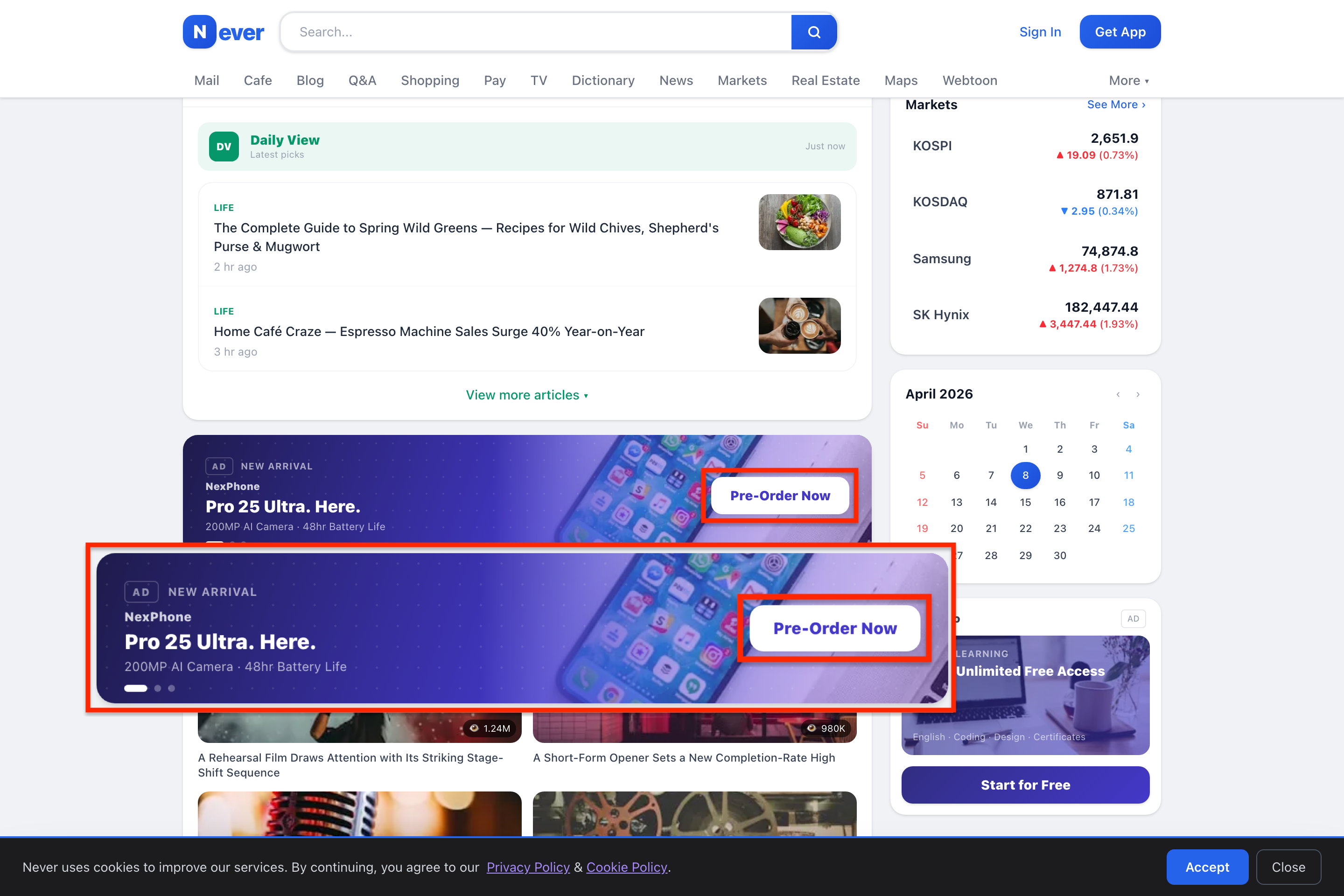}
  \caption{Final state at click time}
\end{subfigure}
\caption{\textbf{\ctypeb{3}{2}: Dynamic Target Interaction} (News Portal). \textit{``Click `Pre-Order Now' in the `NexPhone' ad.''}}
\label{fig:subtype_l3_2}
\end{figure*}

\begin{figure*}[!htbp]
\centering
\begin{subfigure}[t]{0.49\linewidth}
  \centering
  \includegraphics[width=\linewidth]{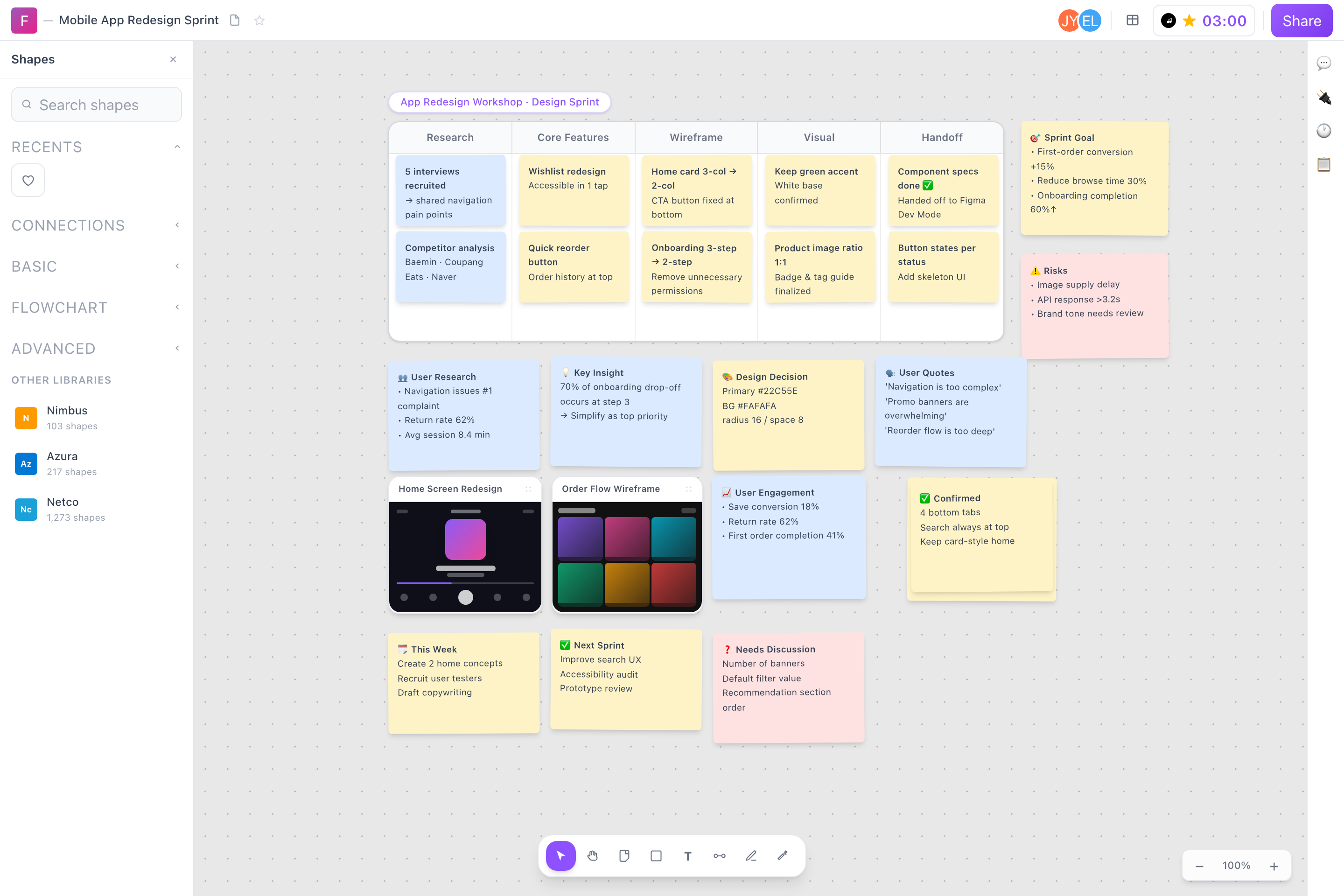}
  \caption{Initial state}
\end{subfigure}
\hfill
\begin{subfigure}[t]{0.49\linewidth}
  \centering
  \includegraphics[width=\linewidth]{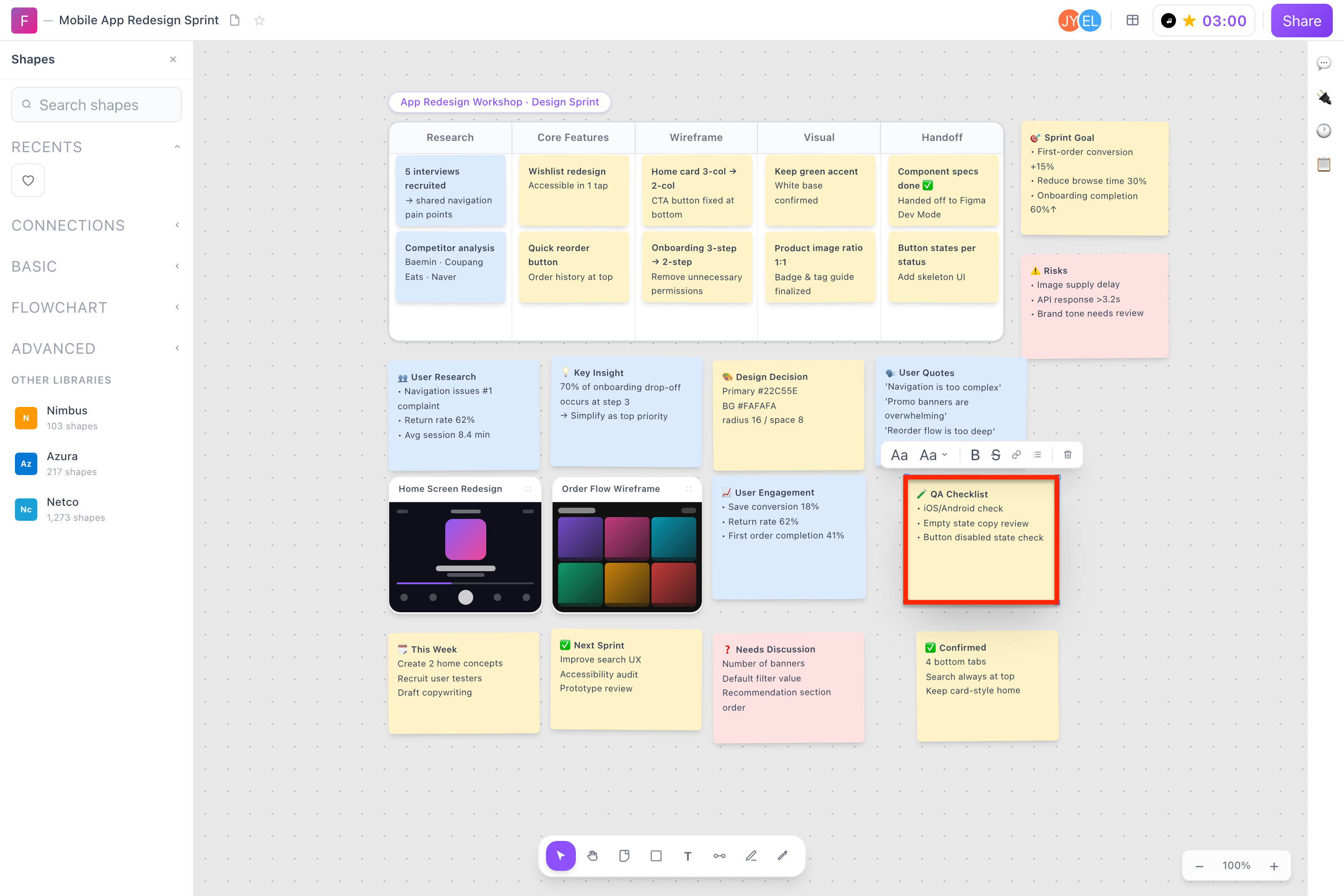}
  \caption{Final state at click time}
\end{subfigure}
\caption{\textbf{\ctypeb{3}{3}: Movable Occlusion} (Design Canvas). \textit{``Click the `QA Checklist' note without deleting any.''}}
\label{fig:subtype_l3_3}
\end{figure*}

Figures~\ref{fig:subtype_l1_1}--\ref{fig:subtype_l3_3} show one representative task for each IGG sub-type. Each example pairs the initial state, where the target is not directly actionable, with the final state at the time of the successful target click.

\section{Failure Case Analysis}
\label{app:failure_analysis}

\subsection{Representative cases}
\label{app:representative-cases}

\begin{figure*}[!htbp]
\centering
\includegraphics[width=\linewidth,height=0.88\textheight,keepaspectratio]{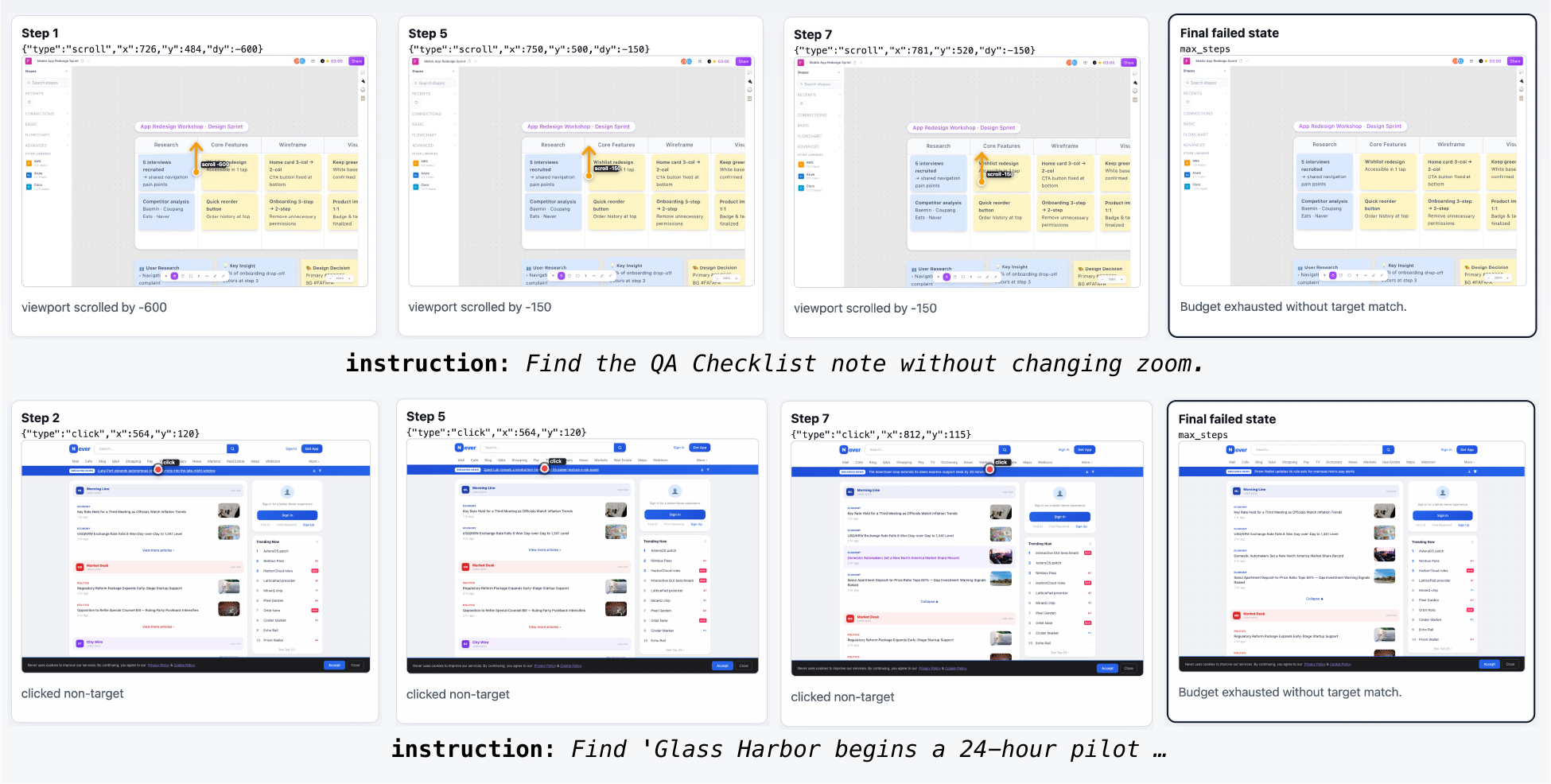}
\caption{\textbf{Representative failed traces.} The figure visualizes example failed traces from two representative cases, showing how models spend the interaction budget without performing the visibility-changing action needed to reveal the target.}
\label{fig:failure_case_grid}
\end{figure*}

Across Qwen2.5-VL, Qwen3-VL, and GUI-Owl-1.5, shared failures often omit the action that would reveal the target. Table~\ref{tab:failure_cases} summarizes three patterns, and Figure~\ref{fig:failure_case_grid} shows representative traces.

\begin{table*}[!htbp]
\centering
\small
\caption{Representative failure cases selected from tasks failed by Qwen2.5-VL-7B, Qwen3-VL-8B, and GUI-Owl-1.5-8B.}
\label{tab:failure_cases}
\setlength{\tabcolsep}{3pt}
\begin{tabular}{p{0.17\linewidth}p{0.14\linewidth}p{0.29\linewidth}p{0.32\linewidth}}
\toprule
\textbf{Failure mode} & \textbf{Sub-type} & \textbf{Task} & \textbf{Observed trace pattern} \\
\midrule
Spatial manipulation is not attempted
& \ctype{3}{3}: Movable Occlusion
& Design Canvas: find the \textit{QA Checklist} note without changing zoom.
& All three models exhausted the budget without using a drag operation. Qwen2.5-VL-7B and Qwen3-VL-8B produced eight scrolls; GUI-Owl-1.5-8B produced one wait followed by seven scrolls. \\
\midrule
Dynamic target state is not tracked
& \ctype{3}{2}: Dynamic Target Interaction
& News Portal: find the \textit{Glass Harbor begins a 24-hour pilot} headline in Breaking News.
& The models interacted with the wrong state of the ticker. Qwen3-VL-8B clicked the ticker controls and ended on a different headline, while the other models scrolled or expanded unrelated page content. \\
\midrule
Hover-based ambiguity is not resolved
& \ctype{2}{1}: Visual Disambiguation
& Community Chat: find the \textit{Creator Dock} server.
& The target server is identifiable only after hovering over ambiguous server icons. Qwen2.5-VL-7B repeatedly scrolled, while Qwen3-VL-8B and GUI-Owl-1.5-8B repeatedly clicked icons without first revealing their labels. \\
\bottomrule
\end{tabular}
\end{table*}

\subsection{Trace-derived action-prior diagnostic}
\label{app:action_prior_diagnostic}

The representative failure cases in Section~\ref{app:representative-cases} show agents spending the budget without the needed reveal action. For each failed Qwen3-VL trajectory, we compare the model actions with the ground-truth trajectory and record whether each required reveal action appears at least once.

\begin{table*}[!htbp]
\centering
\small
\caption{Trace-derived action-prior diagnostic for Qwen3-VL failed trajectories. Counts are non-exclusive because one task can require multiple setup actions.}
\label{tab:action_prior_diagnostic}
\setlength{\tabcolsep}{4pt}
\renewcommand{\arraystretch}{1.08}
\begin{tabular}{@{}lrrrp{0.38\linewidth}@{}}
\toprule
\textbf{Required setup action} & \textbf{Failed tasks} & \textbf{Omitted} & \textbf{Rate} & \textbf{Diagnostic interpretation} \\
\midrule
\textsc{Drag} & 59 & 59 & 100.0\% & Drag is absent from failed traces. \\
\textsc{Scroll} & 247 & 207 & 83.8\% & Scroll is often replaced by clicks or local moves. \\
\textsc{Move} & 96 & 71 & 74.0\% & Hover and disambiguation moves are often skipped. \\
\textsc{Wait} & 3 & 1 & 33.3\% & Few failed tasks require explicit waiting. \\
\bottomrule
\end{tabular}
\end{table*}

\subsection{Target-agnostic reveal heuristics}
\label{app:rule_based_reveal}
To separate reveal patterns that a fixed rule can solve from those that require target-specific decisions, we evaluate five target-agnostic heuristics, each on the sub-type it is designed for: scrolling the page center in fixed steps, hovering over common icon regions in order, clicking common dismiss locations, waiting in one-second steps, and dragging visible occluders toward the screen edge. A heuristic succeeds when the target becomes visible, unobstructed, and actionable (Table~\ref{tab:rule_based_reveal}).

\begin{table*}[t]
\centering
\small
\caption{Reveal success of target-agnostic interaction rules on their
corresponding constraint sub-types.}
\label{tab:rule_based_reveal}
\setlength{\tabcolsep}{5pt}
\renewcommand{\arraystretch}{1.05}
\begin{tabular}{llc}
\toprule
\textbf{Constraint sub-type} &
\textbf{Interaction rule} &
\textbf{Reveal success (\%)} \\
\midrule
\ctype{1}{1}: Scroll-based Exploration
& Scroll center by fixed steps
& 68.8 \\
\ctype{2}{1}: Visual Disambiguation
& Hover common icon areas
& 61.4 \\
\ctype{2}{3}: Removable Occlusion
& Click common dismiss areas
& 17.6 \\
\ctype{3}{1}: Delayed State Activation
& Wait in 1-s steps
& 87.1 \\
\ctype{3}{3}: Movable Occlusion
& Drag visible occluder aside
& 16.9 \\
\bottomrule
\end{tabular}
\end{table*}

Fixed rules often resolve scrolling, hovering, and waiting, but rarely resolve removable or movable occlusions. This agrees with the trace analysis in Section~\ref{app:action_prior_diagnostic}, where \textsc{Drag} is absent from every failed trajectory that requires it.

\section{Training Details}
\label{app:training_details}
Our two-stage procedure uses SFT to introduce rare reveal actions and online GRPO to optimize their execution from browser feedback.

\subsection{Training Setup}
\label{app:training_setup}
Two tasks whose ground-truth trajectories could not be replayed reliably during corpus construction are excluded before splitting the remaining 599 tasks into 479 training tasks and 120 held-out validation tasks. Every included ground-truth trajectory is human-corrected and verified by replay; validation tasks contribute neither SFT examples nor GRPO rollouts.

\subsection{SFT}
\label{app:sft}
The ground-truth trajectories yield 1,141 action-level examples with normalized coordinates. We use LoRA rank 8, alpha 32, dropout 0.05, all-linear language-model target modules, a frozen vision tower and projector, bf16, learning rate $1\times10^{-4}$, effective batch size 8, maximum length 8192, maximum image pixels 1,382,400, and 500 steps with checkpoints every 100 steps. Frequency balancing oversamples \textsc{Click}/\textsc{Scroll}/\textsc{Move}/\textsc{Wait}/\textsc{Drag} by 1/1/2/2/4, expanding the data to 1,460 examples and the drag examples from 55 to 220.

\subsection{Online GRPO}
\label{app:online_grpo}
GRPO starts from an SFT LoRA seed and collects online browser rollouts from the 479 training tasks under the same eight-action budget used in evaluation. We use group size 6, temperature 1.0, top-$p$ 0.95 for rollout sampling, learning rate $2\times10^{-6}$, PPO-style clipping with $\epsilon=0.2$, maximum gradient norm 1.0, and up to 20 update groups. We evaluate saved LoRA checkpoints with greedy decoding on the fixed 120-task validation split.

\paragraph{Online interaction reward.}
The reward is computed from browser-executed traces. Let $S$ denote target-click success; $T$ the action count; $W$ non-empty wrong clicks; $E$ empty clicks; $I$ invalid actions; $A$ repeated ineffective actions; and $R$ repeated wrong-click locations. The bounded bonus $B$ rewards reveal progress (Table~\ref{tab:grpo_reward_components}). Successful traces with prior wrong clicks are capped at 1.2, shortcut successes at 0.8, and runtime or parsing failures are excluded from policy updates.

\subsection{Cross-architecture results}
\label{app:cross_arch}
We apply the same recipe, with the same data, budget, and hyperparameters, to InternVL3.5-8B and GUI-Owl-1.5-8B. All three architectures improve by a similar margin: Qwen3-VL from 40.0\% to 59.2\%, InternVL3.5 from 28.3\% to 47.5\%, and GUI-Owl-1.5 from 35.0\% to 52.5\% (Table~\ref{tab:cross_arch}). The models differ in coordinate tokenization and pretraining, so the gain is not specific to one model family.

\begin{table}[!hbtp]
\centering
\small
\caption{Grounding accuracy before and after SFT+GRPO for three model architectures.}
\label{tab:cross_arch}
\begin{tabular}{lcc}
\toprule
\textbf{Model} & \textbf{Base} & \textbf{SFT+GRPO} \\
\midrule
Qwen3-VL-8B      & 40.0 & 59.2 \\
InternVL3.5-8B   & 28.3 & 47.5 \\
GUI-Owl-1.5-8B   & 35.0 & 52.5 \\
\bottomrule
\end{tabular}
\end{table}

\begin{table*}[!hbtp]
\centering
\small
\caption{GRPO reward components used in the diagnostic training experiments. Positive progress bonuses are bounded per rollout and are applied only when the browser state or target evidence changes in the intended direction.}
\label{tab:grpo_reward_components}
\setlength{\tabcolsep}{3pt}
\renewcommand{\arraystretch}{1.05}

\begin{tabular}{@{}lll@{}}
\toprule
\textbf{Component} & \textbf{Symbol} & \textbf{Value / rule} \\
\midrule
Target-click success & $S$ & $+2.2$ if successful \\
Unsuccessful rollout & $\neg S$ & $-1.0$ base term \\
Step cost & $T$ & $-0.03$ per action \\
Wrong non-empty click & $W$ & $-0.65$ per event \\
Empty click & $E$ & $-1.00$ per event \\
Invalid action & $I$ & $-1.00$ per event \\
Repeated ineffective action & $A$ & $-0.20$ per event \\
Repeated wrong-click location & $R$ & $-0.25$ per event \\
Hover / move progress & $B$ & $+0.45$ \\
Hidden-state progress & $B$ & $+0.45$ \\
Drag progress & $B$ & $+0.75$; $0.7\times$ if no state change \\
Wait progress & $B$ & $+0.35$; $0.5\times$ if no delayed change \\
Clean visible-target click & $B$ & $+0.20$ \\
Reward clipping & -- & $[-4.0,\ 2.8]$ \\
\bottomrule
\end{tabular}
\end{table*}

\begin{table*}[!hbtp]
\centering
\small
\caption{Per-application GA, AC, RER, and TT for Qwen3-VL and GPT-5.5 on all 601 tasks.}
\label{tab:per_app_extended_metrics}
\setlength{\tabcolsep}{3pt}
\renewcommand{\arraystretch}{1.05}
\begin{tabular}{lrrrrrrrr}
\toprule
& \multicolumn{4}{c}{\textbf{Qwen3-VL}} & \multicolumn{4}{c}{\textbf{GPT-5.5}} \\
\cmidrule(lr){2-5}\cmidrule(lr){6-9}
\textbf{App} & \textbf{GA} & \textbf{AC} & \textbf{RER} & \textbf{TT} & \textbf{GA} & \textbf{AC} & \textbf{RER} & \textbf{TT} \\
\midrule
Social Feed & 29.4 & 6.47 & 0.93 & 17514 & 61.8 & 4.59 & 1.04 & 69757 \\
Planner & 21.2 & 6.88 & 0.79 & 13098 & 81.8 & 3.88 & 0.83 & 61011 \\
Developer Hub & 25.7 & 6.94 & 0.66 & 9763 & 57.1 & 5.57 & 0.69 & 138919 \\
Code Editor & 21.9 & 6.81 & 1.26 & 9575 & 87.5 & 3.88 & 0.97 & 59747 \\
Community Chat & 26.5 & 6.82 & 0.64 & 9357 & 61.2 & 4.82 & 1.03 & 108102 \\
Document Editor & 37.5 & 6.05 & 0.99 & 8451 & 85.0 & 3.48 & 1.04 & 63538 \\
Design Canvas & 8.9 & 7.44 & 1.25 & 13951 & 82.2 & 4.33 & 0.94 & 93199 \\
Spreadsheet & 33.3 & 6.24 & 0.84 & 8920 & 83.3 & 2.93 & 1.04 & 124023 \\
Email Client & 50.0 & 5.38 & 0.75 & 10493 & 88.1 & 3.38 & 0.97 & 63120 \\
News Portal & 29.0 & 6.55 & 0.80 & 19204 & 71.0 & 4.77 & 0.80 & 52984 \\
Workspace Pages & 34.3 & 6.40 & 0.86 & 8830 & 85.7 & 3.71 & 0.93 & 30489 \\
Streaming Service & 20.5 & 7.00 & 0.67 & 17224 & 87.2 & 4.05 & 0.71 & 52801 \\
E-commerce & 43.3 & 6.40 & 0.60 & 17047 & 86.7 & 4.40 & 0.72 & 49700 \\
Music Player & 36.0 & 5.80 & 1.15 & 12580 & 86.0 & 3.60 & 1.04 & 63597 \\
Maps & 26.7 & 6.47 & 1.25 & 11972 & 70.0 & 4.53 & 0.95 & 166930 \\
Video Platform & 50.0 & 5.91 & 0.63 & 8158 & 94.1 & 3.15 & 0.88 & 38354 \\
\bottomrule
\end{tabular}
\end{table*}

\section{Human Study Details}
\label{app:human_study}

\subsection{Study setup}

This appendix provides implementation details for the two human studies summarized in Section~\ref{sec:human_study}. Fifteen participants completed two browser-based protocols connected to the same environments as model evaluation. In the solvability study, every participant completed a subtype-balanced suite of 21 tasks---three from each of the seven sub-types and at least one from every environment---yielding 315 trials. In the fidelity study, participants compared seven controlled environments with real-interface references and rated visual realism, task executability, and interaction realism on five-point scales.

\subsection{Task-solvability interface}

\begin{figure*}[!htbp]
\centering

\begin{minipage}{0.33\linewidth}
    \centering
    \includegraphics[width=\linewidth]{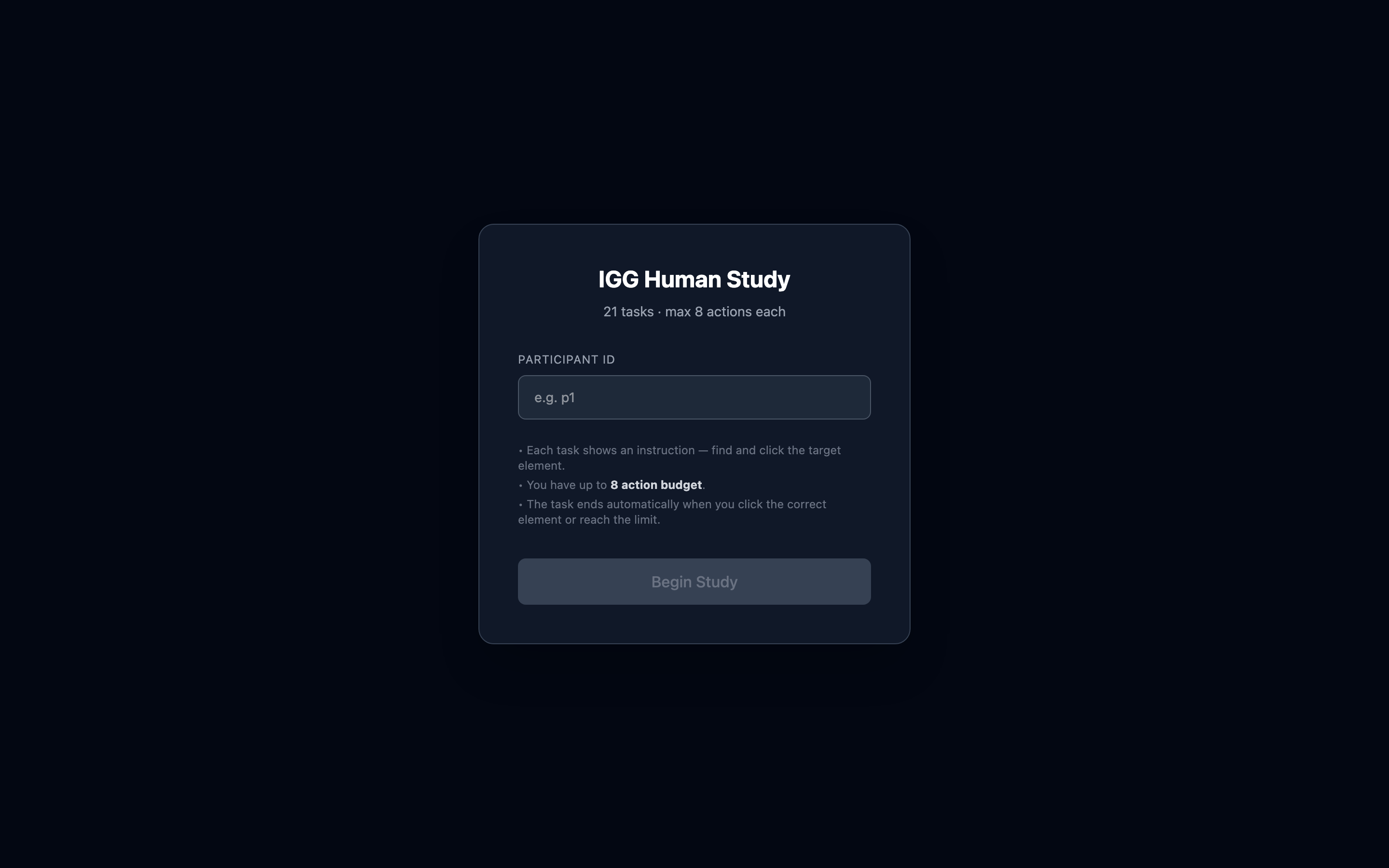}
    \\[2pt]\small (a) Start page
\end{minipage}\hspace{2pt}%
\begin{minipage}{0.33\linewidth}
    \centering
    \includegraphics[width=\linewidth]{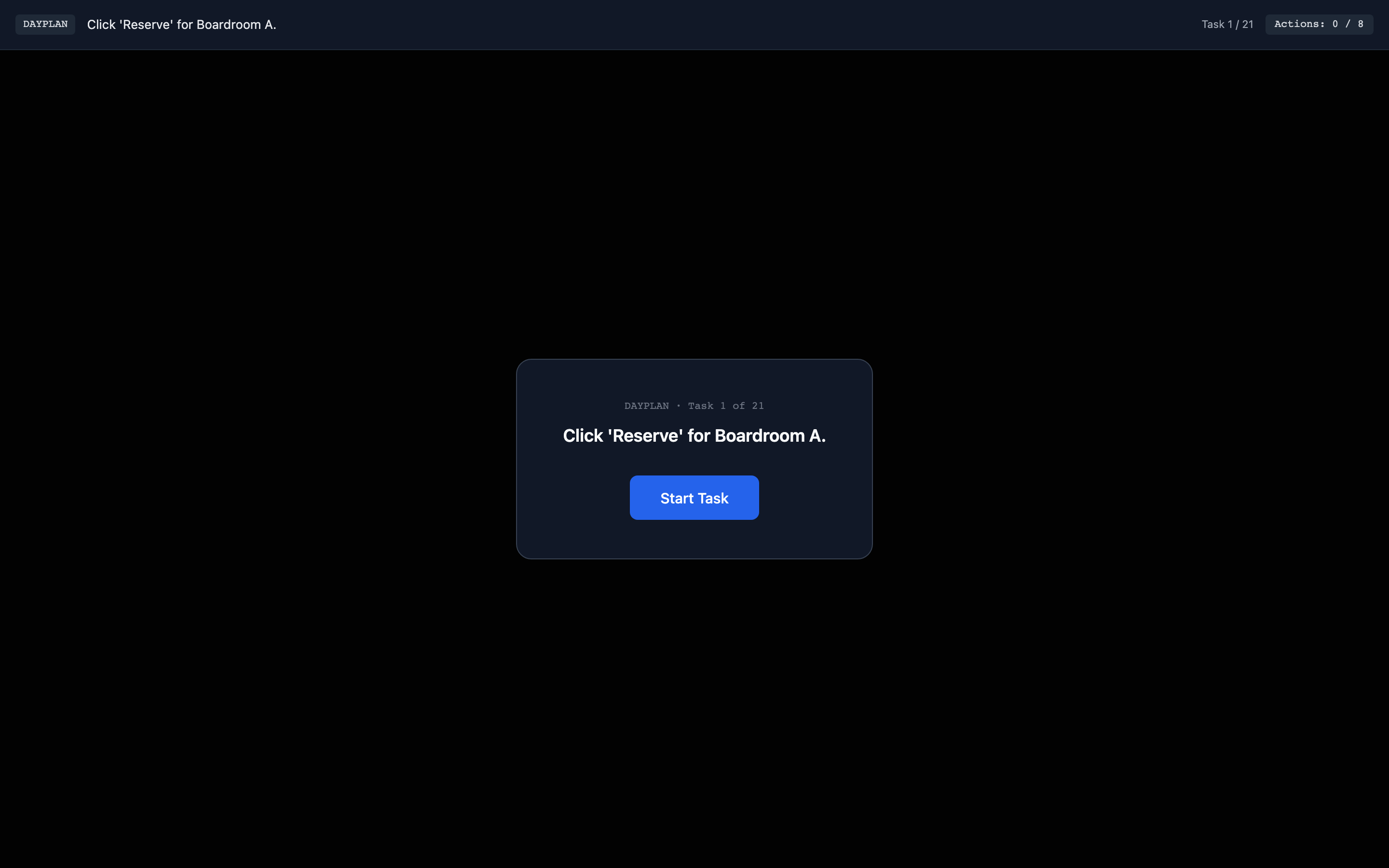}
    \\[2pt]\small (b) Task instruction
\end{minipage}\hspace{2pt}%
\begin{minipage}{0.33\linewidth}
    \centering
    \includegraphics[width=\linewidth]{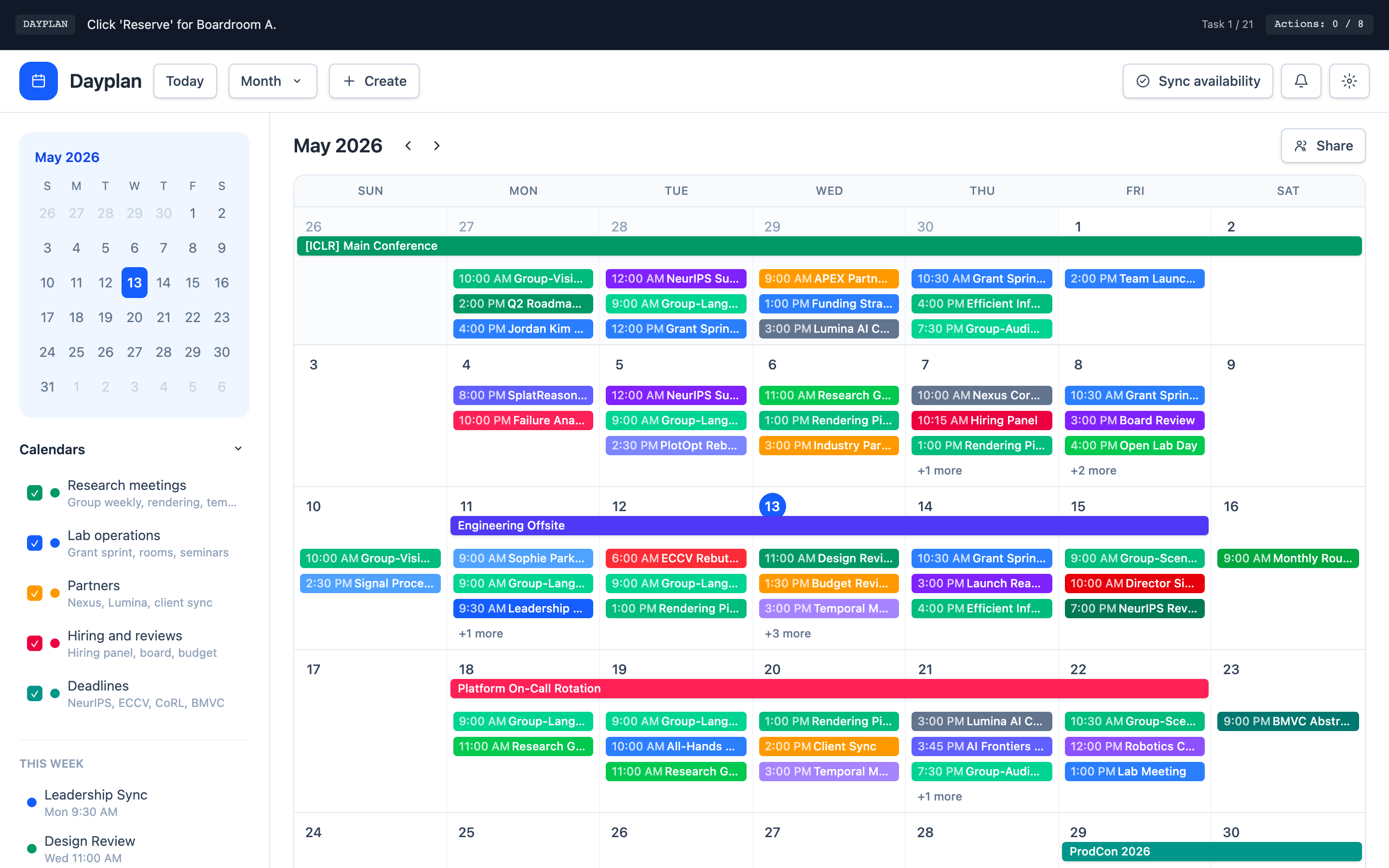}
    \\[2pt]\small (c) Task in progress
\end{minipage}

\vspace{4pt}

\centerline{%
\begin{minipage}{0.495\linewidth}
    \centering
    \includegraphics[width=\linewidth]{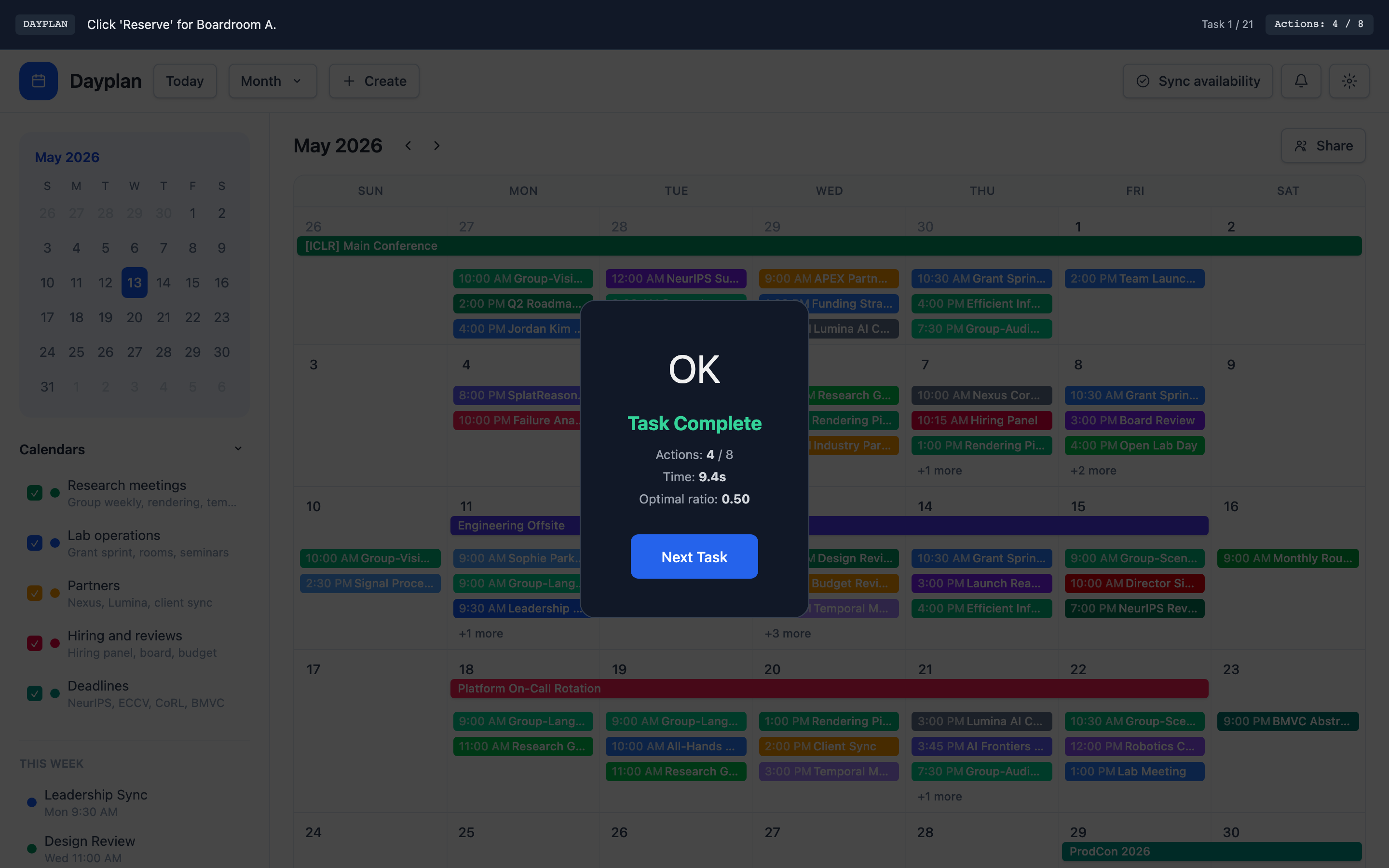}
    \\[2pt]\small (d) Task complete (success)
\end{minipage}\hspace{2pt}%
\begin{minipage}{0.495\linewidth}
    \centering
    \includegraphics[width=\linewidth]{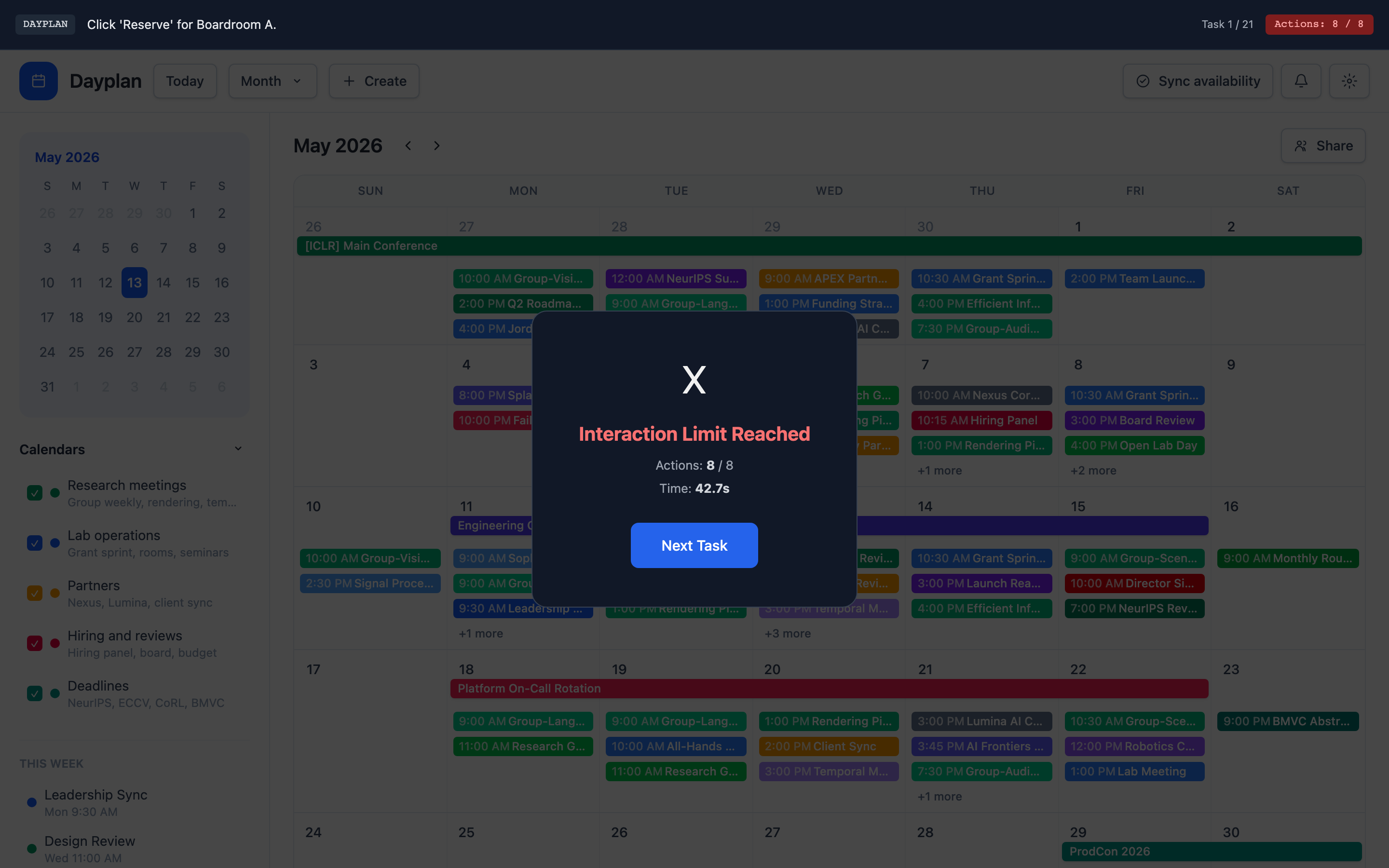}
    \\[2pt]\small (e) Interaction limit reached (failure)
\end{minipage}%
}

\caption{The web-based interface used in the human study.
(a) Participants begin by entering an ID and reviewing the task format.
(b) Each task is introduced with the instruction text.
(c) Participants then interact with the live mockup; the interaction counter is shown in the top-right corner.
(d) The task ends with success when the participant clicks the correct target, or
(e) with failure when the eight click-or-drag attempt limit is reached.}
\label{fig:human_solvability_interface}
\end{figure*}

Participants receive one instruction and interact directly with the browser. The interface ends a trial when the participant clicks the target or reaches eight click or drag attempts. It records scroll sessions after an 800-ms debounce for interaction counts and records hover events without applying them to the attempt limit (Figure~\ref{fig:human_solvability_interface}).

\subsection{Fidelity-test interface}

\begin{figure*}[!htbp]
\centering

\begin{minipage}{0.49\linewidth}
    \centering
    \includegraphics[width=\linewidth]{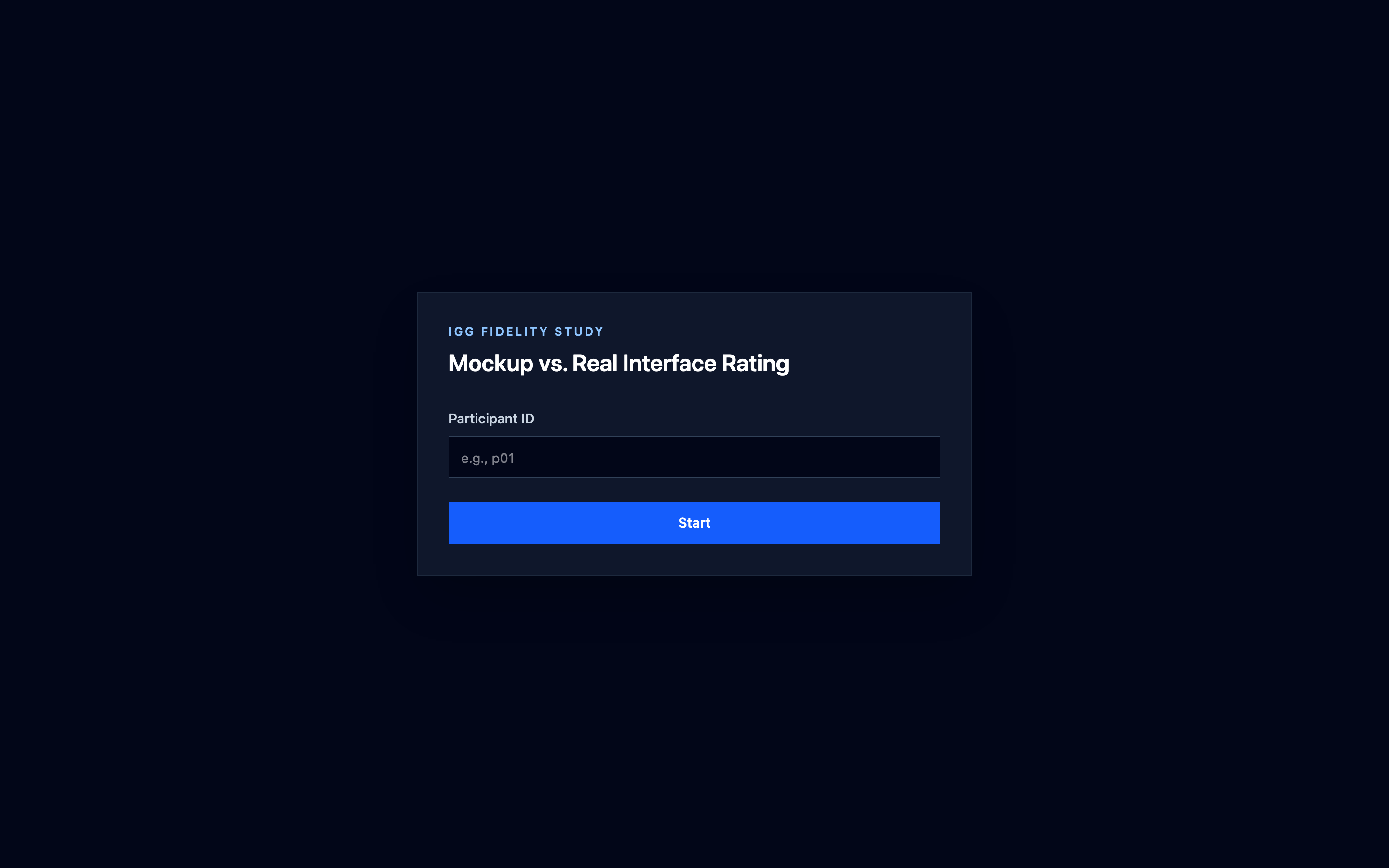}
    \\[2pt]\small (a) Start page
\end{minipage}\hfill
\begin{minipage}{0.49\linewidth}
    \centering
    \includegraphics[width=\linewidth]{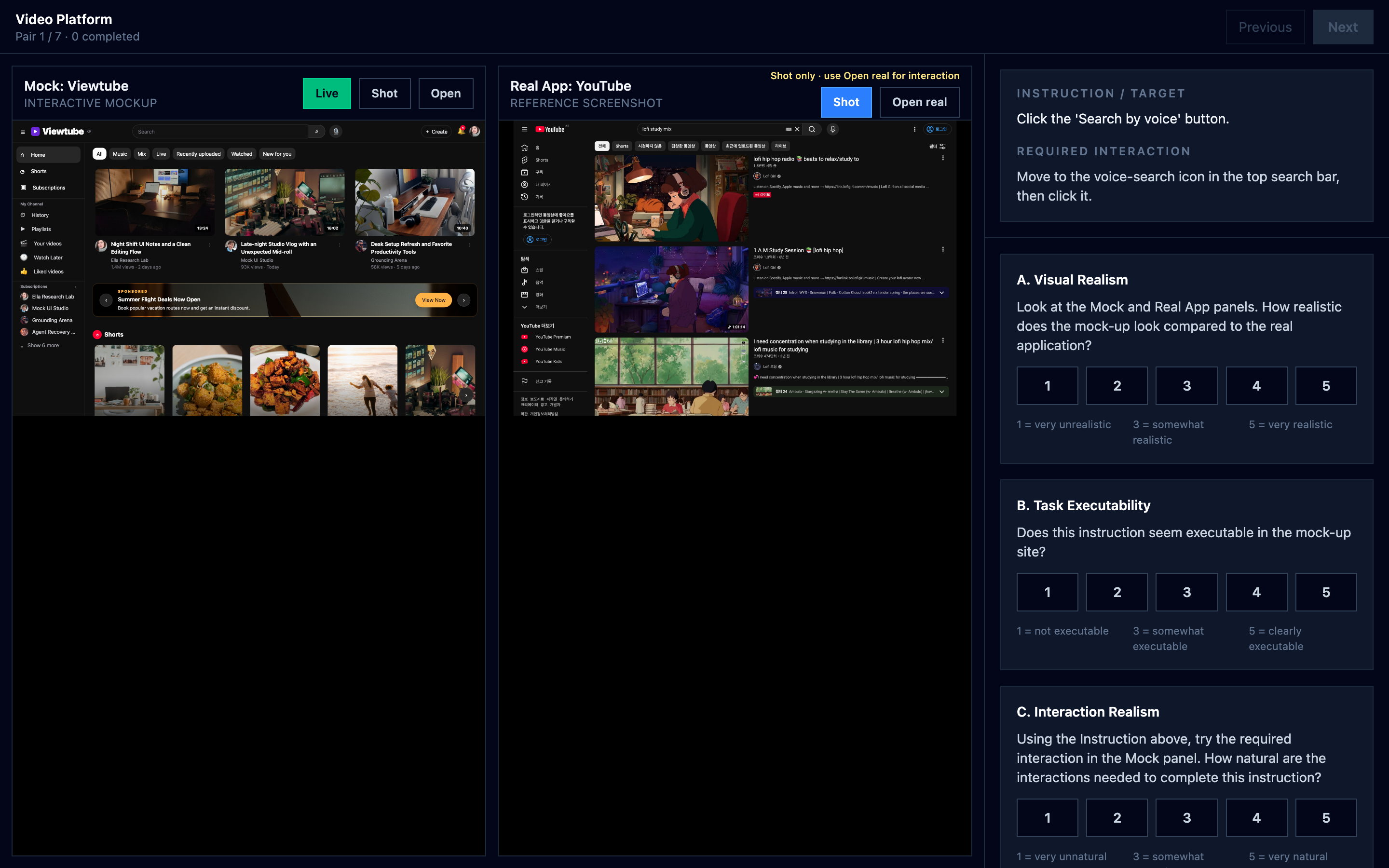}
    \\[2pt]\small (b) Rating interface
\end{minipage}

\caption{The web-based interface used in the human fidelity test.
(a) Participants enter the fidelity-study session.
(b) For each application, participants compare the mock interface with the real-application reference and rate visual realism, task executability, and interaction realism.}
\label{fig:human_fidelity_interface}
\end{figure*}

For each paired example, participants see the instruction, controlled environment, and real-interface reference, then rate visual realism, task executability, and interaction realism (Figure~\ref{fig:human_fidelity_interface}).

\end{document}